\pdfoutput=1
\documentclass{article}

\usepackage{etoolbox, placeins,siunitx,etoc}

\usepackage{tcolorbox, caption}

\definecolor{lightgreen}{HTML}{DFFFD6}
\definecolor{lightred}{HTML}{FFD6D6}
\definecolor{lightyellow}{HTML}{FFFFD6}
\definecolor{lightblue}{HTML}{D6F0FF}
\definecolor{lightgray}{HTML}{E0E0E0}

\definecolor{midblue}{HTML}{B3E6FF}

    \usepackage[main,final]{neurips_2026}
    \usepackage[utf8]{inputenc} 
    \usepackage[T1]{fontenc}    
    \usepackage{hyperref}       
    \usepackage{url}            
    \usepackage{booktabs}       
    \usepackage{amsfonts}       
    \usepackage{nicefrac}       
    \usepackage{microtype}      
    \usepackage{xcolor}   
    \usepackage{algorithm}
    \usepackage{algorithmic}

\usepackage{amsmath}
\usepackage{amssymb}
\usepackage{mathtools}
\usepackage{amsthm}

\newtheorem{assumption}{Assumption}
\newtheorem{proposition}{Proposition}
\newtheorem{theorem}{Theorem}

\newtheorem{remark}{Remark}
\newtheorem{lemma}{Lemma}

    \title{Expert-Guided Bayesian Optimization for Sustainable Protein Design}
\author{%
  \bfseries
  Anna Thomas\textsuperscript{1,2}\thanks{\texttt{thomasat@cs.stanford.edu}} \quad
  Georgios Zaverdinos\textsuperscript{3} \quad
  Petros Mandalis\textsuperscript{3} \quad
  Andreas Orfanoudakis\textsuperscript{3} \\
  \bfseries
  Akihiro Takino\textsuperscript{3} \quad
  Sohum Patnaik\textsuperscript{2} \quad
  David J. Kraus\textsuperscript{3} \quad
  Panos Kostopoulos\textsuperscript{3} \\
  \bfseries
  Caroline Cotto\textsuperscript{4} \quad
  Nikos Tsiaparas\textsuperscript{3} \quad
  Dan Jurafsky\textsuperscript{1} \quad
  Aadit Patel\textsuperscript{5} \\[0.5em]
  \normalfont
  \textsuperscript{1}Stanford Computer Science \quad
  \textsuperscript{2}Food Intelligence Lab \quad
  \textsuperscript{3}Proxy Foods AI \\
  \textsuperscript{4}NECTAR \quad
  \textsuperscript{5}Bayes \& Gambit
}

\begin{document}

    \maketitle
\begin{abstract}
Agentic laboratories give LLM agents control of experiments, alongside scientific tools such as optimizers and code execution. How should such systems choose the next experiment? We study whether an LLM agent should propose experiments directly, delegate them to Bayesian optimization (BO), or select the design space itself and delegate refinement. We formalize the third option as Expert-Guided Bayesian Optimization (EGBO), in which an expert (e.g., an LLM or a human) selects a low-dimensional subspace for BO and may adaptively expand it over time. We decompose EGBO's suboptimality into a selection gap and an optimization gap, and characterize the coverage--dimension tradeoff governing when expert guidance helps. To support \textit{in silico} prototyping before costly laboratory work, we introduce \textsc{FormulateBench}, a suite of 24 sustainable protein formulation tasks. In batched experiments that mirror our laboratory deployment, EGBO outperforms conventional BO. The strongest LLMs match EGBO without noise but underperform it under noise, even when told the noise level. When deployed to optimize two plant-based dairy products, EGBO improves utility, as assessed by a trained human panel, by 29\% and 26\% in 10 batched iterations each. In a comparison with a professional food scientist given the same time budget, EGBO reached a best observed utility of 0.992, vs.\ 0.850 for the food scientist.
\end{abstract}

\section{Introduction}\label{sec:intro}
Agentic laboratories, in which LLM agents control experiments alongside scientific tools such as optimizers and code execution, are increasingly used to accelerate scientific discovery~\citep{abolhasani2023rise,roohani2025biodiscoveryagent,fu2026agentic}. A central question for such systems is how to choose the next experiment. One option is for the LLM to propose experiments directly. LLMs bring broad prior knowledge, but direct comparisons find them competitive mainly at small budgets~\citep{zhang2023using,liu2024large} and insufficiently sensitive to experimental feedback~\citep{gupta2025llms}. Another option is to delegate to Bayesian optimization (BO)~\citep{snoek2012practical,frazier2018tutorial}, which uses feedback efficiently but lacks domain knowledge, and whose sample efficiency deteriorates sharply as the number of variables grows~\citep{srinivas2012information,bull2011convergence}. Existing hybrids let the LLM propose candidates, restart a search that has plateaued, or reshape BO's search space~\citep{chang2025llinbo,cisse2025bora,han2025chembomas,brunzema2026agentic}, but which division of labor works best, and under what conditions, has not been systematically characterized.

We study this question in the challenging testbed of sustainable protein formulation. Animal agriculture accounts for roughly 16.5\% of global greenhouse gas emissions and is a leading driver of pollution, biodiversity loss, and pandemic risk~\citep{gerber2013tackling,twine2021emissions,espinosa2020infectious}. Sustainable proteins, such as plant-based, fermentation-enabled, and cultivated products, can substantially reduce these impacts~\citep{poore2018reducing,gfiEnvironmentalImpacts,henchion2017future}. Adoption, however, depends on matching the sensory and nutritional properties of animal products~\citep{onwezen2021systematic,ulhas2023review}. Designing such a formulation is expensive trial and error, because each candidate must be physically prepared and evaluated by a human sensory panel~\citep{beckley2008accelerating,redefinemeatRoleSensory}. As in much of experimental science, candidates are prepared and evaluated in batches, and the measurements are noisy. The design space is also large. Candidate ingredients and processing steps can number in the thousands, although a good formulation typically uses only a small subset of them.

We formalize a division of labor between the LLM and BO as \emph{Expert-Guided Bayesian Optimization} (EGBO). An expert selects which variables to optimize over, and BO determines their values within the resulting low-dimensional subspace. The expert can be an LLM agent or a human scientist, so EGBO also formalizes common practice in ML-assisted discovery, where an expert narrows the design space before search begins~\citep{lao2023gptuner,roohani2025biodiscoveryagent}. This practice is double-edged. When the expert is right, sample efficiency can improve dramatically, but when the expert is wrong, the optimizer can underperform random search. Because EGBO concentrates the expert's contribution into one interpretable choice, expert reliability reduces to two parameters, recall and false-positive rate, which lets us bound performance as a function of expert quality.

Our contributions include:

\begin{enumerate}
\item \textbf{A formal division of labor between expert and optimizer.} We formalize EGBO, in which an expert selects a subspace and BO optimizes within it, optionally expanding the subspace over time. We decompose EGBO's error into a selection gap, the expert's responsibility, and an optimization gap, BO's responsibility, and we measure both throughout the paper. On a controlled synthetic benchmark (Hartmann6 embedded in 200 dimensions), expert guidance helps only when the expert's recall is high and false positives are rare, and a weak expert can do worse than random search.
\item \textbf{\textsc{FormulateBench} and when to delegate to an optimizer.} To support evaluating formulation methods before costly laboratory work, we introduce \textsc{FormulateBench}, 24 plant-based formulation tasks (14 meat, 10 dairy) grounded in real nutritional targets. We implement the three strategies (direct iterative LLM optimization, conventional BO, and EGBO) as fixed pipelines. For every LLM tested, most of the error in its first recipe comes from the concentrations rather than the choice of ingredients, which is precisely the part that BO refines in EGBO. In our target setting of batched, noisy experiments, EGBO outperforms conventional BO and outperforms direct LLM optimization with all three LLMs tested, significantly for two. Without noise, the strongest LLMs match EGBO. Given code execution, none of three LLM agents ran a BO loop, even when told that BoTorch was installed, and most runs fit no model of their data at all.
\item \textbf{Real-world deployment and human comparison.} We deploy LLM-guided EGBO with a food company to optimize two plant-based dairy products, improving sensory utility, as assessed by a trained human panel, by 29\% and 26\% in 10 iterations. In a comparison with a professional food scientist given the same laboratory time, EGBO reached a best observed utility of 0.992, vs.\ 0.850.
\end{enumerate}

\section{Related Work}
\paragraph{LLMs, BO, and agentic laboratories.} LLMs have been used as optimizers in their own right~\citep{yang2024large,zhang2023using} and as components of BO, for example to warm-start the search or act as the surrogate~\citep{liu2024large}. In experimental design, LLM agents can be insensitive to experimental feedback and outperformed by classical methods~\citep{gupta2025llms}. Hybrid systems divide the work in different ways, from an LLM that intervenes when BO stagnates~\citep{cisse2025bora} or partitions the search space~\citep{han2025chembomas} to an agent that steers a BO backend~\citep{brunzema2026agentic}. Closest to EGBO, GPTuner uses an LLM to select database configuration knobs before optimization~\citep{lao2023gptuner}. We instead compare direct LLM optimization, BO, and expert-guided BO on the same tasks, characterize when each is preferable, and deploy the result in a laboratory.

\paragraph{High-dimensional BO.} %
BO has been scaled to high dimensions with random embeddings (REMBO~\citep{wang2016bayesian}), trust regions (TuRBO~\citep{eriksson2019scalable}), and sparsity-inducing priors or penalties (SAASBO~\citep{eriksson2021high}, SEBO~\citep{liu2023sparse}). These methods must infer the relevant subspace from evaluations, consuming budget that EGBO saves by obtaining the initial subspace from the expert.
We discuss additional related work in Appendix~\ref{app:extended-related-work}.

\section{Expert-Guided Bayesian Optimization}
\label{sec:egbo}
Though directly motivated by sustainable protein formulation, EGBO applies to any setting in which an expensive black-box function \(f: \mathcal{X} \to \mathbb{R}\) must be maximized over a high-dimensional domain \(\mathcal{X} = \prod_{i=1}^N [0,u_i^{\max}] \subseteq \mathbb{R}^N\), and where high-quality solutions are expected to depend on only a small subset of the \(N\) variables. In the sustainable protein setting, \(f\) could be the panel-assessed sensory score of a plant-based formulation parameterized by ingredients and processing steps. 

EGBO maintains an active set of variables \(S_r \subseteq [N]\) at each round $r$, and optimizes over the restricted domain $\mathcal{X}_{S_r} = \{x \in \mathcal{X}: x_i = 0 \text{ for all } i \notin S_r\}$.
The algorithm has \(Q\) rounds. In the first round, the expert proposes an initial active set \(S_1\). In later rounds, the expert may inspect the optimization history and add more variables. The active set grows monotonically: $S_1 \subseteq S_2 \subseteq \cdots \subseteq S_Q$.
\(Q=1\) and \(Q>1\) correspond to one-shot and adaptive EGBO, respectively. Let \(T\) be the total evaluation budget, including an initial-design budget of \(n_{\mathrm{init}}\). We allocate the remaining \(T-n_{\mathrm{init}}\) evaluations across the \(Q\) BO rounds as \(T_1,\ldots,T_Q\), with \(n_{\mathrm{init}}+\sum_{r=1}^Q T_r=T\). The initial design may include an expert-proposed point, as in \textsc{FormulateBench}, or may be generated using a standard space-filling design such as Sobol sampling. Earlier evaluations remain valid when the active set expands because \(\mathcal{X}_{S_r}\subseteq\mathcal{X}_{S_{r+1}}\). Algorithm~\ref{alg:egbo} describes the procedure.

\begin{algorithm}[ht]
\caption{Expert-Guided Bayesian Optimization (EGBO)}
\label{alg:egbo}
\begin{algorithmic}[1]
\REQUIRE Domain \(\mathcal{X}\), expert \(\mathcal{E}\), total budget \(T\), expert rounds \(Q\), initial design budget \(n_{\mathrm{init}}\), round budgets \(T_1,\ldots,T_Q\) satisfying \(n_{\mathrm{init}}+\sum_{r=1}^Q T_r=T\)
\STATE Initialize dataset \(\mathcal{D} \leftarrow \emptyset\)
\STATE Query expert \(\mathcal{E}\) for initial active set \(S_1 \subseteq [N]\)
\STATE Construct and evaluate \(n_{\mathrm{init}}\) initial points in \(\mathcal{X}_{S_1}\), and add them to \(\mathcal{D}\)
\STATE Fit the Gaussian process surrogate using \(\mathcal{D}\)
\FOR{\(r = 1, \dots, Q\)}
    \FOR{\(t = 1, \dots, T_r\)}
        \STATE Choose \(x \in \mathcal{X}_{S_r}\) using the BO acquisition function
        \STATE Evaluate \(y = f(x) + \varepsilon\)
        \STATE Add \((x,y)\) to \(\mathcal{D}\) and update the Gaussian process surrogate
    \ENDFOR
    \IF{\(r < Q\)}
        \STATE Query expert \(\mathcal{E}\) with the current history \(\mathcal{D}\) for additional variables \(A_{r+1} \subseteq [N]\setminus S_r\)
        \STATE Set \(S_{r+1} \leftarrow S_r \cup A_{r+1}\)
    \ENDIF
\ENDFOR
\STATE \textbf{return} \(\hat{x}_T \in \arg\max_{(x,y)\in \mathcal{D}} y\)
\end{algorithmic}
\end{algorithm}

\subsection{Theoretical Analysis}\label{sec:theory}
To provide a simple analysis of the performance of Algorithm~\ref{alg:egbo}, we make three assumptions. 
A central feature of sustainable protein formulation problems is that high-quality sparse solutions are rarely unique. A target sensory profile may be achievable via multiple disjoint ingredient sets.

\begin{assumption}[Family of sparse approximate optima]\label{asm:sparsity}
There exist $k \in \mathbb{N}$ and $\eta \geq 0$ such that the family
\begin{equation}
\begin{aligned}
  \mathcal{R}_\eta = \bigl\{ R \subseteq [N] : {}& |R| \leq k, \\
    & \exists\, \tilde{x}^{(R)} \in \mathcal{X} \text{ with } \mathrm{supp}(\tilde{x}^{(R)}) \subseteq R \\
    & \text{and } f(\tilde{x}^{(R)}) \geq f(x^*) - \eta \bigr\}
\end{aligned}
\end{equation}
is non-empty. We refer to $\tilde{x}^{(R)}$ as a \emph{witness} for support $R$, a specific $k$-sparse near-optimum supported on $R$. We call $\eta$ the \emph{sparsity gap}, $k$ the \emph{target sparsity level}, and $M = |\mathcal{R}_\eta|$ the \emph{multiplicity}. The \emph{support union} $\mathcal{U}_\eta = \bigcup_{R \in \mathcal{R}_\eta} R$ collects all variables that appear in some sparse near-optimum, with $\kappa = |\mathcal{U}_\eta|$.
\end{assumption}

Assumption~\ref{asm:sparsity} is strictly weaker than assuming a unique sparse global optimum, since it requires only that \emph{some} $k$-sparse point achieves near-optimal value.

We characterize expert quality relative to $\mathcal{U}_\eta$:

\begin{assumption}[Expert quality]\label{asm:expert}
At each round $q$, the expert's additions to $S_{q+1}$ are independent across coordinates conditional on $S_q$, with fresh draws each round. Every variable $i \in \mathcal{U}_\eta$ not yet in $S_q$ is added with probability $p \in (0, 1]$ (\emph{recall}); every $j \notin \mathcal{U}_\eta$ not yet in $S_q$ is added with probability $q_{\mathrm{fp}} \in [0, 1)$ (\emph{false-positive rate}). In the one-shot setting ($Q = 1$), each $i \in \mathcal{U}_\eta$ is included in $S$ with probability $p$ and each $j \notin \mathcal{U}_\eta$ with probability $q_{\mathrm{fp}}$.
\end{assumption}

We additionally require a guarantee from the BO subroutine used within each restricted domain.

\begin{assumption}[BO subroutine guarantee]\label{asm:bo}
For any active set $S \subseteq [N]$, when the BO subroutine is run on $\mathcal{X}_S$ with $T_S$ evaluations whose acquisition functions are each maximized over $\mathcal{X}_S$, it returns $\hat{x}_{T_S}$ satisfying
$f_S^\star - f(\hat{x}_{T_S}) \leq \varepsilon_{\mathrm{BO}}(T_S, |S|)$
with probability at least $1 - \delta_{\mathrm{BO}}$, where $\varepsilon_{\mathrm{BO}}(T, d)$ is non-increasing in $T$ and non-decreasing in $d$.
\end{assumption}

We use this as an abstract finite-sample guarantee for the BO subroutine. Different BO algorithms satisfy different versions of such a guarantee under different smoothness, kernel, and noise assumptions. We state our results in terms of the generic $\varepsilon_{\mathrm{BO}}$. %
Throughout, we assume the expert's variable selection randomness and the BO subroutine's randomness are independent.

EGBO helps only if two events occur. First, the expert must choose an active set that contains a good sparse solution. Second, BO must successfully optimize inside that active set. We capture this with the decomposition
\begin{equation}\label{eq:decomp}
  f(x^*) - f(\hat{x}_T)
  =
  \underbrace{f(x^*) - f_S^\star}_{\text{selection gap}}
  +
  \underbrace{f_S^\star - f(\hat{x}_T)}_{\text{optimization gap}},
\end{equation}
where $f_S^\star = \max_{x \in \mathcal{X}_S} f(x)$.
The selection gap measures the cost of restricting the search to the expert-selected subspace. The optimization gap measures BO's suboptimality within that subspace. If \(S\) contains the support of a sparse near-optimal witness \(R \in \mathcal{R}_\eta\), then the restricted domain \(\mathcal{X}_S\) contains a point whose value is within \(\eta\) of the global optimum. Therefore,
$f(x^*) - f_S^\star \leq \eta$.
The remaining error is then controlled by the BO subroutine, which operates in dimension \(|S|\) rather than the ambient dimension \(N\).

\begin{theorem}[One-shot EGBO]\label{thm:oneshot}
Suppose Assumptions~\ref{asm:sparsity},~\ref{asm:expert}, and~\ref{asm:bo} hold. Consider one-shot EGBO: the expert is queried once to obtain an active set \(S\), and BO is then run on \(\mathcal{X}_S\) for \(T\) evaluations. Fix any witness support \(R \in \mathcal{R}_\eta\). With probability at least $p^{|R|}(1-\delta_{\mathrm{BO}})
  \geq
  p^k(1-\delta_{\mathrm{BO}})$,
the returned point \(\hat{x}_T\) satisfies
\begin{equation}\label{eq:oneshot_bound}
  f(x^*) - f(\hat{x}_T)
  \leq
  \eta + \varepsilon_{\mathrm{BO}}(T, |S|).
\end{equation}
Moreover, under Assumption~\ref{asm:expert}, $\mathbb{E}[|S|] =
  \kappa p + (N-\kappa)q_{\mathrm{fp}}$,
where $\kappa = |\mathcal{U}_\eta|$.
\end{theorem}

The theorem says that one-shot EGBO succeeds when the expert-selected active set contains a sparse near-optimal support. In that case, the total error is at most the sparsity gap \(\eta\) plus the error from running BO in the lower-dimensional space \(\mathcal{X}_S\). When there are many different sparse near-optimal supports, the probability of covering at least one of them can be much higher. If \(\mathcal{R}_\eta\) contains \(M'\) pairwise disjoint supports of size at most \(k\), then
\[
  \Pr(\exists R \in \mathcal{R}_\eta : R \subseteq S)
  \geq
  1 - (1-p^k)^{M'}.
\]

\begin{proposition}[Coverage under adaptive expansion]\label{prop:adaptive_coverage}
Suppose EGBO performs \(Q\) expert-query rounds with monotonic active-set growth, $S_1 \subseteq S_2 \subseteq \cdots \subseteq S_Q$.
Under Assumption~\ref{asm:expert}, define $\pi_Q = 1-(1-p)^Q$.
For any fixed witness support \(R \in \mathcal{R}_\eta\), $\Pr(R \subseteq S_Q)
  =
  \pi_Q^{|R|}$.
In particular, $\Pr(R \subseteq S_Q)
  \geq
  \max\{0,\,1-k(1-p)^Q\}$.
The expected final active-set size is
\begin{equation}\label{eq:adaptive_size}
  \mathbb{E}[|S_Q|]
  =
  \kappa\bigl(1-(1-p)^Q\bigr)
  +
  (N-\kappa)\bigl(1-(1-q_{\mathrm{fp}})^Q\bigr).
\end{equation}
\end{proposition}

Proposition~\ref{prop:adaptive_coverage} shows the main tradeoff in adaptive EGBO. More expert-query rounds increase the probability of covering a sparse near-optimal support, because each relevant variable has more chances to be added. However, more rounds also increase the expected number of false-positive variables. Thus adaptivity can improve the selection gap, but may make the BO problem harder by increasing \(|S_Q|\).

\begin{remark}[When one-shot EGBO improves over vanilla BO]\label{rem:crossover}
The one-shot bound shows that EGBO improves over vanilla BO when the cost of using the expert-selected subspace is smaller than the benefit of optimizing in lower dimension. In particular, if \(S\) contains a witness support, then EGBO is better than vanilla BO whenever
\begin{equation}\label{eq:crossover_oneshot}
  \eta + \varepsilon_{\mathrm{BO}}(T, |S|)
  <
  \varepsilon_{\mathrm{BO}}(T, N).
\end{equation}
The left-hand side is the EGBO bound, a selection gap \(\eta\) plus a BO optimization gap in dimension \(|S|\). The right-hand side is the corresponding BO optimization gap in the full ambient dimension \(N\).
\end{remark}

All proofs are in Appendix~\ref{app:proofs}.

\section{Experimental Evaluation}\label{sec:exp-evaluation}
We first evaluate EGBO under controlled expert quality on Hartmann6 embedded in 200 dimensions (Section~\ref{sec:synthetic_benchmark}), then with Claude Opus 4.7 as the expert on \textsc{FormulateBench}'s 24 plant-based formulation tasks. We compare against vanilla BO, random search, REMBO, SEBO, TuRBO, SAASBO, BO on \(k\) randomly selected variables, and fixed-support and unrestricted iterative LLM optimization. All methods receive \(T=20\) objective evaluations (30 in the batched experiments). Stochastic optimizers use 10 seeds (20 for the Hartmann6 $Q$-sweep), while each temperature-zero iterative LLM baseline contributes one cached trajectory per category. Appendix~\ref{app:exp-details} gives full experimental details. Our code is available at \url{https://github.com/Food-Intelligence-Lab/egbo}.

\subsection{Synthetic Experiments on Hartmann6}\label{sec:synthetic_benchmark}

\begin{figure}[t]
\centering
\includegraphics[width=\textwidth]{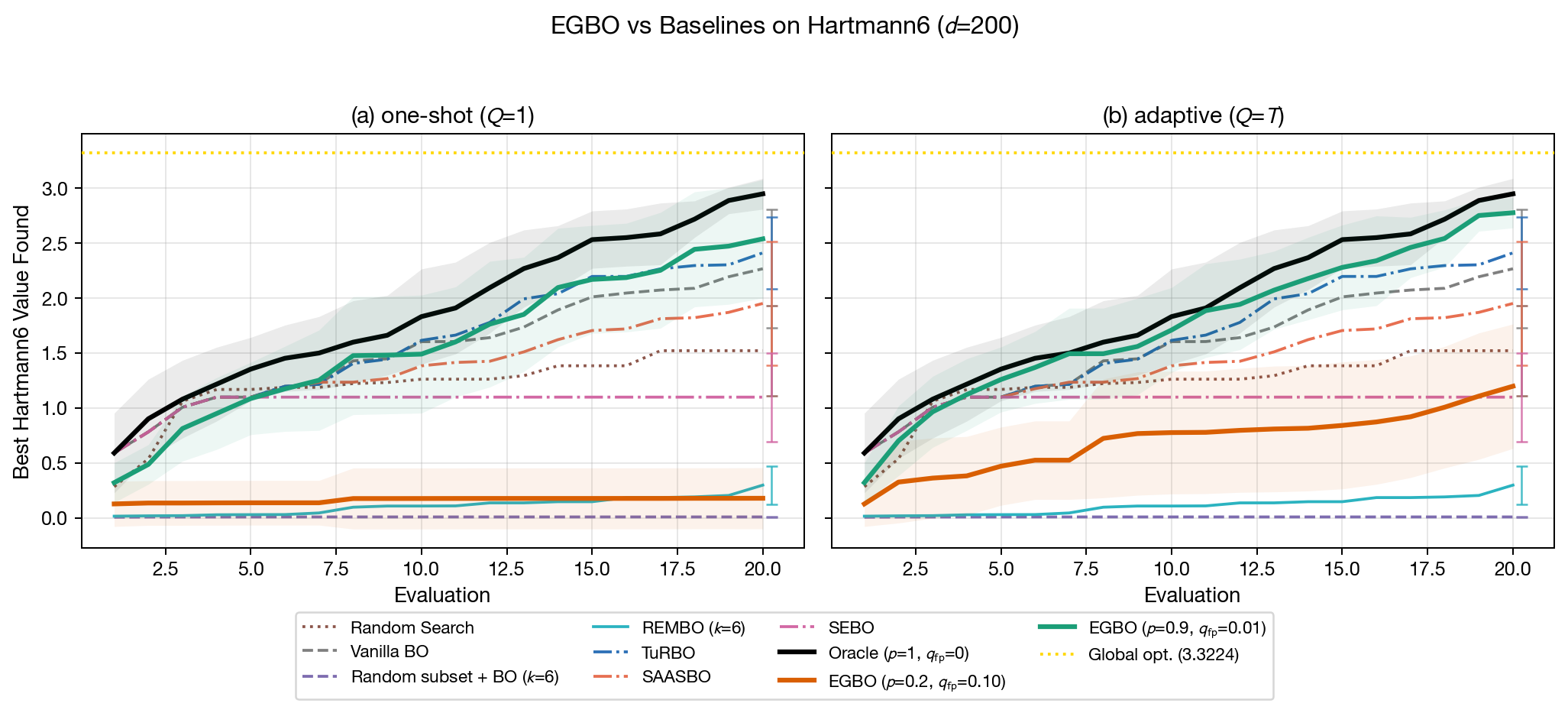}%
\caption{EGBO vs.\ baselines on Hartmann6 embedded in $N=200$ dimensions, with 95\% CIs. CIs for all methods are shown in Figure~\ref{fig:hartmann6_cis_for_all} of Appendix~\ref{app:hartmann6}.
\textbf{(a)} One-shot expert query ($Q=1$): the simulated expert proposes an active variable set once before BO begins, and BO runs in that fixed subspace for the remaining budget. \textbf{(b)} Fully adaptive ($Q=T$): the expert is queried before each of the $T=20$ evaluations, including the five initial Sobol points, with monotonically growing active set. The strong-expert variant ($p=0.9$, $q_{fp}=0.01$) closely tracks the oracle in both regimes; the weak-expert variant ($p=0.2$, $q_{fp}=0.10$) underperforms random search under one-shot querying but improves under adaptive querying.} 
\label{fig:synthetic_main}
\end{figure}

\paragraph{Methods.} To Hartmann6, we append 194 dummy coordinates that do
not affect the objective. The random subset + BO baseline is given the
oracle sparsity level $k=6$. We sweep two axes: (i) the number of expert query rounds $Q$, and
(ii) the expert quality parameters.%
\paragraph{Results.} As shown in Figure~\ref{fig:synthetic_main}, EGBO with a strong expert ($p=0.9$, $q_{fp}=0.01$) performs second best, after the oracle. EGBO with a weak expert ($p=0.2$, $q_{fp}=0.1$), in contrast, is outperformed by random search at $Q=1$, because with low coverage it spends its entire budget in a subspace without a good solution. The $(p, q_{fp})$ heatmap (Appendix~\ref{app:pq_heatmap}) maps this selection gap and shows that EGBO closes most of the gap to the oracle only when recall is high and false positives are rare. Adaptive querying helps the weak expert, reducing regret from $3.20$ at $Q=1$ to $2.68$ at $Q=16$, as Proposition~\ref{prop:adaptive_coverage} predicts. For the strong expert, whose one-shot selection already covers a near-optimal support, the effect of adaptivity is small and depends on when the expert is queried (Appendix~\ref{app:qsweep}). %

\subsection{\textsc{FormulateBench}}\label{sec:formulateBench}
\paragraph{Methods.}
To compare EGBO to baselines on sustainable protein formulation tasks, we construct \textsc{FormulateBench}, 24 tasks (14 meat, 10 dairy) in which the goal is to select plant-based ingredients and concentrations to match an animal product's nutritional profile. Specifically, the objective is z-score RMSE over eight nutrients, computed from USDA FoodData Central~\citep{usda2019fooddata}. All methods use the same filtered pools of 106 meat or 90 dairy ingredients drawn from NECTAR~\citep{nectar2025taste}; full construction details are in Appendix~\ref{app:exp-formulatebench}. The problem's mathematical structure---non-negative least squares (NNLS)---guarantees an optimum with at most \(m+1=9\) ingredients under the simplex constraint (Proposition~\ref{prop:nnls}).

We additionally report two analytic floors per category: $\mathrm{Oracle}_{\text{full}}$, the constrained NNLS optimum over all $d$ ingredients, and $\mathrm{Oracle}_{S_{\mathrm{LLM}}}$, the same restricted to the LLM's chosen subset. These give an empirical decomposition of the suboptimality bound, with $\mathrm{Oracle}_{\text{full}}\!\to\!\mathrm{Oracle}_{S_{\mathrm{LLM}}}$ the selection gap and $\mathrm{Oracle}_{S_{\mathrm{LLM}}}\!\to\!f(\hat x_T)$ the residual optimization gap.

\paragraph{Context provided to the LLM expert.}
In the LLM-guided and LLM-only settings, Claude Opus 4.7 is told only that it is optimizing the overall similarity of a plant-based formulation to the target animal product and is shown the available ingredient names. %
After each evaluation, the iterative LLM methods receive only a scalar target-dissimilarity score (lower is better). This abstracts a setting in which the relationship between formulation and perceived similarity, e.g. as judged by a sensory panel, is unknown.

EGBO and the two iterative LLM baselines share the LLM expert's first proposed recipe, which counts as evaluation 1. EGBO fixes the ingredients in this recipe and uses BO to optimize their mixture fractions over the simplex. Its initial design contains the LLM expert's recipe and four seeded local perturbations, after which BO begins at evaluation 6. The fixed-support iterative LLM is restricted to the same ingredients but directly proposes new concentrations from scalar feedback. The unrestricted iterative LLM may change both ingredients and concentrations. We call the better of the two in each category \emph{direct LLM optimization}. EGBO uses 10 BO seeds; each LLM baseline uses temperature-zero decoding and one cached trajectory per category.

\paragraph{BO configuration.}
We evaluate two one-shot EGBO variants that share the expert-selected ingredients, first recipe, initialization, and budget. \emph{EGBO (default BO)} uses BoTorch~0.18.1's native \texttt{SingleTaskGP} defaults (RBF kernel, learned homoskedastic noise) with qLogNEI. \emph{EGBO (LLM-configured BO)} uses one configuration chosen once by Claude Opus~4.7 from a prespecified menu, given only a high-level task description and the budget (Mat\'ern-$5/2$ ARD kernel, default lengthscale prior, learned noise, qLogNEI), fixed across all 24 categories. Our theory does not cover expert configuration of the BO subroutine, so we treat it as an empirical ablation.

\begin{figure}[t]
\includegraphics[width=\textwidth]{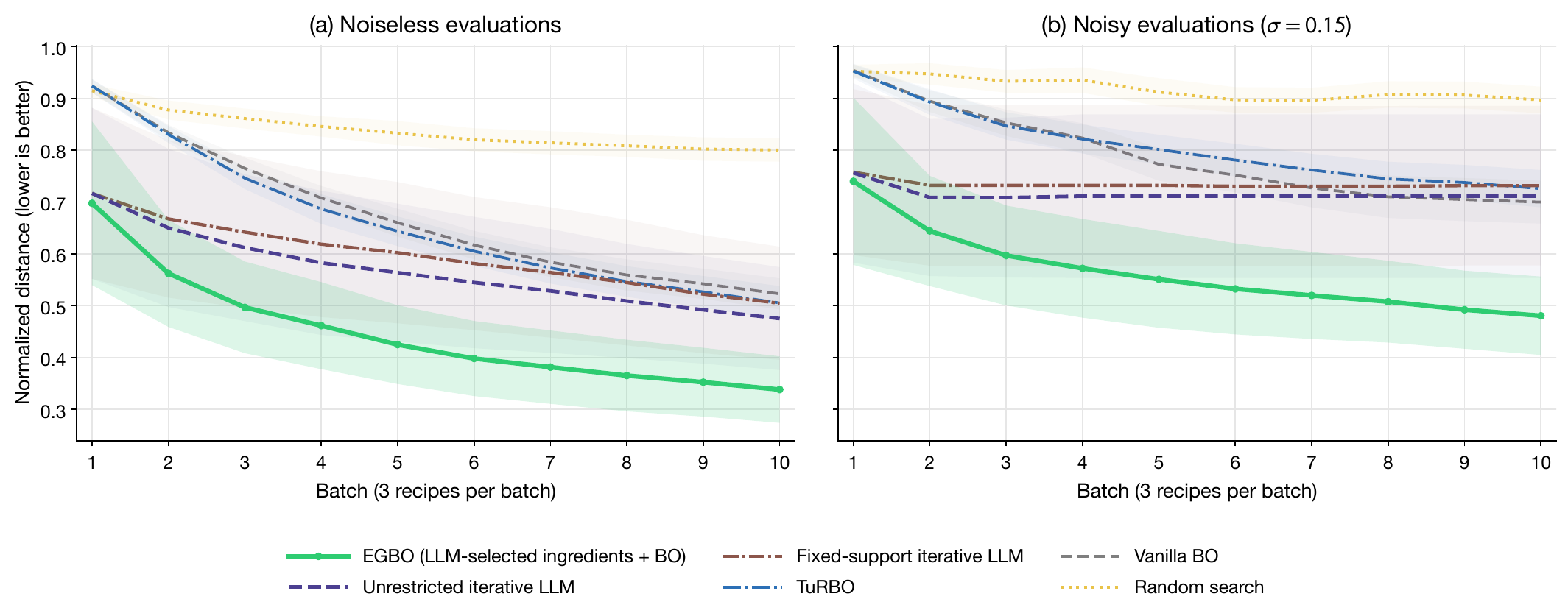}
\caption{\textsc{FormulateBench} in our target setting of 10 batches of 3 recipes with Claude Opus~4.7 as the LLM, (a) without and (b) with Gaussian observation noise. Curves show the mean over 24 categories of the true normalized distance of each method's recommended recipe (lower is better); the shaded region shows a 95\% confidence interval across categories for EGBO (lighter for other methods). EGBO and the conventional baselines average 10 seeds per category; each iterative LLM arm is one trajectory per category. Results with other LLMs are in Table~\ref{tab:batched}; one-recipe-at-a-time results with all baselines are in Appendix~\ref{app:formulatebench-sequential}, and at other noise levels in Appendix~\ref{app:noise}.}
\label{fig:fb-batched}
\end{figure}

\paragraph{Results in the target setting.}
Our target setting is that of our deployment (Section~\ref{sec:real}) and of much experimental science, in which candidates are evaluated in batches and measurements are noisy. We therefore evaluate 10 batches of 3 recipes with Gaussian observation noise of standard deviation $0.15$ times the category's normalizer, and report the noiseless case for comparison (Figure~\ref{fig:fb-batched}, Appendix~\ref{app:batched}). Each method is scored by the true distance of its recommended recipe, the one with the best observed score. With noise, both iterative LLM arms stop improving after the first batch, and telling the LLM the noise level does not help (Appendix~\ref{app:noise}). EGBO keeps improving and ends below direct LLM optimization ($0.48$ vs.\ $0.68$; 19/24 categories, $p<0.001$, Wilcoxon signed-rank). It also outperforms vanilla BO and TuRBO, on 21/24 and 22/24 categories, respectively. Without noise, EGBO also reaches the lowest distance after 30 evaluations, $0.34$, compared with $0.44$ for direct LLM optimization (19/24 categories, $p<0.001$) and $0.51$--$0.52$ for TuRBO and vanilla BO. With GPT-6.1 Sol or Gemini~4 Argon as the LLM (Table~\ref{tab:batched}), direct LLM optimization matches EGBO without noise (differences not significant). Under noise, it falls behind for GPT-6.1 Sol (18/24, $p<0.001$) but not significantly for Gemini~4 Argon (15/24, $p=0.12$).

\paragraph{One recipe at a time.}
With one evaluation per step and no noise (Appendix~\ref{app:formulatebench-sequential}), LLM-configured EGBO has the lowest mean distance at 10 and 20 evaluations among all methods, including SAASBO, SEBO, and REMBO. It outperforms the unrestricted iterative LLM on 20/24 categories at evaluation 20. By 20 evaluations, the two NNLS oracles show that most of the remaining error comes from the choice of ingredients rather than their concentrations. Yet letting the expert add ingredients midway (adaptive EGBO, $Q=2$) does not improve on one-shot EGBO on average (Appendix~\ref{app:formulatebench-adaptive}).

\paragraph{When does each strategy win?}
We compare direct LLM optimization, full-space BO (the better of vanilla BO and BO warm-started from the LLM's recipe), and EGBO per category (Appendix~\ref{app:regimes}). Direct LLM optimization is best on 14/24 categories at evaluation 5 and ties EGBO at evaluation 10 (11 each). EGBO is best on 18/24 at evaluation 20, and full-space BO is best on at most 2/24 after evaluation 5. EGBO's advantage over direct LLM optimization at evaluation 20 grows with the LLM's \emph{concentration error}, the gap between its first recipe and the best mixture of the same ingredients (Spearman $\rho=0.54$, $p=0.007$). Its advantage over full-space BO shrinks as the LLM's \emph{ingredient choice} worsens ($\rho=-0.51$, $p=0.011$). Together, these results suggest a division of labor: the LLM is most reliable at choosing what to vary, and BO at determining how much.

\paragraph{Do LLM agents delegate on their own?} The strategies above are fixed pipelines. To test whether an LLM agent would choose to delegate, we gave three LLM agents two tools in the noisy setting, one that evaluates a recipe and returns its score and one that executes Python code (Appendix~\ref{app:agents}). None ran a BO loop, even when told that BoTorch was installed, and 100 of 168 runs fit no model of their data at all. No agent significantly outperformed EGBO, and Gemini~4 Argon, which fit Gaussian processes most often, performed worst ($0.66$, vs.\ $0.55$ for EGBO).

\section{Real-World Deployment: Sustainable Dairy Formulation}\label{sec:real}

We deploy EGBO in two real-world sustainable protein applications in partnership with a food company, each in our target setting of batched experiments with noisy panel evaluations. In each, the goal is to move a plant-based product (yogurt or cottage cheese) toward a target sensory profile under a limited budget. Dairy products are particularly high-impact targets for plant-based reformulation due to the impact of the dairy industry on deforestation, climate change, land use, and animal welfare~\citep{pendrill2022disentangling,gerber2013tackling,ipbes2019global,vonkeyserlingk2013sustainability}. For yogurt, we conduct a controlled comparison against a professional human food scientist. The food company has developed a commercially available, LLM-based agentic framework, which we denote Food Scientist Agent (FSA) and use as the expert in EGBO.  To ensure reproducibility, we release FSA's output, the optimization variables, and all code. We set $Q=1$, as both FSA's strong initial recipes (utility 0.767, 0.776) and performance of LLM-guided EGBO on \textsc{FormulateBench} suggest one-shot selection is reasonable for this setting. We leave real-world evaluation of adaptive expansion to future work. %

\subsection{Methods.}
The full candidate ingredient pool is the food company's database ($N \approx 88{,}000$). Prior to optimization, FSA performed one-shot variable selection. Given the product category, the target dairy sensory profile, and the plant-based constraint, it selected a small set of active ingredients from the full pool. %
For yogurt, FSA selected $d = 7$ active ingredients. The sensory objective was defined over four attributes assessed on a $0$ to $15$ scale: Consistency (target $t_1 = 12$), Creaminess (target $t_2 = 5$), Tanginess (target $t_3 = 5$), and Uniformity (target $t_4 = 13$). The initial recipe had sensory scores $s^{(0)} = (5, 3, 10, 13)$, creating a nontrivial multi-objective problem in which one attribute already matches the target exactly while three require coordinated improvement. %
For cottage cheese, FSA selected $d = 9$ active ingredients. The sensory objective was defined over six attributes assessed on a $0$ to $15$ scale: Homogeneity (target $t_1 = 6$), Dispersion Phase (target $t_2 = 6$), Slipperiness (target $t_3 = 6$), Sourness (target $t_4 = 6.5$), Dairy (target $t_5 = 7$), and Off-Notes (target $t_6 = 0$). The initial recipe had sensory scores $s^{(0)} = (4, 8, 9, 10, 3, 7)$, presenting a more complex optimization landscape, since five of six attributes require substantial correction, with Off-Notes requiring near-complete elimination and Dairy requiring more than a doubling. In both cases, the total laboratory budget was $T = 10$ iterations of 3 recipes each (30 total recipes).

\paragraph{Scalar utility.}
In both campaigns, we aggregate the multi-objective sensory targets into a scalar utility score. For a candidate formulation $x$, let $s_j(x)$ denote the observed panel score for attribute $j$, $t_j$ its target, $m$ the number of sensory attributes, and $S_{\max}$ the maximum value of the sensory scale. With uniform weights $w_j = 1/m$, the scalar utility, which lies in $[0,1]$ with $1$ for an exact match, is
\begin{equation}\label{eq:scalar-utility}
f(x) = \frac{1}{m}\sum_{j=1}^{m}\left(1-\frac{|s_j(x)-t_j|}{S_{\max}}\right).
\end{equation}
At each iteration, EGBO proposed a batch of recipes within the active subspace, which were prepared under standardized protocols and scored by the sensory panel.

\subsection{Results}
\begin{figure}[t]
\centering
\begin{minipage}[c]{0.55\textwidth}
\centering
\small
\vspace{0.5em}
\renewcommand{\arraystretch}{1.15}          %
\begin{tabular}{@{}l
    S[table-format=2.3]                      %
    S[table-format=2.3]                      %
    S[table-format=2.3]                      %
    S[table-format=1.3]                      %
    S[table-format=1.3]@{}}                  %
\toprule
{Attribute} & {Target} & {Baseline} & {Best}
            & \multicolumn{2}{c@{}}{Error} \\
\cmidrule(l){5-6}
            &          &            &
            & {Before} & {After} \\
\midrule
{Consistency} & 12    & 5     & 12    & 7     & 0     \\
{Creaminess}  & 5     & 3     & 5     & 2     & 0     \\
{Tanginess}       & 5     & 10    & 5     & 5     & 0     \\
{Uniformity}  & 13    & 13    & 12.5  & 0     & 0.5   \\
\midrule
{Utility $f(x)$} & 1.000 & 0.767 & 0.992 & 0.233 & 0.008 \\
\bottomrule
\end{tabular}
\captionof{table}{%
Overall utility of the plant-based yogurt,
as assessed by the median scores of a trained panel of five participants,
improved 29\% relative to baseline. %
}
\label{tab:yogurt_results}
\end{minipage}%
\hfill
\begin{minipage}[c]{0.42\textwidth}
\centering
\includegraphics[width=0.75\linewidth]{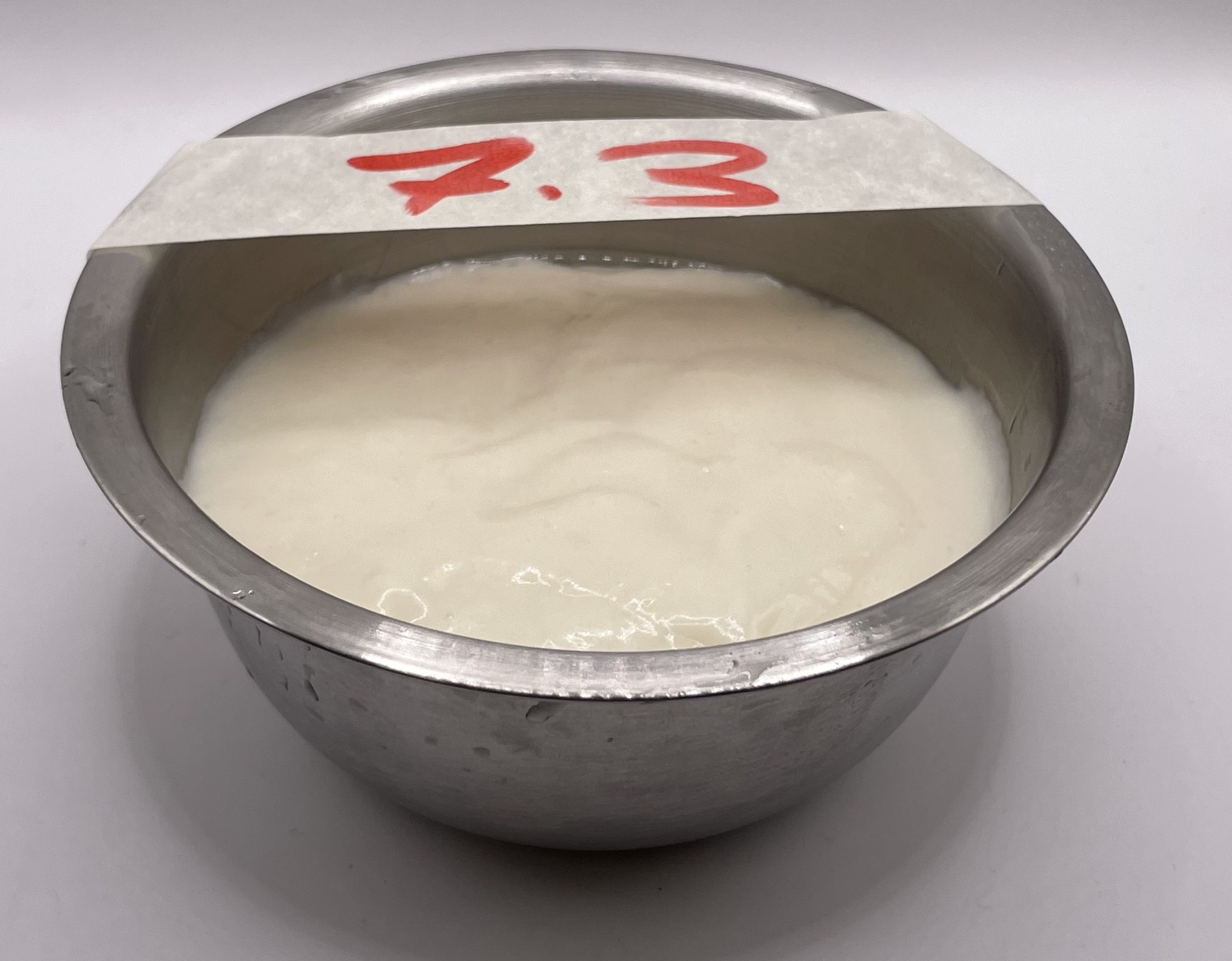}
\captionof{figure}{Best plant-based yogurt formulation
(iteration~7, recipe~3).}
\label{fig:yogurt_best}
\end{minipage}
\end{figure}

EGBO developed a plant-based yogurt formulation with a baseline utility of $f(x^{(0)}) = 0.767$ (recipe produced from FSA) and improved it to a best observed utility of $f(\hat{x}_T) = 0.992$ over 10 laboratory iterations (30 total recipes), a 29\% improvement. The mean absolute error across the four sensory attributes fell from 3.5 to 0.13 points on a 0--15 scale. The first, space-filling batch within the FSA-selected subspace already raised utility to 0.917; BO, beginning at iteration~3, raised it further to 0.992. %
For cottage cheese, EGBO improved a plant-based formulation from a baseline utility of $f(x^{(0)}) = 0.776$ to a best observed utility of $f(\hat{x}_T) = 0.975$ over 10 laboratory iterations (30 total recipes), a 26\% improvement.\footnote{In the cottage cheese campaign, the six attribute weights were entered as $w_j = 0.17$ (summing to $1.02$) rather than $1/6$, so all reported cottage cheese utilities are on that scale, with a maximum of $1.02$. Rescaling to $w_j = 1/6$ gives a baseline of $0.761$ and a best of $0.956$, the same 26\% improvement. Because the scaling is uniform, it does not affect the recipes proposed by BO.} Mean absolute attribute error fell from 3.6 to 0.67 points. Space-filling batches within the FSA-selected subspace raised utility to 0.850 by iteration~2, and BO raised it to 0.975. In both campaigns, the expert-selected subspace accounts for much of the gain and BO for the refinement, consistent with the decomposition in Equation~\eqref{eq:decomp}. Tables~\ref{tab:yogurt_results} and~\ref{tab:cottage_results} summarize the per-attribute results for yogurt and cottage cheese respectively.

\subsection{EGBO vs.\ Human Food Scientist}\label{sec:human-comparison}

We compare EGBO with a professional human food scientist (HFS) on the yogurt task. HFS had prior experience in recipe formulation but no prior experience specifically with plant-based dairy, to avoid the situation of HFS simply reproducing a suitable recipe from memory. This mirrors the setting of a company entering a new product category rather than reproducing a known recipe. HFS had unrestricted internet and literature access and reported consulting scientific and technical sources before beginning. HFS did not have access to EGBO's suggestions or results.
We report two comparisons under the same approximate laboratory time (40 hours). In the workflow comparison, HFS develops a yogurt from scratch and reaches utility 0.850, while EGBO (FSA selection plus BO) reaches 0.992. The same-start comparison isolates refinement. From FSA's recipe (0.767), HFS reaches 0.867 and BO reaches 0.992. It is the laboratory analogue of our \textsc{FormulateBench} comparison between direct and BO-based refinement, with a human expert in place of the LLM.

\paragraph{Experiment~1: full workflow comparison.}
HFS was asked to develop, from scratch, a plant-based yogurt recipe targeting the same four sensory objectives used in the EGBO campaign (Consistency~12, Creaminess~5, Tanginess~5, Uniformity~13). Despite having full freedom to perform batching, HFS chose a conventional iterative workflow, preparing a candidate formulation, letting it cool, tasting it, deciding on the next modification, and repeating. A total of 29 trials were executed over approximately 40~hours. Of these, 19 were classified as failed trials (the operator abandoned the attempt and restarted from scratch), and 10 produced formulations deemed sufficiently promising to evaluate for utility. Figure~\ref{fig:leg1_utility} shows the utility progression across the 10 evaluated trials. The best utility achieved was $0.850$ (trial~28), with a general upward trend as HFS progressively refined the plant base and texture. %

\paragraph{Experiment~2: same-start refinement comparison.}
To isolate the refinement step, HFS was given FSA's recipe ($f(x^{(0)}) = 0.767$) and asked to optimize it toward the same four sensory targets. HFS was allocated 10 iterations, matching the EGBO campaign. Similarly, HFS chose to produce one recipe per iteration, and EGBO proposed batches of three recipes per iteration. 
Each HFS iteration required approximately 4~hours of laboratory time, for a total of approximately 40~hours.

Figure~\ref{fig:leg2_comparison} shows the utility for both EGBO and HFS starting from the same baseline. EGBO reached a utility of $0.950$ by iteration~4 and peaked at $0.992$ by iteration~7. HFS improved more slowly, reaching $0.825$ at iteration~3 and $0.867$ at iteration~10. The gap was largest around iterations~6--7; by iteration~10, EGBO's best-so-far ($0.992$) exceeded HFS's best-so-far ($0.867$) by 14.4\%.

\paragraph{Time and utility comparison.}
Table~\ref{tab:human_comparison} summarizes the comparison. The results highlight two distinct advantages of the EGBO workflow. First, EGBO allows \emph{parallelism}. It proposes a batch of 3 recipes per iteration, all of which can be prepared and evaluated in a single laboratory session, whereas HFS's sequential reasoning requires one-at-a-time execution. Second, EGBO is more \emph{consistent}. HFS's trajectory shows regressions (iterations~4--5 and~8 in Experiment~2), reflecting the difficulty of mentally tracking coupled ingredient interactions across multiple sensory dimensions, whereas EGBO's GP surrogate maintains a global model, which may reduce such backsliding. %

\begin{table}[t]
\centering
\small
\setlength{\tabcolsep}{4pt}
\begin{tabular}{llcccc}
\toprule
Comparison & Method & Start & Best & Relative Gain & Lab time \\
\midrule
Experiment 1: Overall workflow & HFS from scratch & -- & 0.850 & -- & ${\sim}$40\,h \\
Experiment 1: Overall workflow & EGBO: FSA + BO & 0.767 & 0.992 & +17\% vs. HFS & ${\sim}$40\,h \\
\midrule
Experiment 2: Same-start refinement & HFS refinement & 0.767 & 0.867 & -- & ${\sim}$40\,h \\
Experiment 2: Same-start refinement & BO refinement & 0.767 & 0.992 & +14\% vs. HFS & ${\sim}$40\,h \\
\bottomrule
\end{tabular}
\vspace{6pt}
\caption{Human comparison on plant-based yogurt. We first compare the overall workflows under the same approximate laboratory time budget; the same-start comparison isolates the refinement step. EGBO's refinement includes two initial space-filling batches (Section~\ref{sec:real}).}
\label{tab:human_comparison}
\end{table}

\section{Discussion}\label{sec:discussion}
The current pace of development in sustainable proteins is incompatible with the urgency needed for climate change mitigation~\citep{zurek2022climate}. Self-driving and agentic laboratories offer a path to the throughput this challenge requires~\citep{abolhasani2023rise,fu2026agentic}, but their architecture---how frontier models, optimization algorithms, and experimental access divide the work---remains open. We study one interface in this architecture, when the model should delegate the choice of experiments to an optimizer, and find that EGBO outperforms a professional food scientist on plant-based yogurt formulation. %

\paragraph{When should an agent delegate to an optimizer?} Our results suggest a decision rule. Direct LLM optimization is preferable when the budget is very small, when evaluations are nearly noiseless and the LLM is a strong optimizer, or when the LLM's first proposal is already close to optimal. EGBO is preferable when the LLM can identify which variables matter but not their values, and when the budget is moderate or evaluations are noisy. When the LLM's choice of variables is poor, the agent should revise the design space (e.g., unrestricted LLM optimization or adaptive EGBO) or fall back to BO over the full space.  Since an agent rarely knows its regime in advance, a natural next step is to diagnose it from early evaluations (e.g., replicating a recipe to estimate noise) before delegating. Because the LLM agents we tested with code execution did not delegate on their own (Section~\ref{sec:formulateBench}), the interface between agent and optimizer must currently be designed into the system. Training agents to recognize when to delegate, e.g., by post-training on noisy closed-loop tasks, is a promising direction.

\paragraph{Limitations.}
We study one interface of an agentic laboratory, the choice of experiments, rather than its full architecture. \emph{Benchmark:} the main comparison uses Claude Opus~4.7, with replications across two other LLMs (Appendices~\ref{app:batched}--\ref{app:crossmodel}). These LLMs have no tools such as literature search or code execution (Appendix~\ref{app:agents} tests agents with code execution in a minimal tool-calling loop), and \textsc{FormulateBench}'s nutritional objective and synthetic Gaussian noise are proxies for sensory evaluation. Learned taste and texture surrogates~\citep{lee2023principal}, robotic execution, and multi-fidelity BO~\citep{pmlr-v206-mikkola23a,sabanza2025best} could further ease the evaluation bottleneck. \emph{Deployment:} each campaign is a single run, compared with one food scientist but not with direct LLM optimization. The best utilities of both EGBO and the food scientist are maxima over single panel evaluations, which, as in \textsc{FormulateBench} (Appendix~\ref{app:noise}), are biased upward, more so for EGBO's 30 recipes. Natural next steps are replicated re-scoring of top recipes, an iterative-LLM arm, and a subspace-sampling control. \emph{Theory:} Assumption~\ref{asm:expert}'s independent coordinate inclusion is used only for the closed-form coverage bound, whereas LLM errors may be correlated across ingredients. The expert-quality parameters \((p,q_{\mathrm{fp}})\) are defined relative to the unknown \(\mathcal{U}_\eta\), and the analysis does not cover LLM-configured BO or shrinking active sets.

\paragraph{Societal impact.} %
EGBO could be used to optimize formulations for cost or palatability at the expense of nutrition, so we focus on applications with environmental and public-health benefits. %
It could also reduce demand for food scientists. In our framework, however, humans remain responsible for objective specification and quality oversight, while EGBO automates iterative refinement, not scientific judgment.
    \begin{ack}
    Thank you to Theodore John Leonsis, the Bezos Earth Fund, Food System Innovations, and Stanford HAI for supporting this work.
    \end{ack}

    \bibliographystyle{plainnat} %

\bibliography{references}

\newpage

\appendix

\section{Symbols and Abbreviations}
Table~\ref{tab:symbols} contains a list of symbols and abbreviations introduced throughout this paper.
\begin{table*}[h!]
\centering
\small
\begin{tabular}{@{}ll@{}}
\toprule
\textbf{Symbol} & \textbf{Used For} \\
\midrule
\multicolumn{2}{@{}l}{\textit{Acronyms}} \\
EGBO & Expert-Guided Bayesian Optimization \\
BO & Bayesian optimization \\
GP & Gaussian process \\
NNLS & nonnegative least squares \\
LLM & large language model \\
\midrule
\multicolumn{2}{@{}l}{\textit{Problem parameters}} \\
$N$ & ambient input dimension \\
$\mathcal{X}$ & input domain, $\mathcal{X} \subseteq \mathbb{R}^N$ \\
$f$ & black-box objective, $f : \mathcal{X} \to \mathbb{R}$ \\
$x^*$ & global maximizer of $f$ on $\mathcal{X}$ \\
$\|x\|_0$ & number of nonzero (active) coordinates of $x$ \\
$u_i^{\max}$ & upper bound of the $i$-th coordinate \\
\midrule
\multicolumn{2}{@{}l}{\textit{Sparsity structure}} \\
$k$ & target sparsity level (max size of any sparse near-optimal support) \\
$\eta$ & sparsity gap: $f(\tilde{x}) \geq f(x^*) - \eta$ for some $k$-sparse $\tilde{x}$ \\
$\tilde{x}^{(R)}$ & witness for support $R$: a $k$-sparse near-optimum supported in $R$ \\
$R$ & a sparse near-optimal support, $R \subseteq [N]$, $|R| \leq k$ \\
$\mathcal{R}_\eta$ & family of all $k$-sparse $\eta$-near-optimal supports \\
$M$ & multiplicity, $M = |\mathcal{R}_\eta|$ \\
$M'$ & number of pairwise disjoint supports in $\mathcal{R}_\eta$ used in bounds \\
$\mathcal{U}_\eta$ & support union, $\bigcup_{R \in \mathcal{R}_\eta} R$ \\
$\kappa$ & size of the support union, $\kappa = |\mathcal{U}_\eta|$ \\
\midrule
\multicolumn{2}{@{}l}{\textit{Expert quality}} \\
$E$ & domain expert (human, LLM, or other knowledge source) \\
$p$ & per-variable expert recall on $\mathcal{U}_\eta$ \\
$q_{fp}$ & per-variable expert false-positive rate outside $\mathcal{U}_\eta$ \\
\midrule
\multicolumn{2}{@{}l}{\textit{Algorithm and active set}} \\
$T$ & total BO budget (function evaluations) \\
$Q$ & number of expert query rounds \\
$\tau$ & BO iterations per round, $\tau = \lfloor T / Q \rfloor$ \\
$S$, $S_q$ & active variable set (at round $q$) \\
$d$, $d_Q$ & active set size, $d = |S|$ (resp.\ $d_Q = |S_Q|$) \\
$\mathcal{X}_S$ & restricted domain $\{x \in \mathcal{X} : x_i = 0 \text{ for } i \notin S\}$ \\
\midrule
\multicolumn{2}{@{}l}{\textit{Convergence quantities}} \\
$\hat{x}_T$ & best point observed by EGBO after $T$ evaluations \\
$f_S^*$ & best achievable value in $\mathcal{X}_S$, $\max_{x \in \mathcal{X}_S} f(x)$ \\
$\Delta_S$ & selection gap, $f(x^*) - f_S^*$ \\
$r_T$ & optimization gap, $f_S^* - f(\hat{x}_T)$ \\
$\varepsilon_{\mathrm{BO}}(T, d)$ & BO subroutine convergence rate as a function of budget $T$ and dimension $d$ \\
$\delta_{\mathrm{BO}}$ & failure probability of the BO subroutine \\
\midrule
\multicolumn{2}{@{}l}{\textit{NNLS objective (Section~\ref{sec:formulateBench})}} \\
$V$ & ingredient-to-nutrient matrix, $V \in \mathbb{R}_{\geq 0}^{m \times N}$ \\
$v^*$ & target nutrient profile, $v^* \in \mathbb{R}^m$ \\
$m$ & number of nutrient targets \\
\bottomrule
\end{tabular}
\caption{Symbols and abbreviations used in this paper.}
\label{tab:symbols}
\end{table*}

\section{Extended Related Work}\label{app:extended-related-work}

\paragraph{LLMs in optimization and experimental design.} A growing body of work uses LLMs as components of optimization or experimental design loops. LLAMBO \citep{liu2024large} uses LLMs to inform surrogate modeling and candidate generation for BO, while OPRO \citep{yang2024large} treats the LLM itself as the optimizer over natural-language solution descriptions. Closer to our setting, GPTuner \citep{lao2023gptuner} uses LLMs to select database configuration knobs and propose search ranges before optimization.
GPTuner can be viewed as a special case of EGBO with $Q=1$ and a discrete database-knob domain. EGBO formalizes this class, extends it to continuous high-dimensional formulation problems with adaptive expansion, characterizes its regime of applicability, and deploys it for sustainable protein formulation.

\paragraph{LLMs versus BO, and hybrids.} Head-to-head comparisons of LLM and BO optimizers report results that depend on the setting. LLMs match or beat BO for hyperparameter optimization under small budgets~\citep{zhang2023using}, and LLM-based warm-starting helps most when observations are scarce~\citep{liu2024large}. In contrast, in genetic-perturbation and molecular design, LLM agents were insensitive to experimental feedback---replacing outcomes with permuted labels did not change their performance---and were outperformed by linear bandits and GP optimization~\citep{gupta2025llms}. LLM features improve BO over molecules only when the model is pretrained or fine-tuned on domain data~\citep{kristiadi2024sober}. Hybrid systems divide the work in different ways: an LLM intervenes when BO stagnates~\citep{cisse2025bora}, LLM proposals are screened by a statistical surrogate~\citep{chang2025llinbo}, an LLM partitions a chemical search space for BO~\citep{han2025chembomas}, or an LLM agent steers the search while delegating modeling and acquisition to a BO backend~\citep{brunzema2026agentic}. Data-driven alternatives to expert variable selection select active variables from evaluations~\citep{song2022monte,shen2023computationally}, at the cost of the budget spent identifying them. Our work differs in four ways. It compares direct LLM optimization, BO, and expert-guided BO on the same tasks. It characterizes when each is preferable as a function of budget, observation noise, batching, and the LLM. It isolates the LLM's contribution to a single interpretable decision, the choice of subspace, with accompanying theory. Finally, it deploys the result in a laboratory with a human sensory panel.

\paragraph{ML-driven formulation optimization.} Within food and formulated-product development, \citet{cao2021optimization} coupled Thompson Sampling Efficient Multi-objective Optimization (TSEMO) with a naive Bayes stability classifier to optimize a five-ingredient liquid formulation, finding nine acceptable recipes in 15 working days. \citet{becker2023optimization} applied TSEMO to a three-ingredient whey protein formulation using a fully automated milli-fluidic platform. Neither approach exploits expert priors to reduce the search dimension. \citet{thomas2025what} found that LLMs can exceed the performance of human food scientists in experimental design for plant-based meats, as judged by other expert food scientists.

\section{Theory and Proofs}\label{app:proofs}

Throughout this appendix, we use the independence between the expert's variable-selection randomness and the BO subroutine's randomness assumed in Section~\ref{sec:egbo}.

\subsection{Suboptimality Decomposition}
\begin{proof}[Proof of Equation~\eqref{eq:decomp}]
By definition,
\[ f_S^\star = \max_{x \in \mathcal{X}_S} f(x). \]
Adding and subtracting \(f_S^\star\) gives
\[ f(x^*) - f(\hat{x}_T) = \bigl(f(x^*) - f_S^\star\bigr) + \bigl(f_S^\star - f(\hat{x}_T)\bigr). \]
The first term is the selection gap and the second term is the optimization gap.

Now suppose \(R \subseteq S\) for some \(R \in \mathcal{R}_\eta\). By Assumption~\ref{asm:sparsity}, there exists a witness \(\tilde{x}^{(R)}\) with
\[ \mathrm{supp}(\tilde{x}^{(R)}) \subseteq R \qquad\text{and}\qquad f(\tilde{x}^{(R)}) \geq f(x^*)-\eta. \]
Since \(R \subseteq S\), we have \(\tilde{x}^{(R)} \in \mathcal{X}_S\). Therefore,
\[ f_S^\star = \max_{x \in \mathcal{X}_S} f(x) \geq f(\tilde{x}^{(R)}) \geq f(x^*)-\eta. \]
Thus \( f(x^*) - f_S^\star \leq \eta. \)
\end{proof}

\subsection{One-Shot Coverage (Theorem~\ref{thm:oneshot})}
\begin{proof}
Fix a witness support \(R \in \mathcal{R}_\eta\). In the one-shot setting, each variable \(i \in R\) is included in the active set \(S\) independently with probability \(p\). Hence \( \Pr(R \subseteq S) = p^{|R|}. \) Since \(|R| \leq k\) and \(p \in (0,1]\), \( p^{|R|} \geq p^k. \)

On the event \(R \subseteq S\), the subspace \(\mathcal{X}_S\) contains the witness \(\tilde{x}^{(R)}\). By the decomposition above, the selection gap is therefore at most \(\eta\): \( f(x^*) - f_S^\star \leq \eta. \)

By Assumption~\ref{asm:bo}, when BO is run on \(\mathcal{X}_S\) for \(T\) evaluations, it returns \(\hat{x}_T\) satisfying \( f_S^\star - f(\hat{x}_T) \leq \varepsilon_{\mathrm{BO}}(T, |S|) \) with probability at least \(1-\delta_{\mathrm{BO}}\).

Combining the selection and optimization gaps gives
\[ f(x^*) - f(\hat{x}_T) \leq \eta + \varepsilon_{\mathrm{BO}}(T, |S|). \]
The coverage event \(R \subseteq S\) and the BO success event are independent by assumption, so the joint event occurs with probability at least \( p^{|R|}(1-\delta_{\mathrm{BO}}) \geq p^k(1-\delta_{\mathrm{BO}}). \)

It remains to compute the expected active-set size. Under Assumption~\ref{asm:expert},
\[ \mathbb{E}[|S|] = \kappa p + (N-\kappa)q_{\mathrm{fp}}. \]
\end{proof}

\subsection{Coverage of Multiple Disjoint Supports}
Theorem~\ref{thm:oneshot} fixes one sparse near-optimal support \(R\). If there are many such supports, the probability that the expert covers at least one of them can be larger.

\begin{lemma}[Coverage of multiple disjoint supports]\label{lem:multi}
Let \(\mathcal{R}' \subseteq \mathcal{R}_\eta\) be a collection of pairwise disjoint supports. Suppose each variable in \(\mathcal{U}_\eta\) is included independently with probability \(\pi\). Then
\[ \Pr(\exists R \in \mathcal{R}' : R \subseteq S) = 1 - \prod_{R \in \mathcal{R}'} \left(1-\pi^{|R|}\right). \]
If additionally \(|R| \leq k\) for every \(R \in \mathcal{R}'\), then
\[ \Pr(\exists R \in \mathcal{R}' : R \subseteq S) \geq 1 - (1-\pi^k)^{|\mathcal{R}'|}. \]
\end{lemma}
\begin{proof}
For a fixed support \(R\), \( \Pr(R \subseteq S) = \pi^{|R|}. \) Because the supports in \(\mathcal{R}'\) are pairwise disjoint, the events \(\{R \subseteq S\}\) depend on disjoint sets of independent Bernoulli variables, so they are independent. Hence
\[ \Pr(\text{no support in } \mathcal{R}' \text{ is covered}) = \prod_{R \in \mathcal{R}'} \left(1-\pi^{|R|}\right). \]
If \(|R| \leq k\) and \(\pi \in [0,1]\), \( \pi^{|R|} \geq \pi^k \), so \( 1-\pi^{|R|} \leq 1-\pi^k \) and the bound follows.
\end{proof}

In the one-shot setting, \(\pi=p\). Therefore, if \(\mathcal{R}_\eta\) contains \(M'\) pairwise disjoint supports of size at most \(k\), then
\[ \Pr(\exists R \in \mathcal{R}_\eta : R \subseteq S) \geq 1 - (1-p^k)^{M'}. \]

\subsection{Adaptive Coverage (Proposition~\ref{prop:adaptive_coverage})}
\begin{proof}
Consider any variable \(i \in \mathcal{U}_\eta\). In each expert-query round, if \(i\) has not already been included, it is added with probability \(p\). The probability that \(i\) is not added in any of the \(Q\) rounds is \( (1-p)^Q, \) so \( \pi_Q = 1-(1-p)^Q. \)

Fix \(R \in \mathcal{R}_\eta\). By Assumption~\ref{asm:expert}, inclusion events are independent across coordinates, so \( \Pr(R \subseteq S_Q) = \pi_Q^{|R|}. \)

A simple union-bound gives \( \Pr(R \not\subseteq S_Q) \leq |R|(1-p)^Q \leq k(1-p)^Q, \) hence \( \Pr(R \subseteq S_Q) \geq \max\{0,\,1-k(1-p)^Q\}. \)

Finally, by linearity of expectation,
\[ \mathbb{E}[|S_Q|] = \kappa\bigl(1-(1-p)^Q\bigr) + (N-\kappa)\bigl(1-(1-q_{\mathrm{fp}})^Q\bigr). \]
\end{proof}

\section{Shared Experimental Details}\label{app:exp-details}

This appendix documents resources, GP/acquisition configuration, and LLM prompts shared by the synthetic Hartmann6 study (Appendix~\ref{app:hartmann6}), \textsc{FormulateBench} (Appendix~\ref{app:formulatebench}), and the real-world dairy deployment (Appendix~\ref{app:deploy}). Exact library versions are pinned in each release directory's
\texttt{requirements.txt}.

\subsection{Compute resources}\label{app:compute}

All experiments in this paper run on CPU; no GPU is required at any point.
We executed runs on a 14-inch MacBook Pro with an Apple M5 Pro chip
(18-core CPU, 20-core GPU, 16-core Neural Engine), 48 GB unified memory,
and OS Tahoe 26.4.1 with the library versions pinned in each release directory's
\texttt{requirements.txt}. Per-experiment
compute requirements are summarized in Table~\ref{tab:compute} and detailed below.

\begin{table}[h]
\centering
\small
\caption{Compute resources per experimental setting. Lab time refers to
physical-laboratory wall-clock time (recipe preparation, cooling,
sensory evaluation), which dominates the deployment budget.}
\label{tab:compute}
\begin{tabular}{@{}lllll@{}}
\toprule
Setting & Hardware & Per-run wall time & Total CPU time & Lab time \\
\midrule
Hartmann6 ($N=200$) & 1 CPU core      & 5--10\,min (SAAS); ${<}1$\,min (others) & ${\sim}50$\,CPU-h & --- \\
\textsc{FormulateBench}     & 1 CPU core      & 2--5\,min (SAAS); ${<}1$\,min (others) & ${\sim}10^{2}$\,CPU-h & --- \\
Yogurt deployment   & 1 CPU core      & ${<}1$\,min per acquisition step       & ${\sim}10$\,CPU-min & ${\sim}40$\,h \\
Cottage cheese deployment & 1 CPU core & ${<}1$\,min per acquisition step       & ${\sim}10$\,CPU-min & ${\sim}40$\,h \\
\textsc{FormulateBench} (batched, other LLMs, agents) & 1 CPU core & ${<}1$\,min per run (BO); LLM-bound & ${\sim}25$\,CPU-h & --- \\
\bottomrule
\end{tabular}
\end{table}

\paragraph{Hartmann6 ($N=200$).} Each main comparison (one-shot or
adaptive) covers 10 methods (including both EGBO presets) $\times$
10 seeds $\times$ 20 evaluations $=$ 2{,}000 function evaluations, and
each heatmap cell adds $10 \times 20 = 200$ EGBO evaluations. Per-seed wall time is dominated by NUTS sampling in the
SAAS-based methods (SAASBO, SEBO) at roughly $5$--$10$ minutes. All other
methods complete in under one minute per seed. Peak resident memory stays
below 4\,GB. The full Hartmann6 sweep (one-shot + adaptive,
all expert regimes) is approximately $50$ CPU-hours on the machine above.

\paragraph{\textsc{FormulateBench}.} The main comparison evaluates nine stochastic optimization methods over 10 seeds and two temperature-zero iterative LLM baselines using one cached trajectory per category, for
\[
24 \times (9 \times 10 + 2) \times 20 = 44{,}160
\]
objective evaluations. The adaptive EGBO ablation adds \(24 \times 10 \times 20 = 4{,}800\) evaluations.

\paragraph{Additional \textsc{FormulateBench} experiments.} The batched experiments (Appendix~\ref{app:batched}) comprise $3$ LLMs $\times$ $2$ noise levels $\times$ $24$ categories of iterative LLM arms (one trajectory each) and EGBO, plus random search, vanilla BO, and TuRBO (10 seeds each), at 30 evaluations per run; the comparison with other LLMs, the noise sweeps with other LLMs, and the noise-aware arm follow the one-at-a-time protocol. Together these use about 25 CPU-hours and about 9{,}000 LLM API calls. The tool-using-agent test (Appendix~\ref{app:agents}) used about 4{,}800 LLM API calls; its sandboxed code execution is negligible. All LLM responses and agent transcripts are released.

\paragraph{Real-world deployment.} The computational footprint of each EGBO iteration is negligible. Acquisition-function optimization with
\texttt{num\_restarts}~$=20$, \texttt{raw\_samples}~$=1024$,
\texttt{sample\_shape}~$=[512]$, \texttt{sequential}~$=$~True completes
in under one minute on a single CPU core, and per-campaign cumulative
acquisition compute is on the order of $10$ minutes. The binding constraint is laboratory wall time, since each batch of $q=3$ recipes
requires approximately $4$ hours of physical preparation, cooling, and
trained-panel sensory evaluation, giving a total of $\sim 40$ hours of
lab time per 10-iteration campaign. Both the yogurt and cottage cheese
campaigns ran on the same compute and panel infrastructure.

\paragraph{Total compute footprint.} Summing across all reported
experiments, the full reproduction of every numeric result and figure
in this paper requires approximately $\mathbf{1.8\times 10^{2}}$ CPU-hours
of single-machine compute and about $2 \times 10^{4}$ LLM API calls (all responses are released). The experiments also used $\mathbf{\sim 160}$ hours of physical
laboratory time (about 80 hours for the two EGBO campaigns and 80 hours for the two HFS
experiments).

\subsection{GP and acquisition configuration}\label{app:exp-shared}

Unless noted otherwise, GP hyperparameters are fit by maximum marginal likelihood using \texttt{fit\_gpytorch\_mll} on \texttt{ExactMarginalLogLikelihood}, with \texttt{Standardize}($m=1$) as the outcome transform. Acquisition functions are optimized with \texttt{optimize\_acqf} using multiple L-BFGS-B restarts, and inputs are normalized to $[0,1]^d$ before being passed to the GP. The synthetic study and conventional \textsc{FormulateBench} baselines use their method-specific configurations documented with the released runners. The native-default EGBO control uses BoTorch~0.18.1's default \texttt{SingleTaskGP}, whose covariance is an RBF kernel, together with learned homoskedastic noise and qLogNEI. The LLM-configured EGBO arm uses a Mat\'ern-$5/2$ ARD kernel, default lengthscale prior, learned default noise, and qLogNEI. Both use a \texttt{SobolQMCNormalSampler} with \texttt{sample\_shape}~$=[256]$, \texttt{num\_restarts}~$=10$, and \texttt{raw\_samples}~$=512$. The deployment uses \texttt{sample\_shape}~$=[512]$ because it operates with batches of $q=3$. The fully Bayesian baselines (SAASBO and SEBO) use \texttt{SaasFullyBayesianSingleTaskGP} with NUTS (warmup steps $=512$, post-warmup samples $=256$); on NUTS-fit failure we fall back to \texttt{SingleTaskGP}.

\subsection{LLM expert: prompts and caching}
\label{app:llm-prompts}

The main \textsc{FormulateBench} comparison uses Claude Opus~4.7
(\texttt{anthropic/claude-opus-4.7} via OpenRouter) at temperature
\(0\) to generate formulations under the limited-information regime used in the main experiment. The LLM is told that it is optimizing the overall similarity
of a plant-based formulation to a target animal product and is shown
only the available ingredient names. It is not shown the ingredient
nutrient profiles, target nutrient vector, or mathematical form of the
objective.

For each category, the first LLM response specifies both an ingredient
set and initial concentrations. This recipe is shared by EGBO and the
two iterative LLM baselines and counts as the first objective
evaluation. EGBO fixes its ingredients and uses BO to optimize their
concentrations. The fixed-support iterative LLM is restricted to the
same ingredients but directly proposes new concentrations, whereas the
unrestricted iterative LLM may change both ingredients and
concentrations. After each evaluation, the iterative LLM methods
receive the previously evaluated formulations and their scalar
target-dissimilarity scores, with lower values indicating better performance. They receive no additional information about the hidden
nutrition objective.

All formulation-expert calls use the system message ``You are an expert
plant-based food scientist.'' The response is parsed as a JSON object
containing one complete recipe with at most \(k=9\) ingredients and
fractions summing to one. The cap \(k=9\) follows from the NNLS
sparsity bound (Proposition~\ref{prop:nnls},
Appendix~\ref{app:formulatebench}). The initial user prompt is
reproduced in Figure~\ref{fig:llm-prompt-formulateBench}.

For reproducibility without API access, each OpenRouter response is
cached on disk using a SHA-256 hash of the model, system prompt, user
prompt, and decoding settings. We use one cached temperature-zero trajectory per iterative LLM method and category, and include the cached trajectories and per-category result JSONs in the code release.

\paragraph{BO-configuration prompt.}
The BO-configuration call is also made once at temperature zero and cached. It states that an expert has proposed a recipe of at most nine ingredients, that BO will refine mixture fractions over the simplex, that the total budget is 20 evaluations, and that only scalar target-dissimilarity feedback will be observed. From a fixed menu of kernel, lengthscale-prior, noise, acquisition, and ARD choices, Claude selected a Mat\'ern-$5/2$ kernel, default lengthscale prior, default/learned noise, qLogNEI, and ARD. This configuration is used unchanged for every category.

The Food Scientist Agent (FSA) used in the real-world deployment
(Section~\ref{sec:real}) is a commercially available agentic system. We use its recipes as one-shot inputs and do not invoke it
during BO.

\begin{figure}[h!]
\begin{tcolorbox}[
    colback=lightyellow,
    colframe=yellow,
    coltitle=black,
    title=User prompt
]
You are optimizing the overall similarity of a plant-based formulation
to animal-based $<$\textit{category}$>$. The evaluator is a black
box.\\[4pt]

Available plant-based ingredients:\\
$<$\textit{ingredient names}$>$\\[4pt]

No formulations have been evaluated yet. Propose the strongest initial
formulation using your prior knowledge.\\[4pt]

Propose exactly one complete formulation. Use at most 9 available
ingredients, nonnegative quantities, and no ingredient outside the
catalog. Return only JSON in this exact shape:\\
\texttt{\{"recipe":[\{"ingredient":"exact catalog name",}\\
\texttt{\hspace*{1em}"fraction":0.25\}]\}}\\
Fractions must sum to 1. Do not include commentary.
\end{tcolorbox}
\caption{Prompt used to obtain the shared initial recipe for the
LLM-informed methods in \textsc{FormulateBench}. The system message is
``You are an expert plant-based food scientist.'' EGBO fixes the
ingredients in this recipe and optimizes their concentrations. At
subsequent evaluations, the iterative LLM baselines instead receive
the previously evaluated formulations and their scalar target-dissimilarity
scores. The cap \(k=9\) is derived from
Proposition~\ref{prop:nnls}.}
\label{fig:llm-prompt-formulateBench}
\end{figure}

\section{Synthetic Benchmark: Hartmann6 ($N=200$)}\label{app:hartmann6}

\subsection{Setup}\label{app:exp-hartmann}

\paragraph{Problem.} We embed the standard 6-D Hartmann function into $N=200$ ambient dimensions. The embedded objective is $f(x) = \mathrm{Hartmann6}(x_{1:6})$, so coordinates $7,\ldots,200$ are inert and never enter the function. The search domain is $[0,1]^{200}$. For \emph{subspace methods} (oracle, random subset, REMBO, EGBO), BO optimizes only over the active set $S$ and inactive coordinates are pinned at $0$ (matching the support convention $\mathrm{supp}(x)\subseteq S$ used in Assumption~\ref{asm:sparsity}). Full-dimensional methods (vanilla BO, TuRBO, SAASBO, SEBO) optimize over all $200$ coordinates. Because $f$ ignores dimensions $7$--$200$, no pinning is required, but TuRBO initializes its trust-region center at $\mathbf{0.5} \in [0,1]^{200}$ following BoTorch defaults.

\paragraph{Budget.} Total budget $T=20$, of which $n_{\mathrm{init}}=5$ are Sobol points (BoTorch \texttt{SobolEngine}, scrambled, with the $i$-th Sobol draw seeded as \texttt{seed}~$+~i$ so that EGBO and Oracle share init points across active sets) and the remaining $15$ are acquisition steps with batch size $q=1$. We run $10$ random seeds per (method, regime). Shaded bands are $95\%$ confidence intervals from the standard error of the mean.

\paragraph{Acquisition optimization.} qLogNEI is optimized with \texttt{num\_restarts}~$=10$, \texttt{raw\_samples}~$=512$.

\paragraph{Simulated expert.} The expert is implemented in \texttt{expert\_step()} as independent per-coordinate Bernoulli draws, in which each of the $6$ relevant coordinates is added to the active set with probability $p$, and each of the $194$ irrelevant coordinates with probability $q_{fp}$. The convergence-curve experiments report two regimes as representatives of the ``EGBO wins'' and ``EGBO loses'' regions: a strong expert $(p, q_{fp}) = (0.9, 0.01)$ and a weak expert $(p, q_{fp}) = (0.2, 0.10)$, plus the oracle $(p, q_{fp}) = (1, 0)$. In adaptive runs the active set grows monotonically across rounds.

\paragraph{Baselines.}
\begin{itemize}
\item \textbf{Random search:} uniform $[0,1]^{200}$.
\item \textbf{Vanilla BO:} \texttt{SingleTaskGP} + qLogNEI on the full $200$-D box.
\item \textbf{Random subset + BO:} fixed random $k=6$ subset (seed offset $+10\,000$); BO is run inside it with the rest pinned at $0$. We intentionally give this baseline the oracle sparsity~$k$ so that the comparison isolates the value of \emph{which} variables are chosen, not \emph{how many}.
\item \textbf{Oracle:} BO restricted to the true relevant $6$-D subspace.
\item \textbf{REMBO}~\citep{wang2016bayesian}: Gaussian random projection $A \in \mathbb{R}^{200 \times 6}$ with $A_{ij} \sim \mathcal{N}(0, 1/\sqrt{6})$. Latent bounds are computed from $4096$ Sobol probes back-projected through $A$ rather than the canonical $[-\sqrt{6}, \sqrt{6}]$ box. This gave consistently better REMBO performance in our setting and we report results under the favorable bounds.
\item \textbf{TuRBO}~\citep{eriksson2019scalable}: trust-region length $L_0 = 0.8$, $L_{\min} = 2^{-7}$, $L_{\max} = 1.6$, success tolerance $3$, failure tolerance $\max(4, d)$ following the original paper's dimension-dependent rule (with batch size $q = 1$).
\item \textbf{SAASBO}~\citep{eriksson2021high}: \texttt{SaasFullyBayesianSingleTaskGP} with NUTS (warmup $512$, samples $256$), with the fallback noted in Appendix~\ref{app:exp-shared}.
\item \textbf{SEBO}~\citep{liu2023sparse}: SAAS GP with smooth-$\ell_0$ penalty, $\lambda_{\max} = 1.0$, smoothing $a = 0.05$, linear homotopy $\lambda_t = \lambda_{\max}\, t / T$, penalty centered at $x_{\mathrm{ref}} = 0$. These values follow the BoTorch reference implementation rather than the original SEBO settings. We adopted the BoTorch values because they were robust across our $N = 200$ regime without per-task tuning.
\end{itemize}

\paragraph{Run counts and compute.} With $10$ seeds, $T = 20$ evaluations, and $10$ methods (including both EGBO presets), each main Hartmann6 comparison is $10 \times 10 \times 20 = 2{,}000$ function evaluations, and each heatmap cell adds $10 \times 20 = 200$ EGBO evaluations. Full per-seed traces are written to \texttt{results/<NAME>/\{oneshot,adaptive\}/raw.json}. Runs were executed on a single CPU node; per-seed wall time is dominated by NUTS in SAASBO/SEBO at roughly $5$--$10$ minutes.

\paragraph{Entry points.} \texttt{run\_hartmann6.py} (per-method runs); \texttt{run\_hartmann6.sh} (fan-out across seeds); \texttt{merge\_hartmann6.py} and \texttt{plot\_hartmann6.py} (aggregation and figures).

\subsection{Best-so-far convergence (95\% CIs)}
\begin{figure}[h]
\centering
\includegraphics[width=\textwidth]{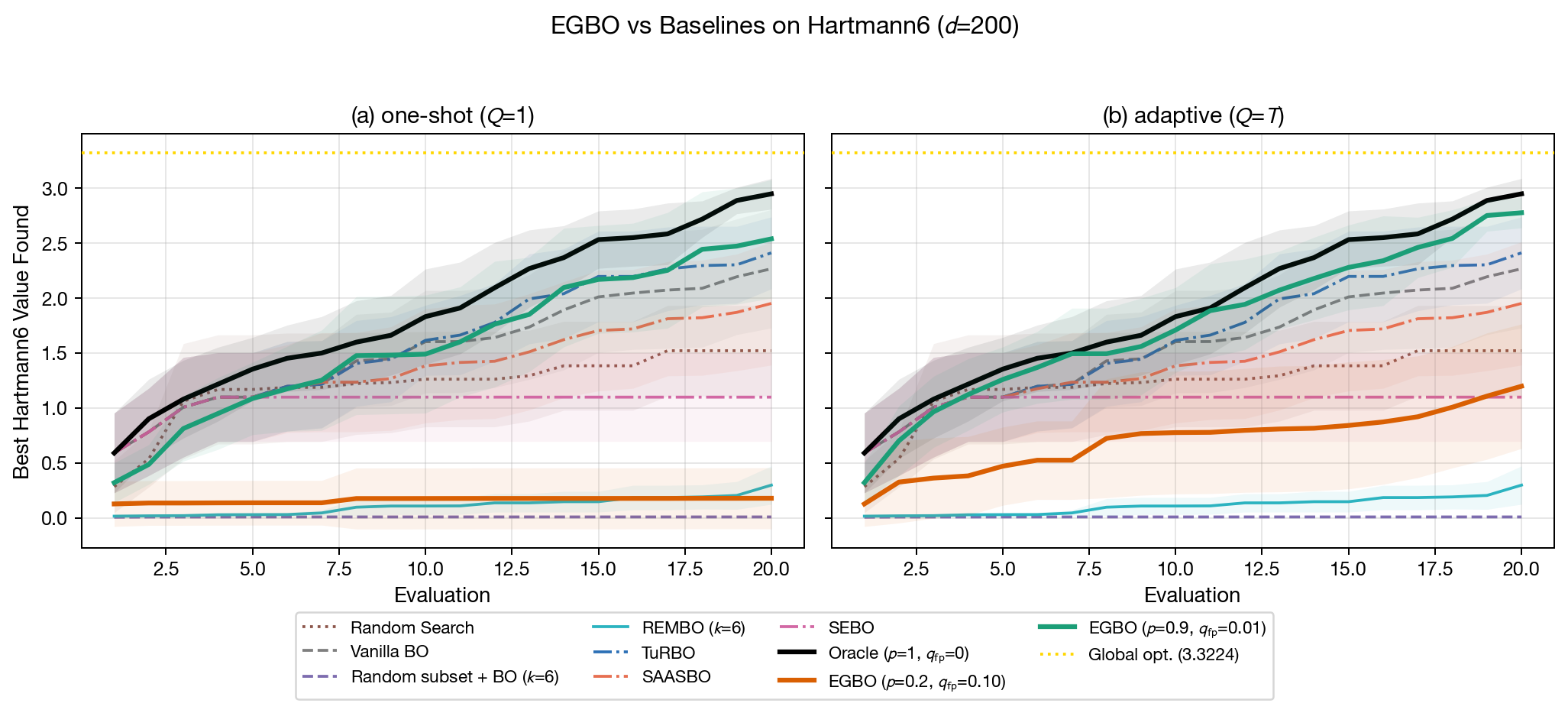}
\caption{Hartmann6: best-so-far convergence with 95\% CIs over 10 seeds.}
\label{fig:hartmann6_cis_for_all}
\end{figure}

Figure~\ref{fig:hartmann6_cis_for_all} shows Figure~\ref{fig:synthetic_main} with 95\% CIs for all methods.

\subsection{Expert-quality heatmap}\label{app:pq_heatmap}

This experiment maps the $(p, q_{fp})$ plane empirically to identify the regimes where EGBO improves over vanilla BO. Recall is swept over $p \in \{0.1, 0.2, 0.4, 0.6, 0.8, 0.9\}$ and false-positive rate over $q_{fp} \in \{0, 0.01, 0.05, 0.10, 0.20\}$, all at adaptive $Q=T$ (the schedule of Figure~\ref{fig:synthetic_main}(b)) to give the expert the maximum number of opportunities to recover from misses. For each cell we run $10$ seeds at $T=20$ evaluations and report the gap-closed fraction $(f_{\text{EGBO}} - f_{\text{vanilla}})/(f_{\text{oracle}} - f_{\text{vanilla}})$, where $1.0$ means EGBO matches the oracle, $0$ means it matches vanilla BO, and negative values indicate EGBO underperforms vanilla BO.

Both recall and false-positive rate matter. With no false positives ($q_{fp}=0$), EGBO closes most of the oracle gap at every $p \geq 0.4$ (gap-closed fractions $0.56$--$0.82$). Once $q_{fp} > 0$, only the high-recall, low-false-positive cells close the majority of the gap ($0.68$ and $0.75$ at $p \in \{0.8, 0.9\}$, $q_{fp}=0.01$). At $q_{fp} \geq 0.05$, EGBO closes at most $38\%$ of the gap even at $p=0.9$, and most cells with $p \leq 0.6$ fall below vanilla BO. At $p \leq 0.2$, EGBO underperforms vanilla BO at every $q_{fp}$. This is consistent with Theorem~\ref{thm:oneshot} and Equation~\eqref{eq:adaptive_size}. False positives enter the bound through $\varepsilon_{\mathrm{BO}}(T, |S|)$, and under adaptive querying they accumulate across rounds, so even a modest $q_{fp}$ inflates $|S_Q|$ enough to erode the benefit of searching a lower-dimensional subspace. The two presets used in Section~\ref{sec:synthetic_benchmark}, \texttt{egbo\_strong} at $(0.9, 0.01)$ (gap-closed fraction $0.75$) and \texttt{egbo\_weak} at $(0.2, 0.10)$ ($-1.57$), sit in the success and failure regions respectively, picked deliberately to illustrate the two regimes.

\begin{figure}[h]
    \centering
    \includegraphics[width=0.7\linewidth]{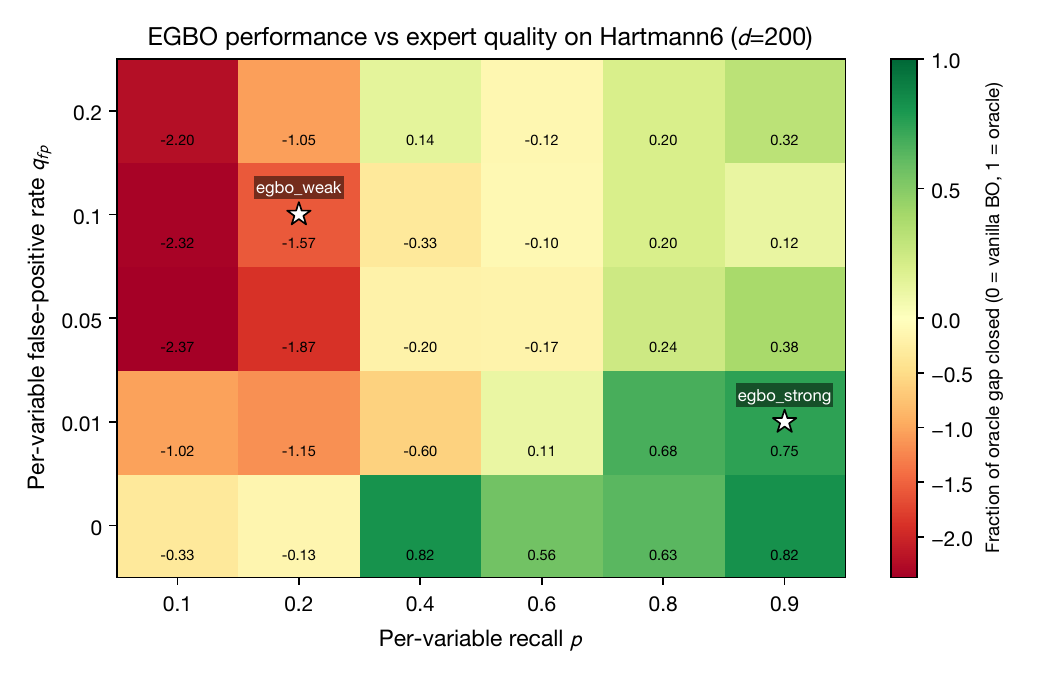}
    \caption{Fraction of the oracle--vanilla gap closed by adaptive EGBO
    ($Q=T$) on Hartmann6 in $N=200$ ambient dimensions, as a function of
    the per-variable recall $p$ (columns) and false-positive rate $q_{fp}$
    (rows). Each cell is the mean over $10$ random seeds, $T=20$
    evaluations; the colour scale is centred at $0$ (vanilla BO), and
    negative values mean EGBO underperforms vanilla BO. Stars mark the two presets used in the main text:
    \texttt{egbo\_strong} at $(p=0.9, q_{fp}=0.01)$ and \texttt{egbo\_weak}
    at $(p=0.2, q_{fp}=0.10)$. EGBO closes the majority of the oracle gap only when
    recall is high and false positives are rare ($p \geq 0.4$ at
    $q_{fp}=0$; $p \geq 0.8$ at $q_{fp}=0.01$). At $q_{fp} \geq 0.05$ it
    closes at most $38\%$ of the gap, and at $p \leq 0.2$ it
    underperforms vanilla BO over the full $200$-dimensional space
    regardless of $q_{fp}$.}
    \label{fig:pq_heatmap}
\end{figure}

\subsection{Intermediate-Q sweep}\label{app:qsweep}

This experiment tests Proposition~\ref{prop:adaptive_coverage}'s prediction that adaptive expert-query rounds help only when the initial prior is incomplete. We sweep $Q \in \{1, 2, 4, 8, 16\}$ for the two representative presets (\texttt{egbo\_strong}, \texttt{egbo\_weak}) at $20$ seeds per cell, with all other settings matching Appendix~\ref{app:exp-hartmann}. In this sweep the expert is queried once before the initial design and once before each of the $T - n_{\text{init}} = 15$ BO steps, so $16$ is the maximum number of distinct query rounds. A run with $Q=20$ reproduces $Q=16$ exactly and is omitted. Figure~\ref{fig:synthetic_main}(b) and the heatmap in Appendix~\ref{app:pq_heatmap} instead query the expert before each of the $20$ evaluations, including the five initial Sobol points. The two schedules give different results on the same seeds. Under the Figure~\ref{fig:synthetic_main}(b) schedule, the strong expert's final regret is $0.55$ (vs.\ $0.79$ one-shot), whereas under the sweep's schedule it is $1.19$ at $Q=16$ (vs.\ $0.98$ at $Q=1$, 20 seeds). Whether adaptivity helps a strong expert therefore depends on when the expert is queried.

The two regimes split along expert quality, as Proposition~\ref{prop:adaptive_coverage} predicts. With a strong expert, regret is essentially flat across $Q$ (gap-to-optimum $0.98$ at $Q=1$, $1.19$ at $Q=16$), because recall is high enough that one round suffices to cover a near-optimal support, and additional rounds only inflate $|S_q|$ with false positives. The trend is within the $95\%$ CIs and we do not claim statistical significance at $\alpha = 0.05$. With a weak expert, regret drops monotonically from $3.20$ at $Q=1$ to $2.68$ at $Q=16$, with the largest improvement between $Q=8$ and $Q=16$, when the expert is consulted before every BO step. The mechanism is made concrete by the $|S_q|$ trajectories in Appendix~\ref{app:active_set}. Weak experts only assemble a usable support after many rounds, while strong experts have nothing to add.

\begin{figure}[h]
    \centering
    \includegraphics[width=0.65\linewidth]{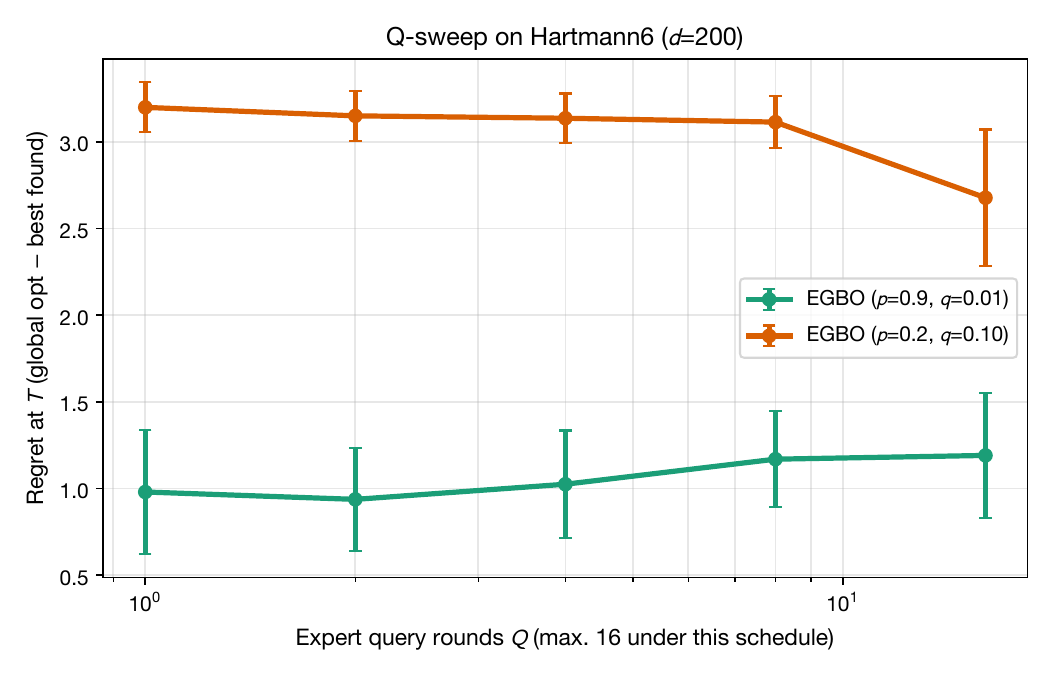}
    \caption{Final regret (gap to the global optimum, lower is better) as a function of
the number of expert query rounds $Q$ on Hartmann6 in $N=200$ ambient
dimensions, $T=20$ evaluations, $20$ random seeds per cell. Error bars
are $95\%$ confidence intervals. With a strong expert ($p=0.9$,
$q_{fp}=0.01$, green), regret is essentially flat across $Q$ and trends
slightly upward from $0.98$ at $Q=1$ to $1.19$ at $Q=16$. Adaptive
querying does not help, and can mildly hurt, a strong expert because
per-iteration false positives accumulate and grow $|S_q|$ beyond the
true $6$-dimensional support, making the BO problem harder under a
fixed evaluation budget (the active-set trajectory in
Appendix~\ref{app:active_set} makes this growth concrete). The trend
is within the $95\%$ CIs and we do not claim statistical significance
at $\alpha=0.05$. With a weak expert ($p=0.2$, $q_{fp}=0.10$, orange),
regret drops from $3.20$ at $Q=1$ to $2.68$ at $Q=16$, with the largest
improvement between $Q=8$ and $Q=16$, once the expert is consulted before
every BO step. $Q=16$ is the maximum number of distinct query rounds
under this schedule ($T - n_{\text{init}} = 15$ BO steps plus the initial query). The two regimes match
the prediction of Proposition~\ref{prop:adaptive_coverage}: adaptivity
helps only when the prior is incomplete.}
    \label{fig:qsweep}
\end{figure}

\subsection{Active-set trajectory}\label{app:active_set}
\begin{figure}[h]
    \centering
    \includegraphics[width=0.95\linewidth]{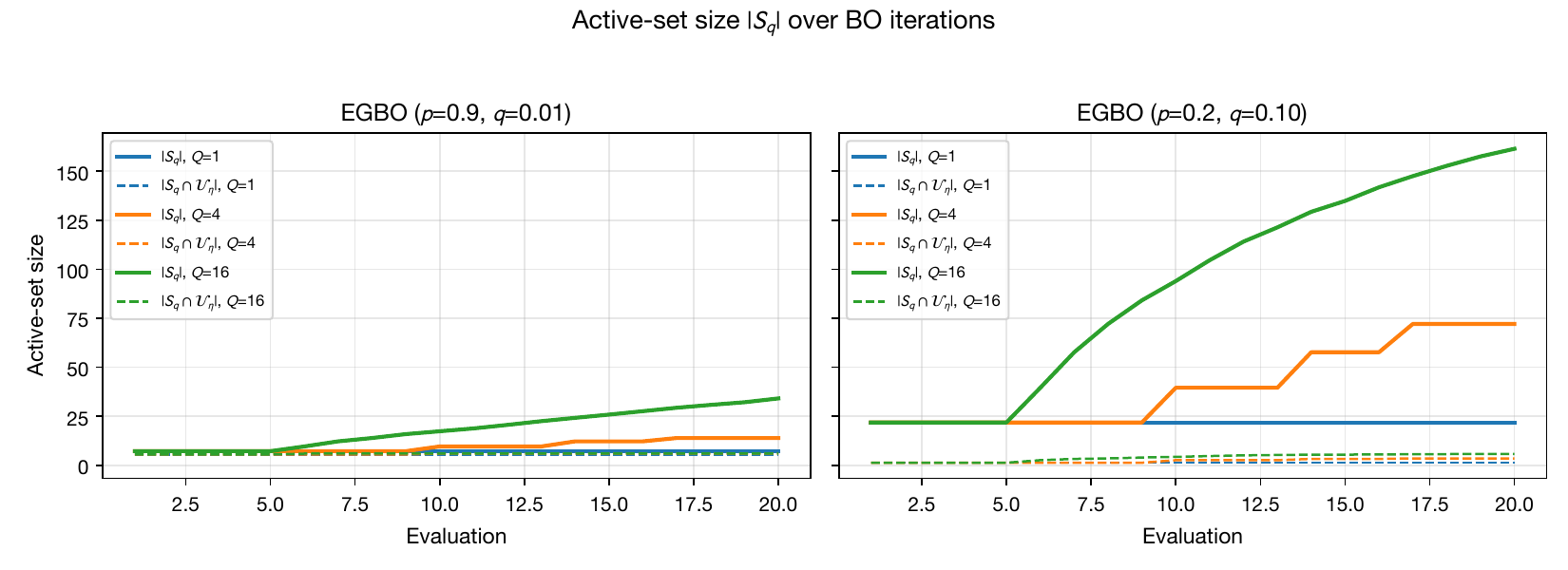}
    \caption{Mean active-set size $|S_q|$ over BO iterations for three
    representative $Q$ values $\{1, 4, 16\}$ on Hartmann6 in $N=200$
    ambient dimensions, $20$ random seeds. \emph{Left:} strong expert
    ($p=0.9, q_{fp}=0.01$). \emph{Right:} weak expert ($p=0.2, q_{fp}=0.10$).
    Solid lines plot $|S_q|$ (total active-set size). Dashed lines plot
    $|S_q \cap \mathcal{U}_\eta|$ (number of relevant variables in the active set;
    $|\mathcal{U}_\eta|=6$ for embedded Hartmann6). Under a strong expert $|S_q|$
    saturates within one round near the true support size; under a weak
    expert, $|S_q|$ grows slowly across many rounds before it covers a
    near-optimal support, and false positives accumulate when $Q$ is
    large. This visualization makes Equation~(\ref{eq:adaptive_size}) of
    Proposition~\ref{prop:adaptive_coverage} concrete: low recall $p$ makes
    round-by-round coverage of $\mathcal{U}_\eta$ unlikely, so weak
    experts require many rounds before the active set is useful.}
    \label{fig:active_set}
\end{figure}

Figure~\ref{fig:active_set} visualizes the active-set dynamics predicted by Equation~(\ref{eq:adaptive_size}) directly on the Hartmann6 task. For each $(Q, \text{expert})$ configuration we record $|S_q|$ (total active-set size) and $|S_q \cap \mathcal{U}_\eta|$ (number of relevant variables in the active set, where $|\mathcal{U}_\eta|=6$ since coordinates $7$--$200$ are inert) at every BO iteration, averaged over $20$ seeds. We plot three representative $Q$ values: $Q=1$ (no expansion after initialization), $Q=4$ (expand every four BO iterations), and $Q=16$ (expand before every BO step), under both expert presets.

The two panels expose the mechanism behind the $Q$-sweep regret pattern in Appendix~\ref{app:qsweep}. Under a strong expert (left), $|S_q|$ jumps near the true support size in the first round and the dashed line ($|S_q \cap \mathcal{U}_\eta|$) tracks the solid one closely throughout. The relevant support is essentially covered after one query, so larger $Q$ only adds the slow accumulation of false positives at rate $q_{fp} = 0.01$ per round per inert coordinate. Equation~(\ref{eq:adaptive_size}) predicts $\mathbb{E}[|S_Q|]$ rises from $\approx 7.3$ at $Q=1$ to $\approx 35$ at $Q=16$ in this regime. The resulting BO dimension cost is what produces the slight upward regret trend in Appendix~\ref{app:qsweep}. Under a weak expert (right), $|S_q|$ grows slowly across rounds for small $Q$ and aggressively for large $Q$, with the dashed line approaching the $|\mathcal{U}_\eta|=6$ ceiling only by mid-trajectory at $Q=16$. This is the regime in which adaptivity pays off, because each additional round provides a fresh chance to recruit a missing relevant variable, at the cost of a large absolute $|S_q|$ dominated by false positives ($\mathbb{E}[|S_{16}|] \approx 164$ by Equation~(\ref{eq:adaptive_size})).

Together, this active-set view, the $Q$-sweep regret in Appendix~\ref{app:qsweep}, and the $(p, q_{fp})$ heatmap in Appendix~\ref{app:pq_heatmap} agree. The green band of the heatmap corresponds to fast $|S_q \cap \mathcal{U}_\eta|$ saturation with controlled $|S_q|$, while the red band corresponds to either incomplete coverage (low $p$) or excessive $|S_q|$ (high $q_{fp}$).

\section{\textsc{FormulateBench}}\label{app:formulatebench}

\subsection{Setup}\label{app:exp-formulatebench}

\paragraph{Problem.}
The ground set is the union of plant-based ingredients in NECTAR with nutrition data in USDA FoodData Central. Before any method or LLM call, we remove animal-derived, ambiguous, or directly target-revealing entries. For meat, we remove \texttt{egg white}, \texttt{egg whites}, \texttt{fat}, \texttt{milk}, and \texttt{milk proteins}; for dairy, we remove \texttt{mozzarella}. This leaves a common eligible pool of \(d=106\) meat ingredients and \(d=90\) dairy ingredients for every method. The objective is the per-dimension \(z\)-score RMSE against an animal-product target on \(m=8\) nutrients (energy, protein, fat, saturated fat, carbohydrates, fiber, sugars, sodium), with a per-category measurement mask \(M\subseteq\{1,\ldots,8\}\) for missing USDA values. Ingredient-set \(z\)-score statistics \((\mu_j,\sigma_j)\) are computed once over the filtered candidate ground set.

\paragraph{Simplex parameterization and initialization.}
Compositions live on \(\Delta^{d-1}\) and are parameterized as logits \(z\in[-12.0,0.0]^d\) mapped through a softmax. The \(-12\) floor (\(e^{-12}\approx1.5\times10^{-6}\)) keeps gradients stable while leaving the simplex effectively open. Conventional BO methods use five method-specific initial evaluations that are not centered on the LLM recipe. Vanilla BO, TuRBO, SAASBO, and SEBO use the uniform mixture plus four seeded Dirichlet draws over the full ingredient simplex. Random-subset BO uses the analogous design within its random \(k=9\) subset, and REMBO uses five seeded Sobol points in its latent space. For EGBO, the LLM's first recipe counts as evaluation 1 and the remaining four initial designs are seeded local logit-normal perturbations with standard deviation \(0.35\) around that recipe. BO acquisition optimization begins at evaluation 6. The fixed-support and unrestricted iterative LLM baselines share the same first recipe and begin feedback-conditioned proposals at evaluation 2.

\paragraph{Normalization across categories.}
Raw distances are not directly comparable because some animal-product targets are intrinsically farther than others from a typical plant-ingredient mixture. We divide each category's trajectory by the mean distance of the first random-search evaluation over 10 seeds. Thus a normalized distance of \(0.5\) means that a method has reduced the distance to half that of an average initial random formulation for that category. Aggregate figures average these normalized best-so-far trajectories across categories; per-category figures report raw distance.

\paragraph{Sparsity bound justifying $k = 9$.}
When the objective is nonnegative least squares with $m$ nutrient targets, a classical result gives a bound on the sparsity of optimal solutions. We use this to set the LLM expert's selection cap to $k = m + 1 = 9$.

\begin{proposition}[Sparsity of NNLS nutrition matching]\label{prop:nnls}
Let $V \in \mathbb{R}_{\geq 0}^{m \times N}$ be a nonnegative ingredient-to-nutrient matrix and $v^* \in \mathbb{R}^m$ a target nutrient profile. Consider the nonnegative least-squares problem
\begin{equation}\label{eq:nnls}
  \min_{x \geq 0} \|Vx - v^*\|_2^2.
\end{equation}
There exists an optimal solution $x^*$ with $\|x^*\|_0 \leq m$.
If the problem additionally imposes a simplex constraint $\sum_{i=1}^N x_i = 1$, there exists an optimal solution with $\|x^*\|_0 \leq m + 1$.
\end{proposition}
\begin{proof}
Among all optimal solutions, pick one for which $|I| = \|x^*\|_0$ is minimal, and let $I = \mathrm{supp}(x^*)$. Suppose for contradiction that $|I| > m$. Then the columns $\{V_i : i \in I\}$ are linearly dependent in $\mathbb{R}^m$, so there exists $d \neq 0$ with $V_I d = 0$. Extending $d$ by zeros outside $I$, the perturbation $x^* + td$ preserves the objective for all $t$. Choose $t$ with largest absolute value such that $x^* + td \geq 0$; at least one coordinate becomes zero, contradicting minimality of $|I|$. Hence $|I| \leq m$.

The simplex-constrained version follows the same argument with $d$ restricted to $\ker V_I \cap \{d : \sum_i d_i = 0\}$, the null space of \(\bigl(\begin{smallmatrix} V_I \\ \mathbf{1}^\top \end{smallmatrix}\bigr) \in \mathbb{R}^{(m+1) \times |I|}\), which has dimension at least $|I| - (m+1)$. If $|I| > m+1$ this is non-trivial, and the rest of the argument gives $|I| \leq m+1$. When no box constraint binds, the argument produces a $k$-sparse optimum for any $k$ at least the stated bound, so $\eta(k) = 0$.
\end{proof}

\paragraph{Budget.}
\(T=20\), \(n_{\mathrm{init}}=5\) for BO methods, and batch size \(q=1\). EGBO and the stochastic conventional optimizers use 10 seeds per category. Each temperature-zero iterative LLM baseline contributes one cached trajectory per category.

\paragraph{Acquisition optimization.}
Both reported EGBO variants use qLogNEI with \texttt{num\_restarts}~\(=10\) and \texttt{raw\_samples}~\(=512\). The default variant uses native BoTorch~0.18.1 \texttt{SingleTaskGP} defaults; the LLM-configured variant uses the single global configuration documented in Appendix~\ref{app:exp-shared}.

\paragraph{Baselines.}
All methods receive the same filtered candidate pool and optimize over the simplex parameterization above. Full-space methods may use every eligible ingredient; the expert-selected support is an EGBO intervention, not a restriction imposed on those baselines. Random-subset-plus-BO uses a fixed random \(k=9\) subset (seed offset \(+10{,}000\)), matching EGBO's maximum initial support size. REMBO uses a Gaussian projection \(A\in\mathbb{R}^{d\times9}\) with latent radius \(\sqrt{9}=3\). TuRBO uses failure tolerance \(\max(4,d)\). We additionally compare with fixed-support and unrestricted iterative LLM optimization as described in Appendix~\ref{app:llm-prompts}.

\paragraph{LLM expert.} Active-set selection uses Claude Opus~4.7 at temperature $0$ as documented in Appendix~\ref{app:llm-prompts}; the prompt is reproduced in Figure~\ref{fig:llm-prompt-formulateBench}.

\paragraph{Compute.} \textsc{FormulateBench} runs were executed on CPU. The SAAS-based methods (SAASBO, SEBO) dominate wall time at $\sim$2--5 minutes per (seed, category, method). The full sweep took on the order of $10^{2}$ CPU hours.

\paragraph{Entry point.} Run \texttt{python run\_formulatebench.py -{}-domain \{meat,dairy\} -{}-budget 20 -{}-seeds 10 -{}-n-init 5}. Per-seed traces are written atomically as JSON.

\subsection{One recipe at a time}\label{app:formulatebench-sequential}
\begin{figure}[h]
\includegraphics[width=\textwidth]{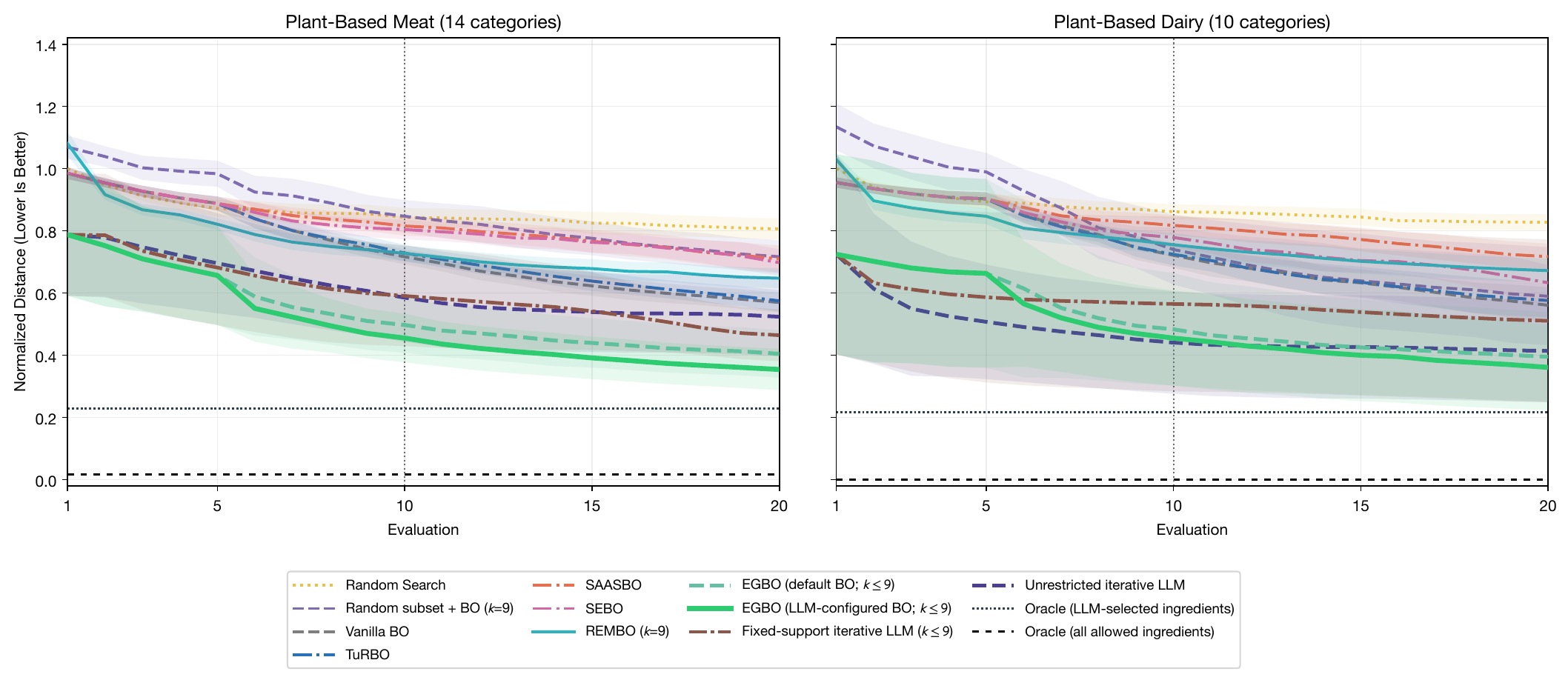}
\caption{\textsc{FormulateBench} convergence for 14 meat and 10 dairy categories. Lower is better. Curves show the mean best-so-far normalized distance across categories, and shaded regions show 95\% confidence intervals across category-level mean trajectories. The two EGBO curves use the same LLM-selected ingredients and initialization and differ only in whether BO uses native BoTorch defaults or the single global LLM-selected configuration. EGBO and the conventional optimization baselines are averaged over 10 optimization seeds; each temperature-zero iterative LLM baseline contributes one cached trajectory per category. The horizontal lines show constrained NNLS oracles using either the LLM-selected ingredients or all allowed ingredients. Results per category are shown in Appendix~\ref{app:formulatebench}.}
\label{fig:meat-dairy-stacked}
\end{figure}

Across all 24 categories, LLM-configured EGBO achieves the lowest mean normalized distance at both 10 and 20 evaluations (lower is better). Figure~\ref{fig:meat-dairy-stacked} reports results separately for meat and dairy. At evaluation 10, LLM-configured EGBO outperforms the unrestricted iterative LLM on meat (\(0.46\) vs.\ \(0.58\)), while the unrestricted LLM has a slightly lower mean distance on dairy (\(0.44\) vs.\ \(0.45\)). By evaluation 20, LLM-configured EGBO has the lower mean distance in both domains, \(0.35\) vs.\ \(0.52\) on meat and \(0.36\) vs.\ \(0.41\) on dairy. 

Across individual categories, LLM-configured EGBO outperforms the fixed-support iterative LLM on 18/24 categories at evaluation 10 and 23/24 at evaluation 20. It outperforms the unrestricted iterative LLM on 16/24 and 20/24 categories, respectively. Per-category results (Appendix~\ref{app:formulatebench_per_category}) show that direct LLM optimization can be strong on individual tasks but substantially worse on others, whereas EGBO is more consistent.

The two NNLS oracles provide an empirical decomposition of the remaining error. At 10 evaluations, the selection and optimization gaps are of comparable magnitude. By 20 evaluations, the residual optimization gap is smaller than the selection gap, indicating that longer runs increasingly become limited by the expert-selected support. Nevertheless, adaptive expert querying (\(Q=2\)), in which the expert may expand the active set after evaluation 10, does not improve performance on average over one-shot EGBO (Appendix~\ref{app:formulatebench-adaptive}).

\subsection{Per-category nutrition results}
\label{app:formulatebench_per_category}
\begin{figure}
\includegraphics[width=\textwidth]{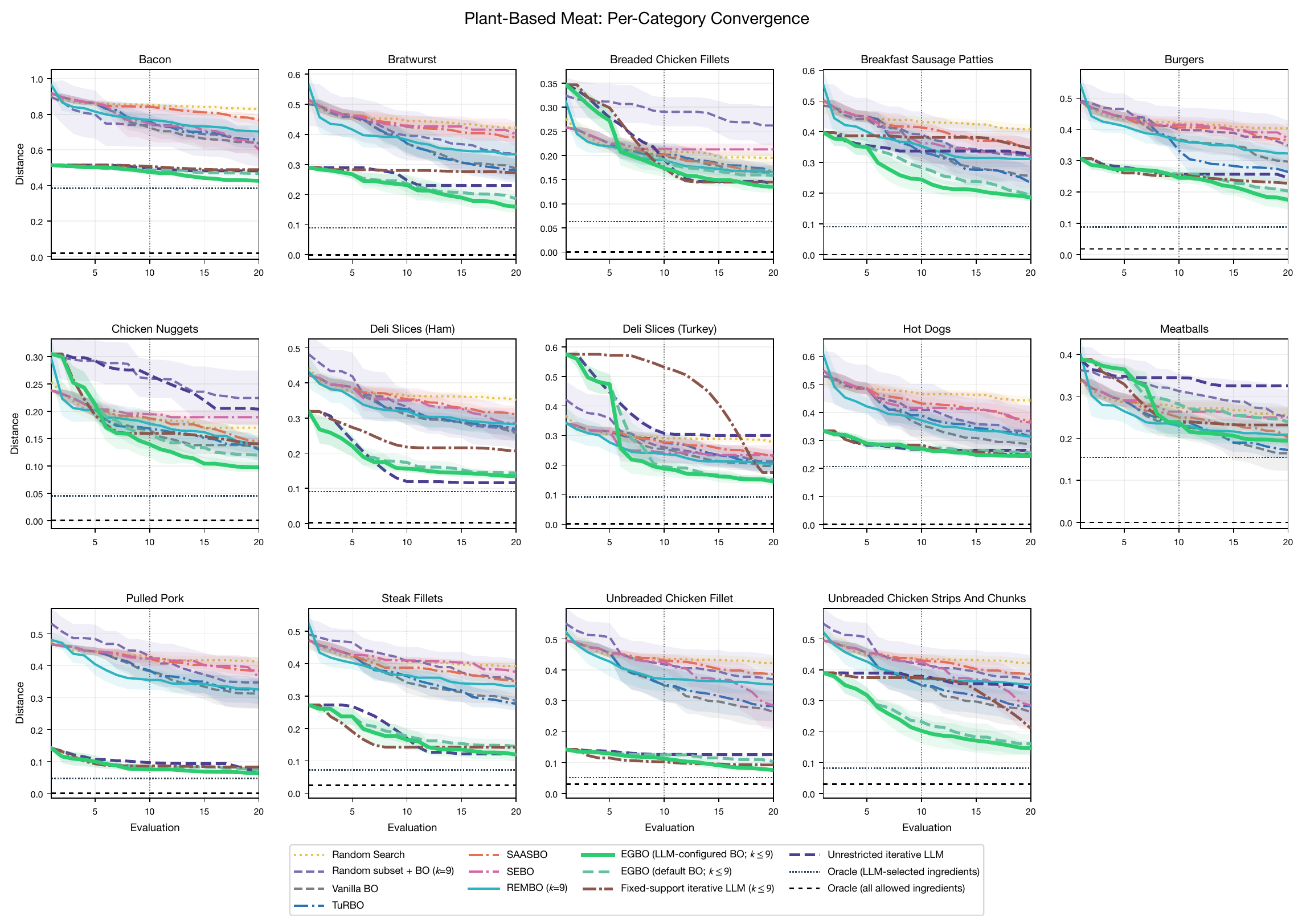}
\caption{\textsc{FormulateBench} per-category convergence for the 14 meat categories. Lower is better. Curves show the mean best-so-far objective distance. The two EGBO variants share the same expert-selected ingredients and initialization and differ only in BO configuration. For EGBO and the conventional optimization baselines, shaded regions show 95\% confidence intervals across 10 optimization seeds; each temperature-zero iterative LLM baseline contributes one cached trajectory per category. The horizontal lines show constrained NNLS oracles using either the LLM-selected ingredients or all allowed ingredients.}
\label{fig:meat-grid}
\end{figure}

\begin{figure}
\includegraphics[width=\textwidth]{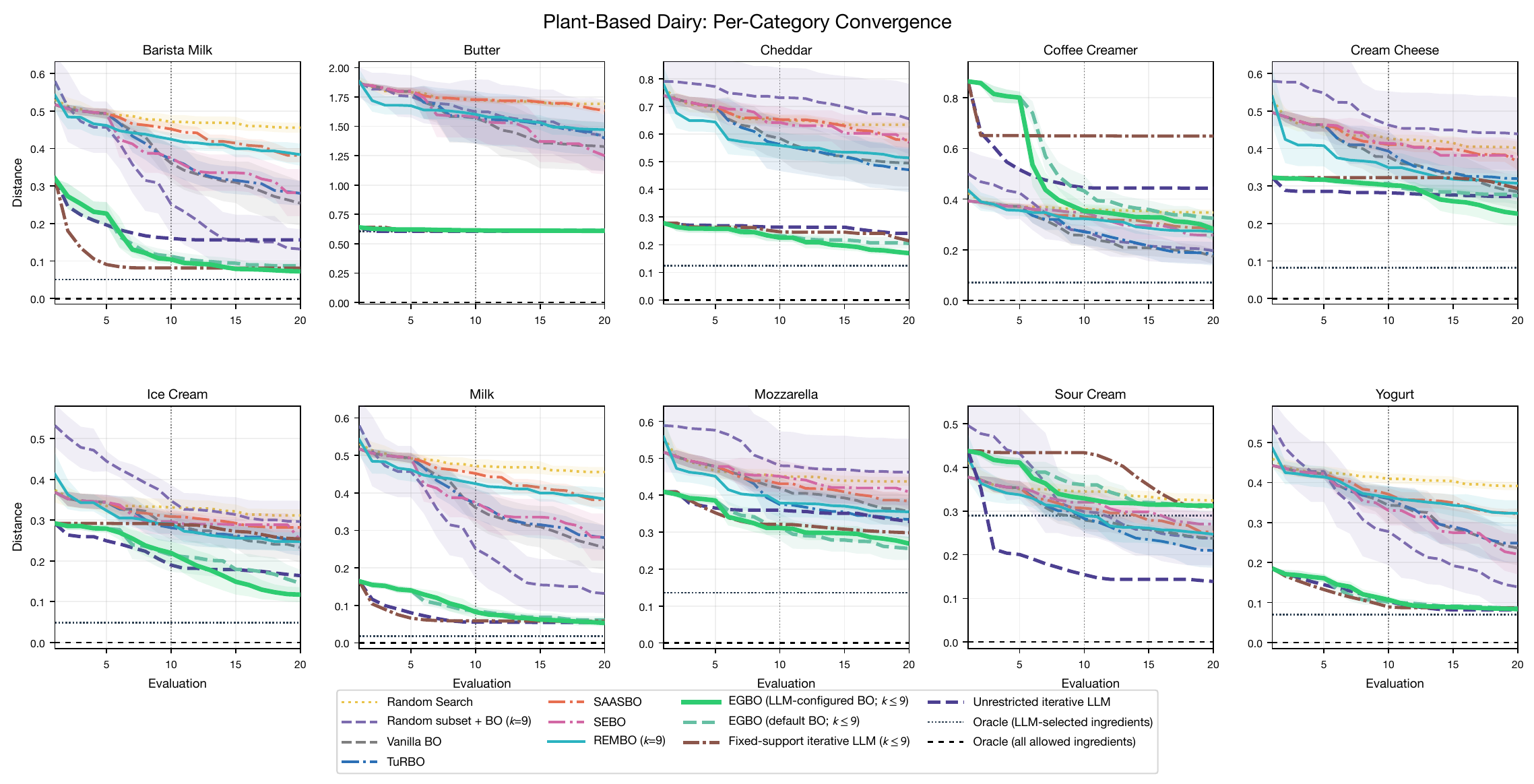}
\caption{\textsc{FormulateBench} per-category convergence for the 10 dairy categories. Lower is better. Curves show the mean best-so-far objective distance. The two EGBO variants share the same expert-selected ingredients and initialization and differ only in BO configuration. For EGBO and the conventional optimization baselines, shaded regions show 95\% confidence intervals across 10 optimization seeds; each temperature-zero iterative LLM baseline contributes one cached trajectory per category. The horizontal lines show constrained NNLS oracles using either the LLM-selected ingredients or all allowed ingredients.}
\label{fig:dairy-grid}
\end{figure}

Figures~\ref{fig:meat-grid} and~\ref{fig:dairy-grid} show results in each category of meat and dairy.

\subsection{Adaptive vs.\ one-shot EGBO}
\label{app:formulatebench-adaptive}
\paragraph{Setup.}
We compare one-shot EGBO (\(Q=1\)) with adaptive EGBO
(\(Q=2\)) under the same information regime as the main experiment. Both methods
begin with the shared LLM recipe. After evaluation 10, the adaptive
expert observes the evaluated formulations and scalar target-dissimilarity
scores and may add up to two ingredients without removing existing
ingredients. Both methods use \(T=20\), \(n_{\mathrm{init}}=5\), and 10 BO seeds per category.

\paragraph{Results.}
Adaptive expansion does not improve performance under the fixed \(T=20\) evaluation budget. On meat, the mean normalized distance at evaluation 20 is \(0.36\) for adaptive EGBO and \(0.35\) for one-shot EGBO. Adaptive EGBO performs better on 4/14 categories. On dairy, the corresponding values are \(0.37\) and \(0.36\), with adaptive EGBO performing better on 2/10 categories. The expert adds an average of 1.81 ingredients for meat and 1.56 for dairy. Thus, expanding the active set increases the BO dimension without producing a reliable reduction in the selection gap at this budget.

\begin{figure}[h]
\centering
\includegraphics[width=\textwidth]{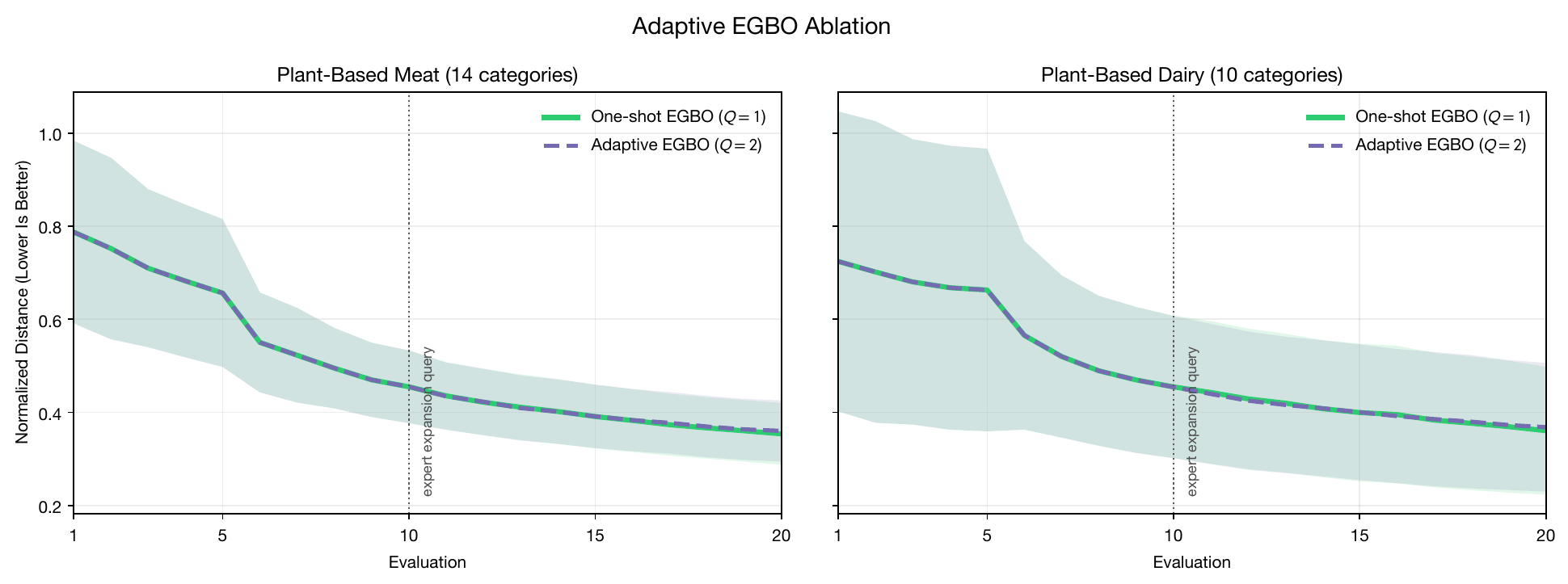}
\caption{One-shot (\(Q=1\)) and adaptive (\(Q=2\)) EGBO on
\textsc{FormulateBench}. The adaptive expert observes
evaluated recipes and scalar target-dissimilarity feedback after
evaluation 10 and may add up to two ingredients without removing
existing ingredients. Lines average category-level means over 10 BO
seeds; bands show 95\% confidence intervals across categories. Adaptive
expansion does not improve performance under the fixed \(T=20\) budget.}
\label{fig:nutrition_pooled}
\end{figure}

\FloatBarrier
\subsection{When does each strategy win?}\label{app:regimes}
\begin{figure}[h]
\centering
\includegraphics[width=\textwidth]{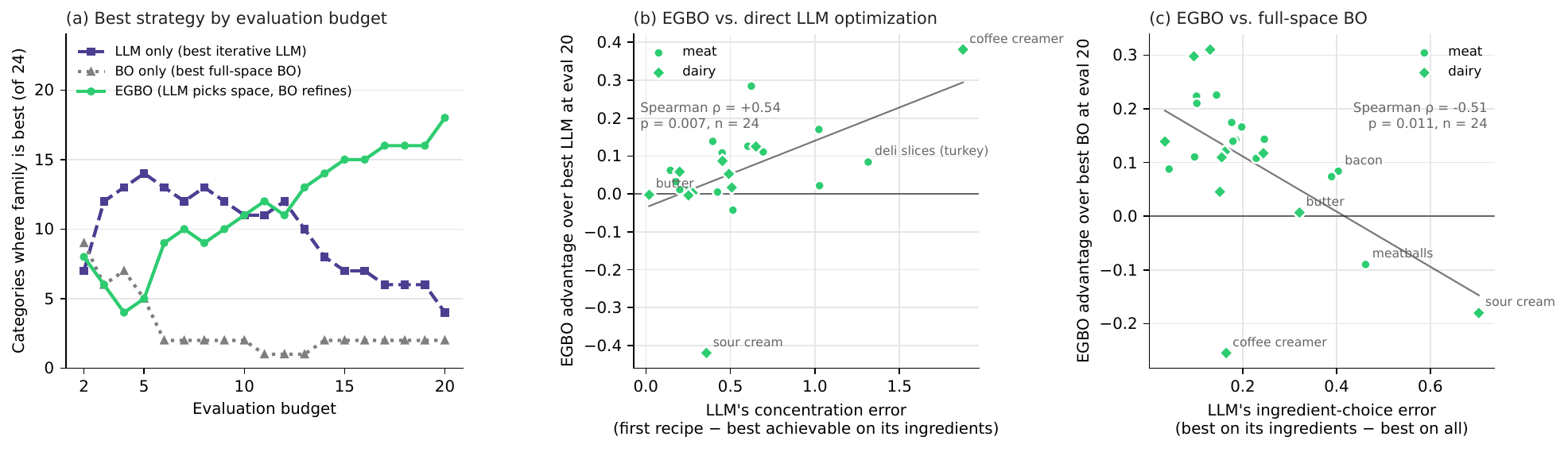}
\caption{Which strategy wins on \textsc{FormulateBench}. \textbf{(a)} Number of categories (of 24) in which each family has the lowest normalized loss at each evaluation; evaluation 1 is omitted because all informed methods share the LLM's first recipe. \textbf{(b)} EGBO's advantage over the best iterative LLM arm at evaluation 20 against the LLM's concentration error. \textbf{(c)} EGBO's advantage over the best full-space BO arm at evaluation 20 against the LLM's ingredient-choice error. Positive advantages favor EGBO; lines are least-squares fits.}
\label{fig:regimes}
\end{figure}

We group the methods into three families: direct LLM optimization (the better of the fixed-support and unrestricted iterative LLM in each category), full-space BO (the better of vanilla BO and BO over all ingredients warm-started from the LLM's first recipe, which uses a Mat\'ern-$5/2$ kernel with a long-lengthscale prior and isotropic lengthscales), and one-shot EGBO with LLM-configured BO. Losses are divided by the category's mean random-search loss at evaluation 1, and per-category values average over seeds. Figure~\ref{fig:regimes}(a) counts the categories in which each family has the lowest loss at each evaluation. To characterize the LLM's knowledge without new experiments, we use the two NNLS oracles: the \emph{concentration error} is the normalized gap between the loss of the LLM's first recipe and $\mathrm{Oracle}_{S_{\mathrm{LLM}}}$, and the \emph{ingredient-choice error} is the normalized gap between $\mathrm{Oracle}_{S_{\mathrm{LLM}}}$ and $\mathrm{Oracle}_{\text{full}}$. For every LLM tested, most of the first recipe's error is concentration error: $0.54$ versus $0.21$ for ingredient choice with Claude Opus~4.7, $0.39$ versus $0.18$ with GPT-6.1 Sol, and $0.71$ versus $0.17$ with Gemini~4 Argon (Appendix~\ref{app:crossmodel}). Both correlations in Figure~\ref{fig:regimes}(b,c) persist when the most extreme category is removed ($\rho=0.53$ and $-0.45$ without sour cream; $0.48$ and $-0.56$ without coffee creamer) and within each domain (meat: $0.35$ and $-0.53$; dairy: $0.64$ and $-0.75$). These are correlations across 24 categories with uncorrected tests. With GPT-6.1 Sol and Gemini~4 Argon as the LLM (Appendix~\ref{app:crossmodel}), the ingredient-choice association replicates ($\rho = -0.52$ and $-0.54$), whereas the concentration association is weaker and not significant. Two caveats apply. First, EGBO spends evaluations 2--5 on local perturbations of the LLM recipe rather than on BO, which contributes to its lag at small budgets. Second, in coffee creamer, whose first recipe is far from optimal, full-space BO outperforms EGBO, suggesting that local initialization around a poor recipe can hurt. The analysis is reproduced from the released results by \texttt{formulateBench/analyze\_regimes.py}.

\subsection{Observation noise}\label{app:noise}
Sensory evaluations are noisy, whereas \textsc{FormulateBench} is deterministic. To test how noise affects each strategy, we rerun the two iterative LLM baselines and one-shot EGBO (LLM-configured BO) with additive Gaussian observation noise, $y_{\mathrm{obs}} = y + \epsilon$ with $\epsilon \sim \mathcal{N}(0, \sigma^2)$, where $\sigma$ is $0.05$ or $0.15$ times the category's mean random-search loss at evaluation 1. All other settings match the main protocol, including the shared first recipe and EGBO's cached ingredient set. EGBO uses 10 seeds; each LLM baseline uses three independent noise draws per category at temperature zero. Because the best observed noisy score is biased toward lucky draws, we score every method by the true (noiseless) loss of the recipe it would recommend, defined as the recipe with the best observed score so far.

\begin{table}[h]
\centering
\small
\setlength{\tabcolsep}{4.5pt}
\begin{tabular}{lccccccc}
\toprule
 & \multicolumn{3}{c}{Evaluation 5} & \multicolumn{3}{c}{Evaluation 20} & EGBO beats \\
\cmidrule(lr){2-4}\cmidrule(lr){5-7}
Noise $\sigma$ & Unrestr. & Fixed & EGBO & Unrestr. & Fixed & EGBO & best LLM \\
\midrule
$0$ & 0.617 & 0.642 & 0.659 & 0.478 & 0.483 & \textbf{0.357} & 20/24 \\
$0.05$ & 0.633 & 0.660 & 0.667 & 0.581 & 0.619 & \textbf{0.404} & 21/24 \\
$0.15$ & 0.688 & 0.750 & 0.698 & 0.688 & 0.748 & \textbf{0.499} & 19/24 \\
\bottomrule
\end{tabular}
\caption{True loss of the recommended recipe under observation noise (normalized; lower is better; mean over 24 categories) for the unrestricted and fixed-support iterative LLM and one-shot EGBO. The last column counts categories (of 24) in which EGBO beats the better LLM arm at evaluation 20. $\sigma$ is a multiple of each category's initial random-search loss; $\sigma=0$ is the main-text setting.}
\label{tab:noise}
\end{table}

\begin{figure}[h]
\centering
\includegraphics[width=\textwidth]{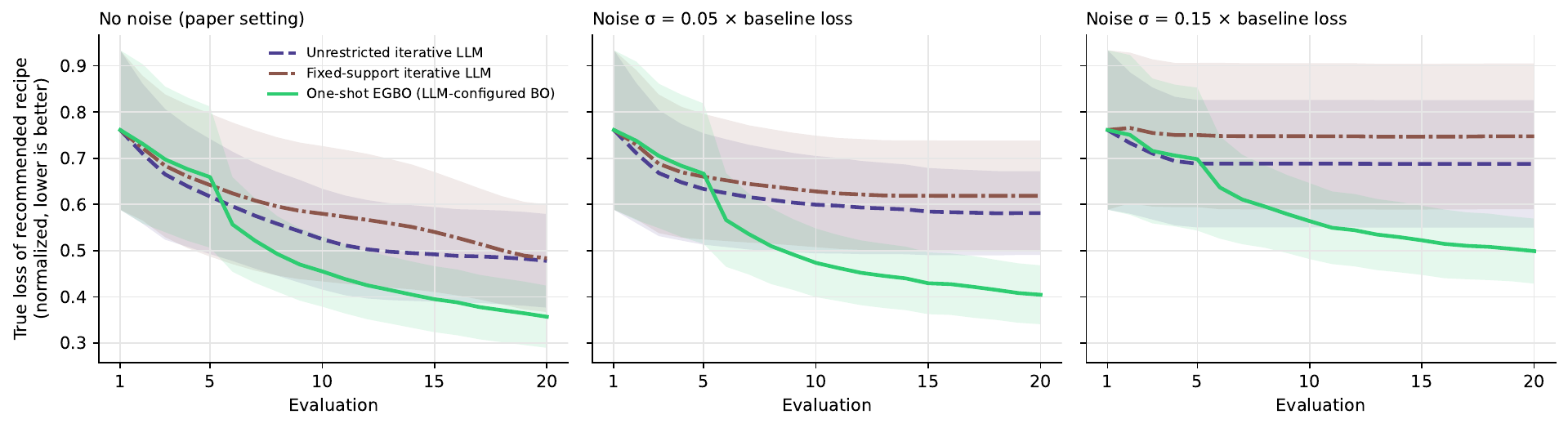}
\caption{\textsc{FormulateBench} under observation noise. True loss of the recommended recipe (normalized; lower is better) for the two iterative LLM baselines and one-shot EGBO, without noise (left) and with Gaussian observation noise of standard deviation $0.05$ (center) and $0.15$ (right) times the category's initial random-search loss. Curves are means over 24 categories (EGBO: 10 seeds; LLM: three noise draws per category). Shaded bands are 95\% CIs across categories and include between-category variation; the paired comparisons in the text are more informative.}
\label{fig:noise}
\end{figure}

Under noise, direct LLM optimization nearly stops improving after evaluation 5. From evaluation 5 to 20, the unrestricted LLM improves by $0.139$, $0.052$, and $0.000$ at $\sigma = 0$, $0.05$, and $0.15$, whereas EGBO improves by $0.302$, $0.262$, and $0.199$ (Table~\ref{tab:noise}, Figure~\ref{fig:noise}). This is a failure of search, not only of selection. At $\sigma=0.15$, the best true loss among all recipes the unrestricted LLM evaluates improves only from $0.644$ to $0.637$ between evaluations 5 and 20. EGBO outperforms the best LLM arm at evaluation 20 at every noise level (Wilcoxon signed-rank $p \le 0.001$), and its deficit at evaluation 5 shrinks from $-0.086$ to $-0.023$ as noise increases (it narrows in 21/24 categories, paired $p = 10^{-4}$). At $\sigma = 0.15$, the two are statistically tied at evaluation 5. The mean advantage at evaluation 20 grows from $0.06$ to $0.18$, but this increase is concentrated in a subset of categories (it increases in 14/24, $p = 0.053$). Our recommendation rule is conservative for EGBO. At $\sigma = 0.15$, the best true recipe EGBO evaluates has loss $0.429$, but the recipe it recommends has $0.499$. Recommending by GP posterior mean narrows this slightly, to $0.489$. Scoring each method by its own pick instead (the GP posterior-mean minimizer for EGBO, and for the LLM the recipe it states it would recommend) leaves the comparison unchanged, with EGBO better in 20/24 categories at evaluation 20. The sweep is reproduced by \texttt{formulateBench/run\_noise\_sweep.py} and analyzed by \texttt{formulateBench/analyze\_noise\_sweep.py}; cached LLM responses are released.

\paragraph{Telling the LLM the noise level.} In a further arm, the iterative LLM's feedback ends with a note that scores include Gaussian measurement noise, stating its exact standard deviation ($\sigma = 0.15$). The note did not help. With Claude Opus~4.7, the better LLM arm's mean distance at evaluation 20 was $0.70$ with the note versus $0.68$ without (EGBO: $0.50$), and the LLM tied or beat EGBO in 6/24 categories versus 7/24. With GPT-6.1 Sol, the values were $0.575$ versus $0.568$ (EGBO: $0.50$), and 4/24 versus 6/24. Both comparisons use the first noise draw of each LLM arm (\texttt{formulateBench/analyze\_noise\_aware.py}).

\subsection{Batched experiments}\label{app:batched}
To mirror the laboratory deployment, which prepares and evaluates three recipes per session, we run 10 batches of 3 recipes (30 evaluations). For a given LLM, the first batch is identical for all LLM-based arms and consists of the LLM's first recipe and two seeded logit-normal perturbations of it ($\sigma = 0.35$, as in the local initialization of Appendix~\ref{app:exp-formulatebench}). In each later batch, the iterative LLM arms are given all previous formulations and scores and asked for three formulations to be evaluated in parallel (the unrestricted arm from the full catalog with at most nine ingredients, the fixed-support arm from the LLM's first-recipe ingredients). The prompt is otherwise that of the one-at-a-time protocol. EGBO fits the LLM-configured GP to all observations on the first-recipe ingredients and proposes three recipes with qLogNEI ($q = 3$, sequential greedy optimization). The conventional baselines do not use the LLM and search the full catalog. Random search draws three Dirichlet(1) recipes per batch. Vanilla BO starts from three simplex-initial points and then proposes batches of three with a BoTorch-default GP and qLogNEI. TuRBO does the same within a trust region (success tolerance 3, failure tolerance $\max(4, \lceil d/3 \rceil)$). Under noise, each evaluation returns the true distance plus Gaussian noise with standard deviation $0.15$ times the category normalizer, and methods are scored by the true distance of the recipe with the best observed score. EGBO and the baselines use 10 seeds per category; each iterative LLM arm uses one temperature-zero trajectory. Batch diversity is the mean pairwise $L_1$ distance between the three recipes of a batch, averaged over batches 2--10. Table~\ref{tab:batched} summarizes the results after 30 evaluations. With Claude Opus~4.7, EGBO outperforms vanilla BO and TuRBO on 20/24 categories each without noise, and on 21/24 and 22/24 with noise. The LLM arms' batches are near-duplicates (mean pairwise $L_1$ distance $0.04$--$0.08$), whereas EGBO's are spread out ($0.43$--$0.47$).

\begin{table}[h]
\centering
\caption{Batched \textsc{FormulateBench} (10 batches of 3 recipes) after 30 evaluations: mean normalized distance over 24 categories (lower is better). ``LLM'' is the better of the two iterative LLM arms per category; ``EGBO better'' counts categories in which EGBO has the lower distance, with a two-sided Wilcoxon signed-rank $p$-value. Vanilla BO and TuRBO do not use the LLM.}
\label{tab:batched}
\small
\setlength{\tabcolsep}{5pt}
\begin{tabular}{llccccc}
\toprule
Noise & LLM & LLM & EGBO & EGBO better & Vanilla BO & TuRBO \\
\midrule
None & Claude Opus 4.7 & 0.44 & 0.34 & 19/24 ($p<0.001$) & 0.52 & 0.51 \\
     & GPT-6.1 Sol     & 0.31 & 0.32 & 9/24 ($p=0.36$)   & & \\
     & Gemini 4 Argon  & 0.32 & 0.35 & 10/24 ($p=0.90$)  & & \\
\midrule
$\sigma = 0.15$ & Claude Opus 4.7 & 0.68 & 0.48 & 19/24 ($p<0.001$) & 0.70 & 0.73 \\
     & GPT-6.1 Sol     & 0.56 & 0.47 & 18/24 ($p<0.001$) & & \\
     & Gemini 4 Argon  & 0.58 & 0.50 & 15/24 ($p=0.12$)  & & \\
\bottomrule
\end{tabular}
\end{table}

\subsection{Other LLMs, one recipe at a time}\label{app:crossmodel}
We repeated the comparison with two other LLMs accessed through OpenRouter, GPT-6.1 Sol and Gemini~4 Argon. Each LLM proposes its own first recipe, which fixes EGBO's ingredient set. Otherwise, the protocol, prompts, and BO configuration are those of the main comparison (one LLM trajectory per category for each iterative LLM arm, ten EGBO seeds). Under observation noise ($\sigma = 0.15$ times the category normalizer, Appendix~\ref{app:noise}), each method is scored by the true distance of the recipe with the best observed score. Table~\ref{tab:crossmodel} reports results at evaluation 20 of the one-recipe-at-a-time protocol (Appendix~\ref{app:formulatebench-sequential}); batched results for GPT-6.1 Sol and Gemini~4 Argon are in Table~\ref{tab:batched}. EGBO's final distance varies little across LLMs, whereas direct LLM optimization varies more. Without noise, GPT-6.1 Sol and Gemini~4 Argon match EGBO (differences not significant). Under noise, direct LLM optimization improves little after evaluation 5 for every LLM tested, and EGBO is better for Claude Opus~4.7 and GPT-6.1 Sol but not for Gemini~4 Argon, whose first proposals were already strong. Letting each LLM choose its own BO configuration, as Claude Opus~4.7 does in the main comparison, changed EGBO's mean final distance by at most $0.02$. The ingredient-choice association of Appendix~\ref{app:regimes} replicates with every LLM (Spearman $\rho = -0.52$ and $-0.54$ for GPT-6.1 Sol and Gemini~4 Argon, both $p < 0.01$). The concentration association, in contrast, is weaker and not significant ($\rho = 0.21$ and $0.31$).

\begin{table}[h]
\centering
\caption{One recipe at a time: direct LLM optimization (better of the two iterative LLM arms) vs.\ EGBO at evaluation 20 across LLMs. Mean over 24 categories of the normalized distance, taking the better LLM arm per category (lower is better); ``EGBO better'' counts categories in which EGBO has the lower distance, with a two-sided Wilcoxon signed-rank $p$-value. Claude Opus~4.7 is the main comparison; under noise, its iterative LLM arms are averaged over three noise draws.}
\label{tab:crossmodel}
\small
\setlength{\tabcolsep}{4.5pt}
\begin{tabular}{lcccccc}
\toprule
 & \multicolumn{3}{c}{No noise} & \multicolumn{3}{c}{Noise ($\sigma = 0.15$)} \\
\cmidrule(lr){2-4}\cmidrule(lr){5-7}
LLM & LLM & EGBO & EGBO better & LLM & EGBO & EGBO better \\
\midrule
Claude Opus 4.7 & 0.42 & 0.36 & 20/24 ($p<0.001$) & 0.68 & 0.50 & 19/24 ($p<0.001$) \\
GPT-6.1 Sol     & 0.37 & 0.37 & 13/24 ($p=0.55$)  & 0.57 & 0.50 & 18/24 ($p<0.001$) \\
Gemini 4 Argon  & 0.34 & 0.39 & 9/24 ($p=0.37$)   & 0.55 & 0.55 & 9/24 ($p=0.29$) \\

\bottomrule
\end{tabular}
\end{table}

\subsection{Do tool-using agents delegate?}\label{app:agents}
The iterative LLM baselines propose recipes directly and have no tools. We tested whether an agent with tools would choose, on its own, to model its experiments statistically and thereby delegate refinement to an optimizer. We gave three LLMs (Claude Opus~4.7, GPT-6.1 Sol, and Gemini~4 Argon) two tools: \texttt{evaluate}, which submits one formulation (at most nine catalog ingredients) and returns its noisy score, and \texttt{python}, which runs code in a sandbox with no network access and no access to the benchmark code. The evaluated history is always available in the sandbox as a file. The task prompt is identical to the unrestricted iterative LLM's. Only its final paragraph differs, to state the budget (20 evaluations, including the LLM's first recipe) and describe the tools, including that ``numpy, scipy, scikit-learn and pandas are installed.'' Nothing in the prompt mentions Bayesian optimization, surrogate models, or noise. We ran 24 categories at noise level $\sigma = 0.15$ (Appendix~\ref{app:noise}), one run per category, in a minimal tool-calling loop rather than a production agent harness, scored by the true distance of the best observed recipe. A second arm appends one sentence: ``Measurements are noisy and each evaluation is expensive, so consider carefully how best to choose each experiment.'' For GPT-6.1 Sol, a third arm instead names BoTorch among the installed libraries. We classified each run from the code the agent executed (no code, analysis only, a fitted model such as a regression, or a Gaussian process) and checked the classification against 22 transcripts read in full.

\begin{table}[h]
\centering
\caption{Tool-using agents on \textsc{FormulateBench} with observation noise ($\sigma = 0.15$; 24 categories, one run each). Columns count runs that wrote no code, fit any model of the data, or fit a Gaussian process, with the median evaluation (rounded) at which the first Gaussian process was fit; no run imported BoTorch. Distances are mean true normalized distances at evaluation 20 (lower is better). ``No tools'' is the unrestricted iterative LLM with the same prompt; EGBO uses the same first recipe. }
\label{tab:agents}
\small
\setlength{\tabcolsep}{4pt}
\begin{tabular}{llcccccccc}
\toprule
LLM & Prompt & No code & Any model & GP & First GP & Agent & No tools & EGBO \\
\midrule
Claude Opus 4.7 & base      & 0  & 0  & 0  & --- & 0.58 & 0.70 & 0.50 \\
                & + hint    & 0  & 0  & 0  & --- & 0.59 &      &      \\
GPT-6.1 Sol     & base      & 20 & 4  & 1  & 9   & 0.48 & 0.58 & 0.50 \\
                & + hint    & 15 & 7  & 4  & 12  & 0.46 &      &      \\
                & + BoTorch & 12 & 10 & 6  & 13  & 0.51 &      &      \\
Gemini 4 Argon  & base      & 0  & 24 & 20 & 13  & 0.66 & 0.57 & 0.55 \\
                & + hint    & 1  & 23 & 22 & 10  & 0.66 &      &      \\
\bottomrule
\end{tabular}
\end{table}

Table~\ref{tab:agents} summarizes the results. No agent ran a BO loop over its remaining budget. Claude Opus~4.7 wrote code in every run, but only to tabulate past results. It never fit a model, and under noise it anchored on the recipe with the luckiest score, re-evaluating and perturbing it (``All perturbations fail. Iter 5 remains best at 0.0236''). Under the base prompt, GPT-6.1 Sol usually wrote no code. When it fit a Gaussian process, it was competently specified (scaled inputs, a noise kernel, and a mean-minus-uncertainty selection rule), but it was fit once, late in the budget, to propose a single recipe, after which the agent returned to proposing recipes directly. Gemini~4 Argon fit models in nearly every run, but ad hoc, using linear regressions, line searches, and Gaussian processes whose posterior mean it minimized within narrow bounds it chose by hand. It had the worst final distance of the three agents, worse than its own iterative LLM baseline (base prompt, $p = 0.03$). The hint changed behavior only modestly, and naming BoTorch reduced the GPT-6.1 Sol runs without code from 20 to 12 but led no run to import it. The agents' final distances are not directly comparable with the iterative LLM baselines, because an agent keeps its own reasoning across turns and knows its budget. The GPT-6.1 Sol agent, for example, outperforms its no-tools baseline even in runs without code. We therefore do not attribute these gains to tool use. EGBO, in contrast, uses the same first recipe and budget, and no agent arm outperformed it significantly. The behavioral conclusion is that these agents could model their data but did not adopt modeling as a strategy.

\FloatBarrier
\section{Real-World Deployment}\label{app:deploy}

\subsection{Shared Setup}\label{app:exp-deploy}

\paragraph{Problem.} The candidate ingredient pool is a food-company catalog of $N \approx 88{,}000$ entries. Prior to optimization, the company's Food Scientist Agent (FSA, an LLM-based agentic framework) performs one-shot variable selection ($Q = 1$) yielding $d = 7$ active ingredients for yogurt and $d = 9$ for cottage cheese. All other $N - d$ ingredients are fixed at zero. Variables are encoded as continuous quantities with per-ingredient $[\min, \max]$ bounds taken from the company catalog. Property and quantity constraints are enforced via linear inequality constraints passed to \texttt{optimize\_acqf}.

\paragraph{Budget and batching.} Each campaign runs $T = 10$ iterations with a laboratory batch of $q = 3$ recipes per iteration, for $30$ total recipes. The application draws scrambled Sobol batches from a pool of $2{,}048$ feasible candidates (Sobol seed equals the number of observations, for reproducibility) until at least $n_{\mathrm{init}} = 5$ observations exist (see below for how each campaign started). Subsequent batches are produced by qLogNEI with \texttt{num\_restarts}~$=20$, \texttt{raw\_samples}~$=1024$, \texttt{sample\_shape}~$=[512]$, and \texttt{sequential}~$=$~True so each of the three batch members is selected greedily conditional on the previously chosen ones. The Streamlit application optionally enables an Input Warping kernel (\texttt{robust}~$=$~True) for non-stationarity; the campaigns reported here use the default unwarped configuration.

\paragraph{Starting recipe.} The FSA's recipe was panel-scored before optimization ($f = 0.767$ for yogurt, $0.776$ for cottage cheese). In the yogurt campaign, this observation seeded the application before iteration~1 and is not among the 30 recipes. Iterations~1--2 are Sobol batches and BO proposals begin at iteration~3. In the cottage cheese campaign, iteration~1 consisted of the FSA recipe (recipe~1.1) and two Sobol recipes generated by the application (recipes~1.2--1.3), all counting toward the 30-recipe budget. Iteration~2 is a Sobol batch, and BO proposals begin at iteration~3.

\paragraph{Utility.} Sensory scores are entered through the Streamlit UI (\texttt{app.py}). \texttt{food\_bo.py:\_compute\_utility} normalizes each attribute to $[0, 1]$ using the configured panel scale and converts it to a per-attribute utility according to the goal (max/min/target). The scalar objective is the weighted sum $\sum_{i} w_{i} u_{i}$ with uniform weights $w_{i} = 1/m$, matching the main-text utility definition. In the cottage cheese campaign the weights were entered as $0.17$ each (summing to $1.02$); see the footnote in Section~\ref{sec:real}.

\paragraph{Reproducibility.} Sobol scrambling and a \texttt{torch.manual\_seed} call both use \texttt{len(self.X\_history)} so reruns at any iteration deterministically reproduce the proposed batch given the recorded panel scores. Campaign state is persisted to pickle and exported as JSON. We release the full optimization variables, ingredient lists, panel protocols, and per-iteration recipes alongside the code.

\paragraph{Compute.} Each iteration's acquisition optimization runs in under a minute on a single CPU. The binding constraint is laboratory time ($\sim$4~hours per recipe batch), giving $\sim$40~hours per campaign.

\subsection{Recipe Formulations}
\label{app:recipes}

This appendix documents the output of FSA, final ingredient compositions and processing
protocols for two plant-based products developed under the EGBO framework:
the plant-based yogurt corresponding to the closed-loop deployment, and an analogous plant-based cottage cheese
formulation. Quantities are reported as supplied to the laboratory.
Ingredients held fixed during optimization are flagged as constants;
all others were varied by EGBO.

\subsubsection{Plant-Based Yogurt}
\label{app:yogurt-recipe}

Table~\ref{tab:yogurt-ingredients} reports the FSA starting recipe
(total batch mass $630.00~\mathrm{g}$) and the best formulation found by
EGBO (iteration~7, recipe~3). Soy protein isolate and tricalcium phosphate were
held constant across all iterations, serving as a functional protein
backbone and a calcium fortificant, respectively. The remaining seven
ingredients constitute the active EGBO design space.

The yogurt was prepared according to the following three-step protocol.

\paragraph{Step 1: Powder dispersion.}
All ingredients listed in Table~\ref{tab:yogurt-ingredients}, including
the calcium fortificant, were combined and dissolved using a high-shear
mixer at $3000~\mathrm{RPM}$, $25~^{\circ}\mathrm{C}$, for
$5~\mathrm{min}$.

\paragraph{Step 2: Cooking.}
The dispersed yogurt base was cooked using a moist-heat boiling protocol
in a covered pot (cover closure $90\%$) at $90~^{\circ}\mathrm{C}$ for
$8~\mathrm{min}$.

\paragraph{Step 3: Chilling.}
The cooked product was transferred to a refrigerator and chilled prior to
sensory evaluation.

\subsubsection{Plant-Based Cottage Cheese}
\label{app:cottage-recipe}

Table~\ref{tab:cottage-ingredients} reports the ingredient composition of
the plant-based cottage cheese formulation. The first two ingredients
form the curd-forming substream, which is subsequently centrifuged to remove whey. The remaining seven ingredients form the cooked dressing,
which is combined with the rinsed curds. The table reports the FSA starting
recipe (recipe~1.1) and the best recipe found by EGBO (recipe~10.1).
Weight fractions are reported as supplied to the process.

The cottage cheese was prepared according to the following seven-step
protocol.

\paragraph{Step 1: Pasteurization.}
Soy milk was heat-treated for microbial-load reduction using a batch
pasteurizer at $90~^{\circ}\mathrm{C}$ for $5~\mathrm{min}$ with full
cover closure ($100\%$). Output: pasteurized soy milk.

\paragraph{Step 2: Acid coagulation.}
Pasteurized soy milk was combined with distilled vinegar in a mixing
vessel operated at $200~\mathrm{RPM}$, $75~^{\circ}\mathrm{C}$, for
$2~\mathrm{min}$ with $80\%$ cover closure. Output: coagulated soy
mixture.

\paragraph{Step 3: Curd separation.}
The coagulated mixture was separated by basket centrifugation at
$800~\mathrm{RPM}$ ($50~\mathrm{Hz}$ drive frequency),
$25~^{\circ}\mathrm{C}$, for $3~\mathrm{min}$ in a single pass. Outputs:
soy curds; soy whey (discarded).

\paragraph{Step 4: Curd rinsing.}
Soy curds were washed with tap water in a washing machine at
$10~^{\circ}\mathrm{C}$ for $1~\mathrm{min}$ (single pass). Outputs:
rinsed soy curds; spent rinse water.

\paragraph{Step 5: Dressing cooking.}
Coconut cream, tap water, lemon juice, salt, nutritional yeast, tapioca
starch, and xanthan gum were combined and cooked using a moist-heat
boiling protocol in a covered pot (cover closure $90\%$) at
$90~^{\circ}\mathrm{C}$ for $8~\mathrm{min}$. Output: cooked plant-based
dressing.

\paragraph{Step 6: Combination.}
Rinsed soy curds and cooked dressing were combined in a planetary mixer
at $60~\mathrm{RPM}$, $20~^{\circ}\mathrm{C}$, for $2~\mathrm{min}$.
Output: unchilled plant-based cottage cheese.

\paragraph{Step 7: Chilling.}
The product was cooled to $4~^{\circ}\mathrm{C}$ in a refrigerator.
Output: plant-based cottage cheese (final). 

\subsection{Sensory evaluation protocol}\label{app:sensory-protocol}

The utility scores reported in Section~\ref{sec:real} were obtained via a structured descriptive sensory evaluation conducted by a trained panel of five participants. The procedure was held fixed across all $T = 10$ EGBO iterations to ensure that observed utility differences reflect genuine recipe-level differences. The sensory panel is treated as a noisy black-box oracle, where for a formulation $x \in \mathbb{R}^N$, the observed score is $y = f(x) + \varepsilon$, with $\varepsilon$ capturing inter-panelist and session variability.

\paragraph{Panel composition and preparation.} All panelists were members of a trained sensory panel with prior experience in structured attribute-based evaluation, and each session was preceded by a calibration briefing aligning the panel on the target attributes and on the $0$--$15$ intensity scale. To ensure a neutral palate at the time of evaluation, panelists abstained from food consumption for at least two hours before the session, and from strong-flavored stimuli (spicy or acidic foods, alcohol, coffee, smoking) for at least six hours before the session.

\paragraph{Evaluation environment.} Sessions were conducted in a dedicated sensory evaluation room designed to minimize external bias. The space was isolated from ambient odors and distractions, and environmental conditions (temperature, noise level, airflow) were held stable across sessions. Controlled lighting was applied to suppress visual bias on attributes such as color and surface appearance. Panelists worked in individual booths, did not communicate during the session, and did not discuss or compare samples. Each panelist evaluated samples in a predefined sequence determined prior to the session.

\paragraph{Calibration system.} A Golden Standard reference sample was made available throughout each session as a continuous calibration anchor. Panelists were instructed to revisit the reference between samples to maintain a stable internal reference point, and printed $0$--$15$ attribute scales were provided before the start of evaluation. This combination of a physical reference and an explicit scale was used to control for within-session and between-session drift in panelist perception.

\paragraph{Evaluation procedure.} The protocol followed a comparative descriptive analysis approach: attribute-based, comparative against the Golden Standard, and non-hedonic (preference and liking are not recorded). Each sample was assessed via a fixed sequence of stages, each applied where relevant to the product class:
\begin{itemize}
\item \textbf{Visual assessment.} Density, surface characteristics, structural integrity, and overall first impression.
\item \textbf{Aroma.} Identification of intensity and key aromatic notes under controlled exposure to avoid sensory fatigue.
\item \textbf{Texture and mouthfeel.} Initial perception on the first spoon or bite, creaminess and slipperiness, graininess, and cohesion or structural behavior under deformation.
\item \textbf{Flavor.} Basic taste profile (sweet, sour, salty, bitter, umami), flavor identity (e.g., dairy, fermented, clean, off-notes), balance, and aftertaste.
\item \textbf{Overall assessment.} Holistic evaluation and identification of key strengths and weaknesses.
\end{itemize}
For the plant-based yogurt campaign, the four target attributes (Consistency, Creaminess, Tanginess, Uniformity) were drawn from the Texture and Flavor stages above and recorded on the $0$--$15$ scale. For plant-based cottage the same was applied for the six target attributes (Homogeneity, Dispersion Phase, Slipperiness, Sourness, Dairy and Off-Notes).

\paragraph{Between-sample protocol.} To preserve sensory clarity across samples, panelists cleansed their palate with water between evaluations, with neutral carriers (bread and crackers) available as an option, and observed a short reset period before proceeding to the next sample. The Golden Standard was re-tasted as needed during this interval to realign perception.

\paragraph{Objectivity, fatigue management, and performance principle.} The panel operated under strict objectivity principles. Evaluation was based on the defined sensory attributes rather than personal preference, no response was treated as right or wrong, and consistency across samples was prioritized over individual bias. When sensory fatigue or attribute confusion arose, panelists followed a four-step recovery protocol: pausing evaluation, cleansing the palate, re-evaluating the Golden Standard, and then resuming testing. The overall protocol prioritized accuracy, consistency, and repeatability of the recorded scores; speed of evaluation was treated as secondary to data quality.

\paragraph{Data aggregation.} Panelist scores were aggregated immediately after evaluation into a single set of attribute-level scores. Individual panelist responses were not retained, and no individual-level data were used in subsequent analysis. The utility values reported in the main text were computed from these aggregated scores using the scalar utility function defined in Section~\ref{sec:real}.

\subsection{Yogurt campaign}\label{app:yogurt}

\begin{figure}[t]
    \centering
    \includegraphics[width=\linewidth]{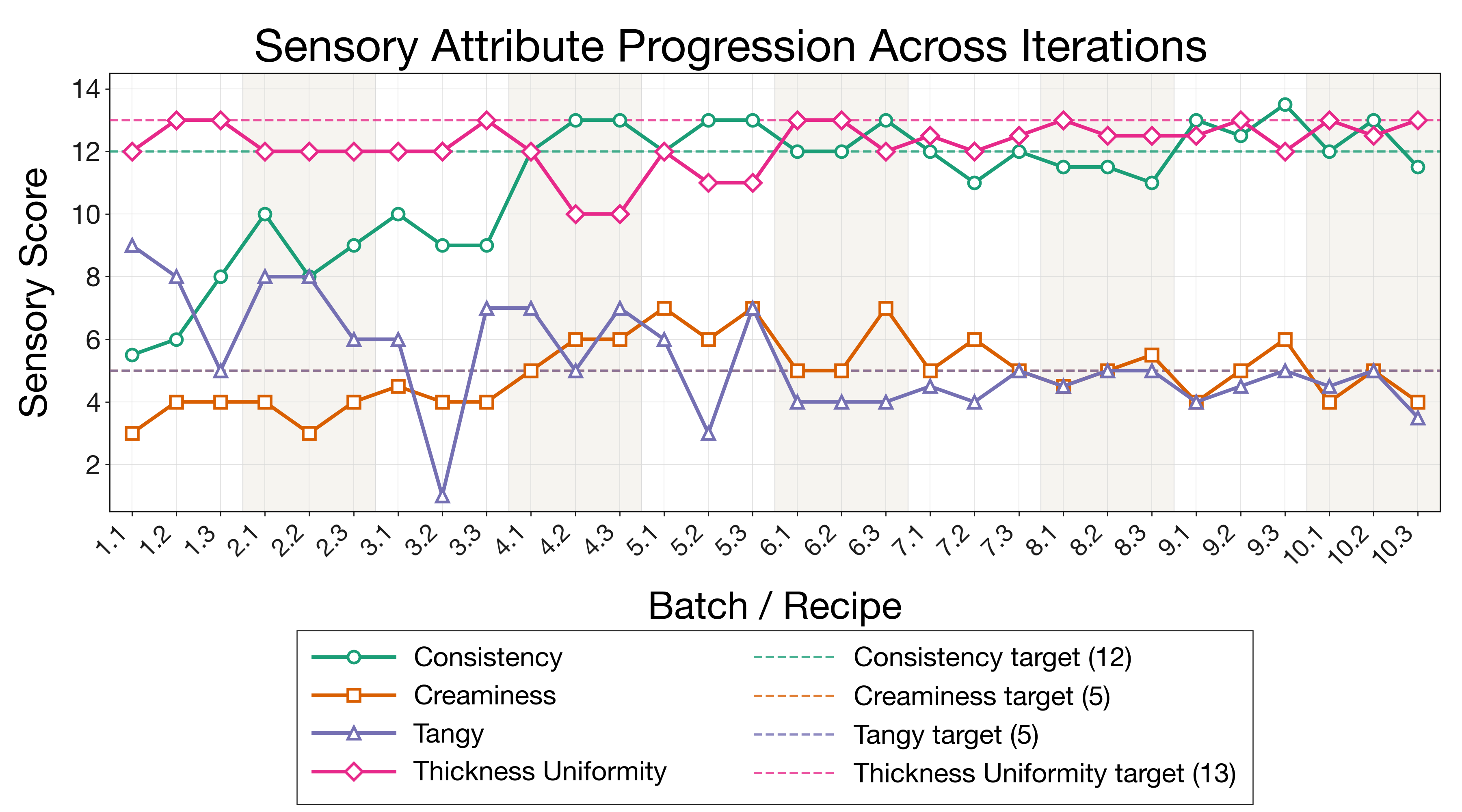}
    \caption{Sensory attribute trajectories across the 10 EGBO iterations (3 recipes per iteration) for the plant-based yogurt campaign. Solid lines show observed panel scores; dashed lines indicate the target value for each attribute. Consistency and Creaminess converge to their targets by iteration~7; Tanginess exhibits the highest volatility before settling at target; Uniformity remains near target throughout with minor drift.}
    \label{fig:yogurt_sensory_trajectory}
\end{figure}

\paragraph{Sensory convergence.} Figure~\ref{fig:yogurt_sensory_trajectory} shows the trajectory of all four sensory attributes. Consistency rose from 5 at baseline to 12 (exact target) by iteration~4, and remained within $\pm 1$ of target for all but one subsequent recipe (13.5 for recipe~9.3). Creaminess increased from 3 to 5 (exact target), first reaching it at iteration~4 and stabilizing from iteration~7 onward. Tanginess, which required a decrease from 10 to 5, proved the most volatile attribute, dropping sharply in early iterations but oscillating between 3 and 7 through the middle of the campaign before converging to 5 at iteration~7. Uniformity, which started at 13 (on target), drifted to values between 10 and 13 during the exploration-heavy early iterations but recovered to 12.5--13 in the final batches. The best single recipe (iteration~7, recipe~3) achieved scores of $(12,\; 5,\; 5,\; 12.5)$, matching the target exactly on three of four attributes and falling only 0.5 points short on Uniformity.

\begin{figure}[t]
    \centering
    \includegraphics[width=\linewidth]{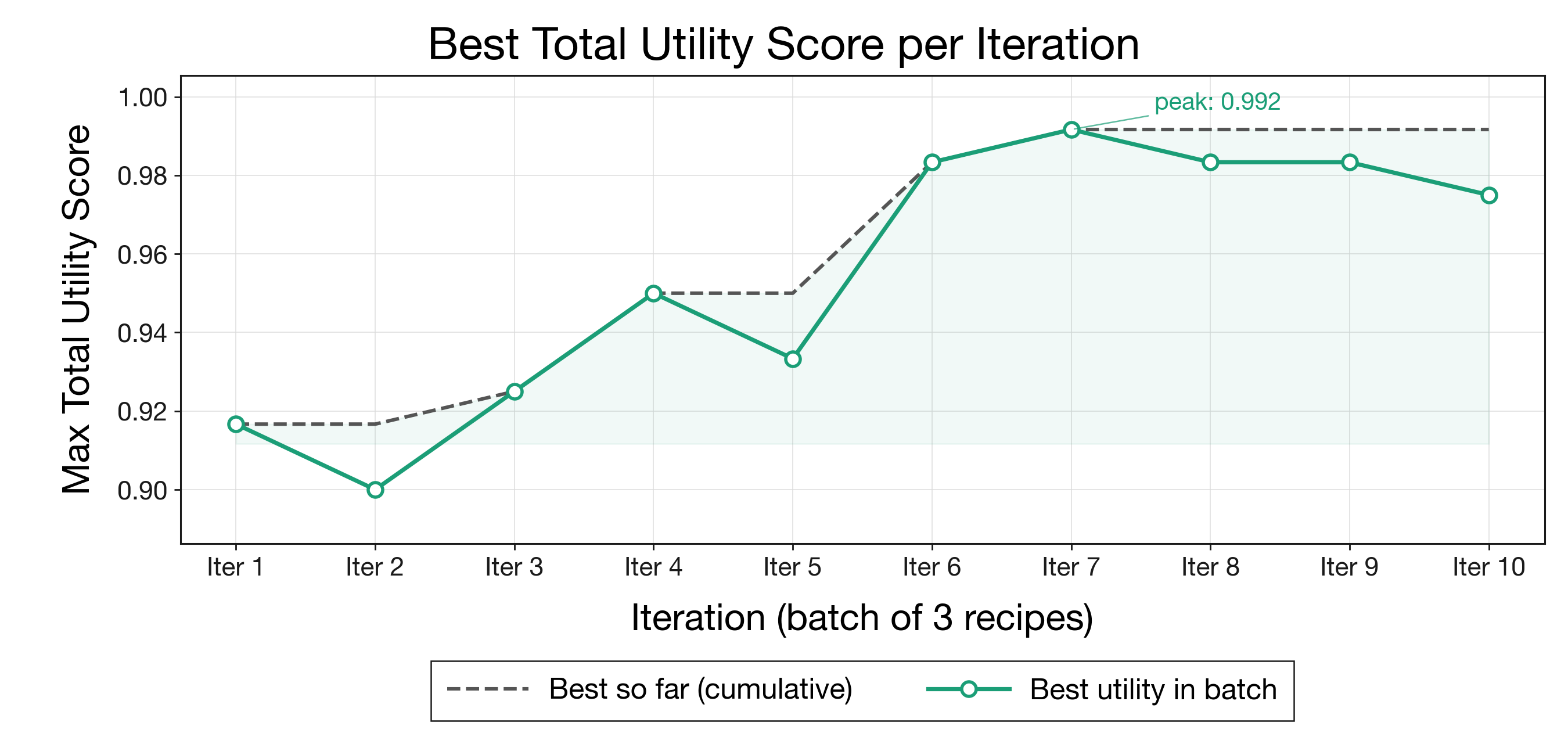}
    \caption{Best total utility score per iteration (solid) and cumulative best-so-far (dashed) across the 10 EGBO iterations. Utility rises from $f(x^{(0)}) = 0.767$ at baseline to a peak of $0.992$ at iteration~7, with the best-so-far remaining at or above $0.983$ from iteration~6 onward.}
    \label{fig:yogurt_utility_trajectory}
\end{figure}

\paragraph{Utility progression.} Figure~\ref{fig:yogurt_utility_trajectory} reports the maximum utility per iteration and the cumulative best-so-far. The utility climbed steeply in the first four iterations, reaching $0.950$ by iteration~4. A second jump to $0.983$ occurred at iteration~6, and the campaign peak of $0.992$ was reached at iteration~7. The best-so-far remained at or above $0.983$ from iteration~6 onward, indicating that the optimizer had largely converged by the sixth batch and subsequent iterations refined rather than explored. The residual gap to perfect utility is $1 - 0.992 = 0.008$.

\begin{figure}[t]
    \centering
    \includegraphics[width=\linewidth]{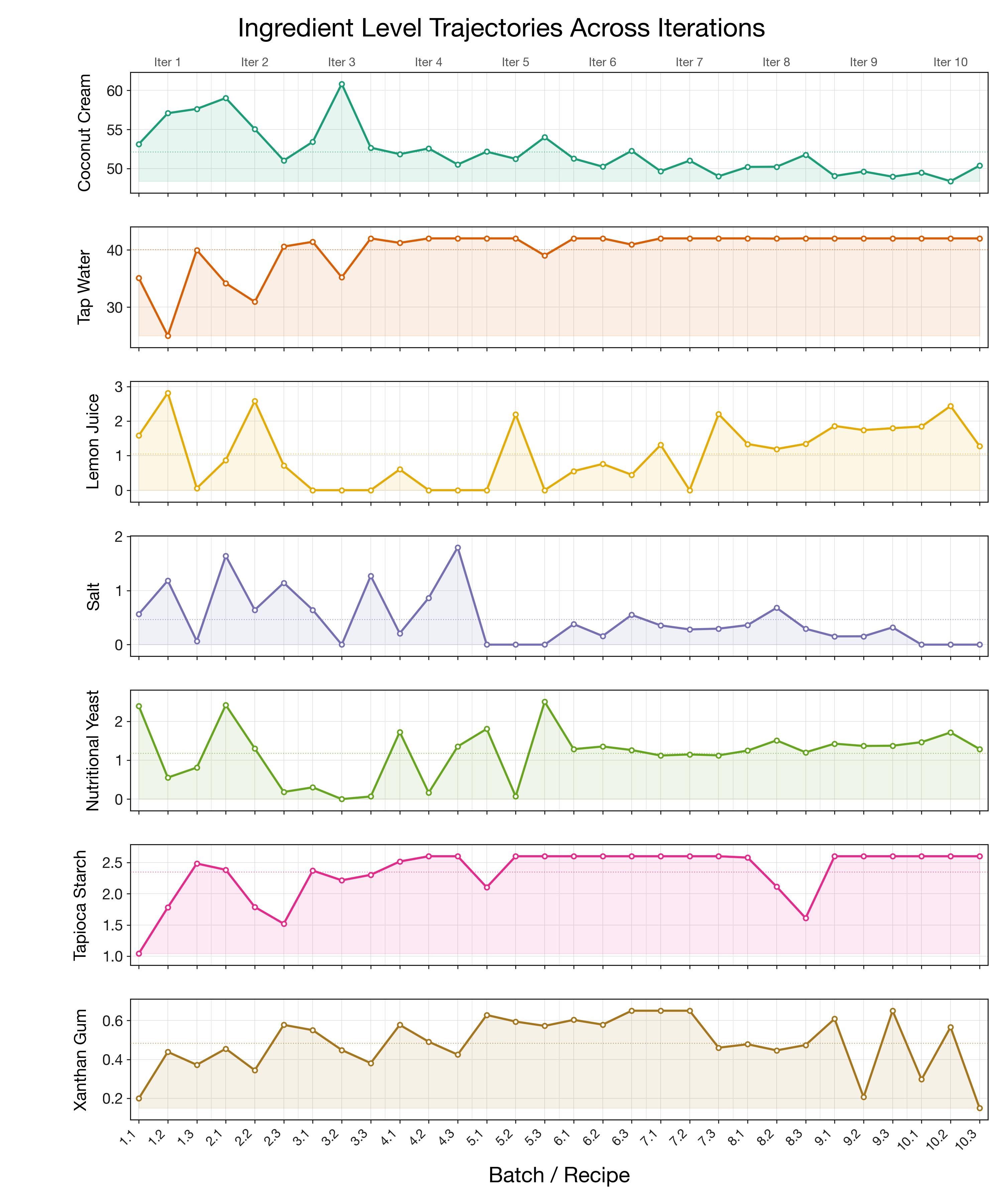}
    \caption{Ingredient concentration trajectories across the 10 EGBO iterations for the seven varying formulation dimensions. Each panel is independently scaled to reveal fine-grained adjustments: the dominant Coconut cream and Tap water rebalancing is visible alongside the smaller but consequential shifts in Lemon juice, Salt, Nutritional yeast, Tapioca starch, and Xanthan gum.}
    \label{fig:yogurt_active_set}
\end{figure}

\paragraph{Ingredient concentration trajectories.} Figure~\ref{fig:yogurt_active_set} shows how the seven active ingredients evolve over the campaign. Tap water and tapioca starch repeatedly take the same maximum values (42.0\% and 2.6\% of the recorded formulation; Table~\ref{tab:yogurt-ingredients} rescales the best recipe to a 630\,g batch) in most recipes from iteration~4 onward, consistent with binding upper bounds. The optimum may therefore lie outside the campaign's box, and widening these bounds is a natural next step.

\paragraph{Final formulation.} Table~\ref{tab:yogurt-ingredients} reports the FSA starting recipe and the best formulation found by EGBO (iteration~7, recipe~3). The formulation and preparation protocol are described in Appendix~\ref{app:yogurt-recipe}.

\begin{table}[h!]
\centering
\caption{Plant-based yogurt: ingredient composition of the FSA starting recipe ($f = 0.767$) and the best recipe found by EGBO (iteration~7, recipe~3; $f = 0.992$). Both recipes are scaled to a total batch mass of $630.00~\mathrm{g}$.}
\label{tab:yogurt-ingredients}
\begin{tabular}{lrrrrl}
\toprule
 & \multicolumn{2}{c}{Starting recipe} & \multicolumn{2}{c}{Best recipe} & \\
\cmidrule(lr){2-3} \cmidrule(lr){4-5}
Ingredient & Mass (g) & \% w/w & Mass (g) & \% w/w & Status \\
\midrule
Coconut cream, canned, sweetened & 340.00 & 53.97 & 311.92 & 49.50 & varied   \\
Tap water                        & 237.75 & 37.74 & 267.27 & 42.42 & varied   \\
Tapioca starch                   &   6.00 &  0.95 &  16.55 &  2.63 & varied   \\
Lemon juice, raw                 &  20.00 &  3.17 &  14.05 &  2.23 & varied   \\
Nutritional yeast                &  10.00 &  1.59 &   7.18 &  1.14 & varied   \\
Soy protein isolate              &   6.30 &  1.00 &   6.30 &  1.00 & constant \\
Xanthan gum                      &   1.00 &  0.16 &   2.89 &  0.46 & varied   \\
Tricalcium phosphate             &   1.95 &  0.31 &   1.95 &  0.31 & constant \\
Salt                             &   7.00 &  1.11 &   1.89 &  0.30 & varied   \\
\midrule
Total                            & 630.00 & 100.00 & 630.00 & 100.00 &         \\
\bottomrule
\end{tabular}
\end{table}

\begin{figure}\includegraphics[width=\textwidth]{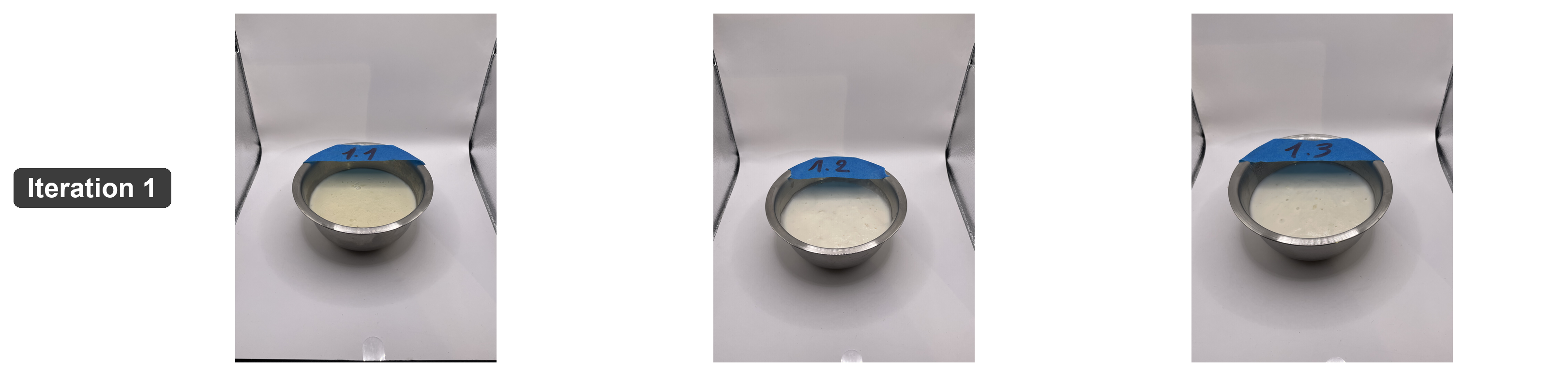}\caption{Plant-based yogurt: iteration 1.}\label{fig:iteration-01}\end{figure}

\begin{figure}\includegraphics[width=\textwidth]{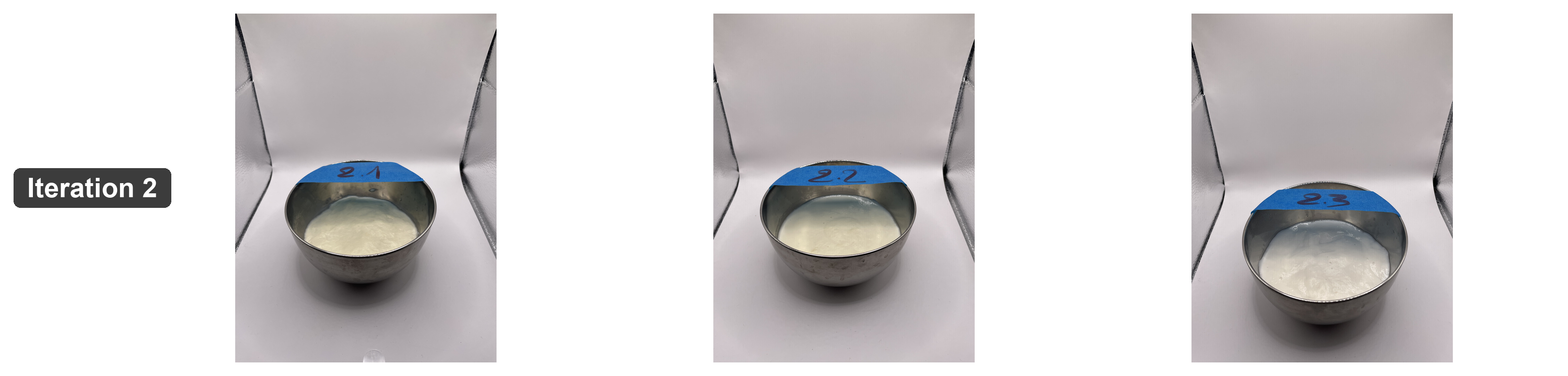}\caption{Plant-based yogurt: iteration 2.}\label{fig:iteration-02}\end{figure}

\begin{figure}\includegraphics[width=\textwidth]{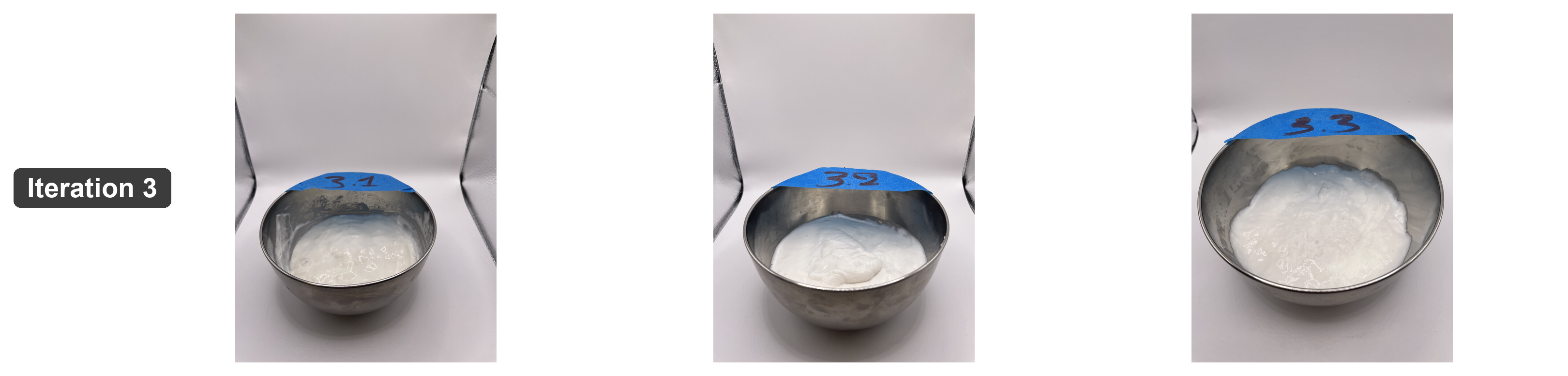}\caption{Plant-based yogurt: iteration 3.}\label{fig:iteration-03}\end{figure}

\begin{figure}\includegraphics[width=\textwidth]{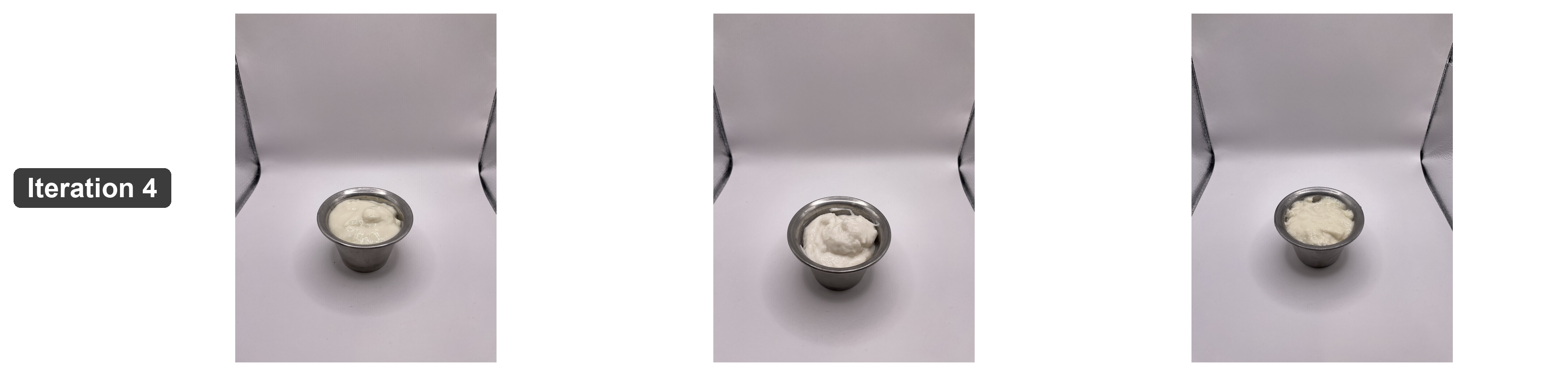}\caption{Plant-based yogurt: iteration 4.}\label{fig:iteration-04}\end{figure}

\begin{figure}\includegraphics[width=\textwidth]{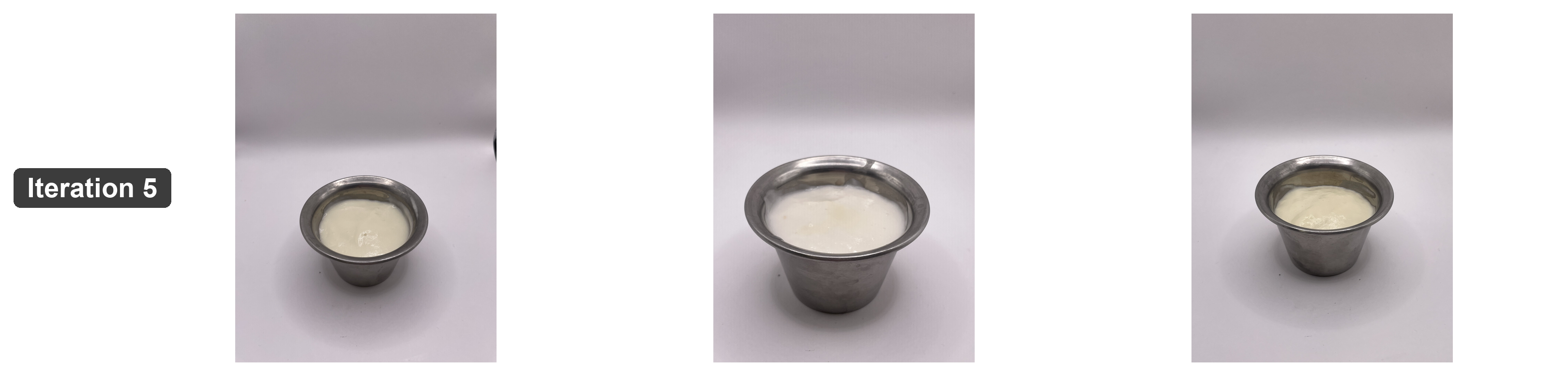}\caption{Plant-based yogurt: iteration 5.}\label{fig:iteration-05}\end{figure}

\begin{figure}\includegraphics[width=\textwidth]{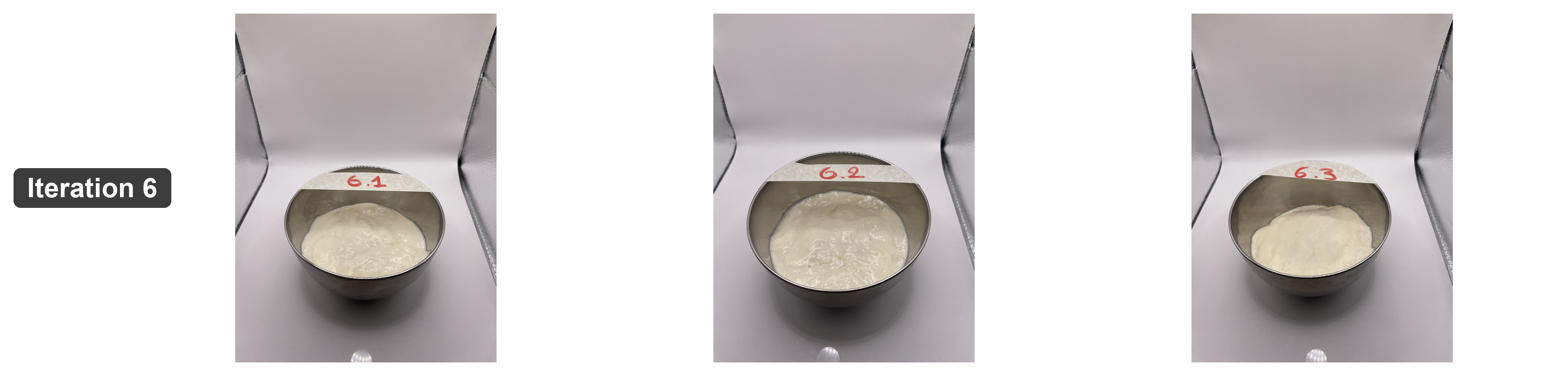}\caption{Plant-based yogurt: iteration 6.}\label{fig:iteration-06}\end{figure}

\begin{figure}\includegraphics[width=\textwidth]{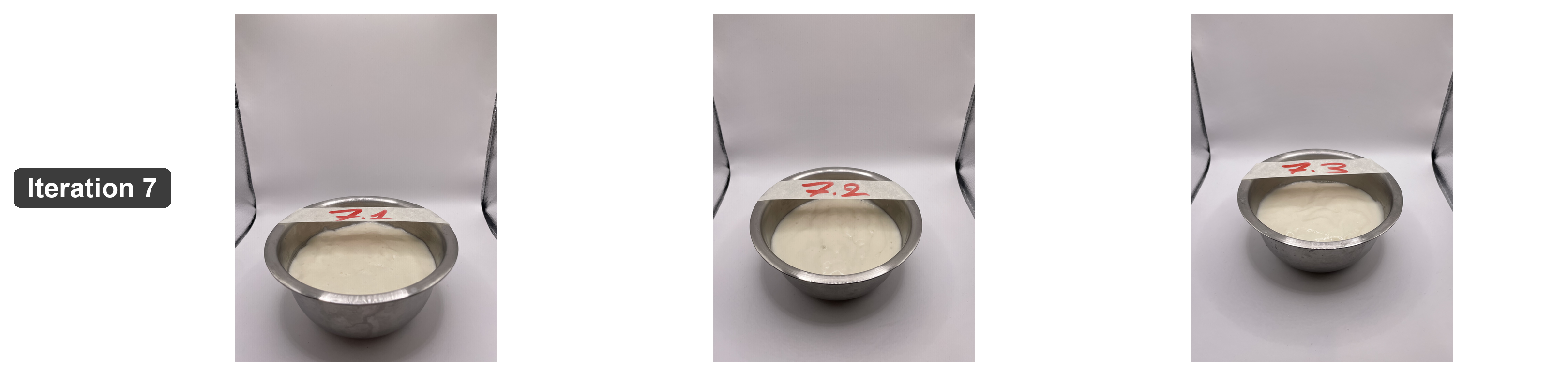}\caption{Plant-based yogurt: iteration 7.}\label{fig:iteration-07}\end{figure}

\begin{figure}\includegraphics[width=\textwidth]{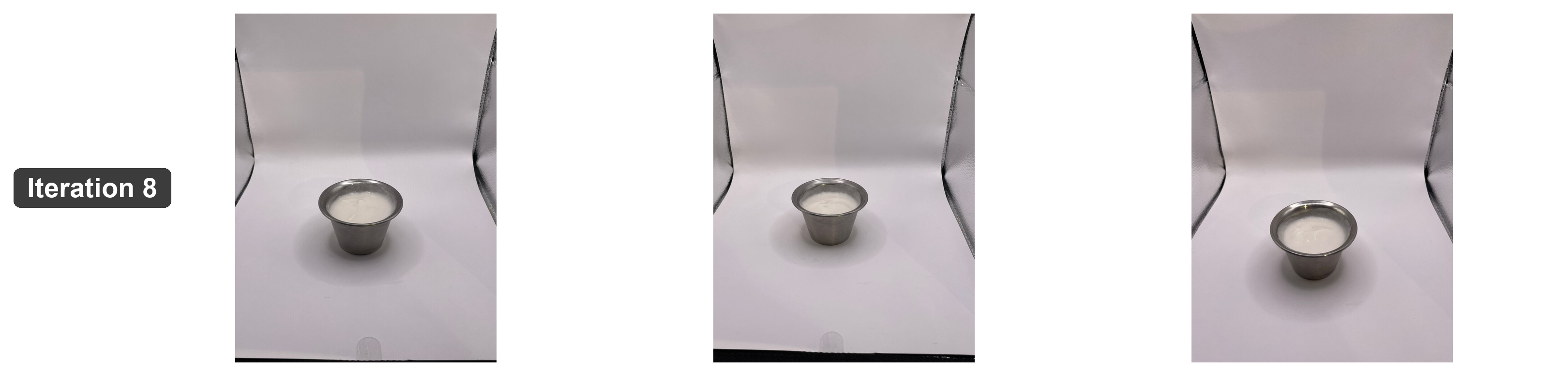}\caption{Plant-based yogurt: iteration 8.}\label{fig:iteration-08}\end{figure}

\begin{figure}\includegraphics[width=\textwidth]{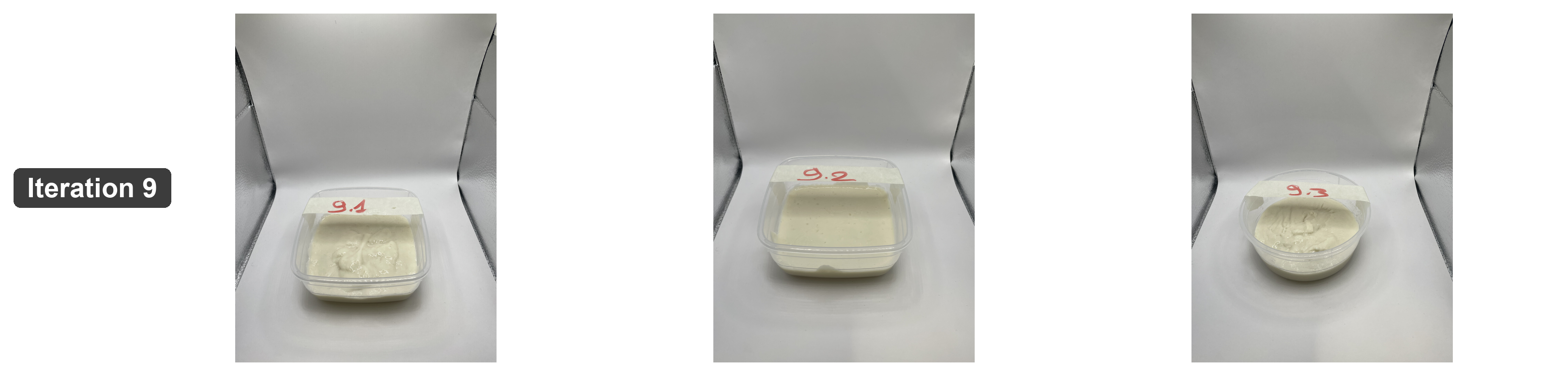}\caption{Plant-based yogurt: iteration 9.}\label{fig:iteration-09}\end{figure}

\begin{figure}\includegraphics[width=\textwidth]{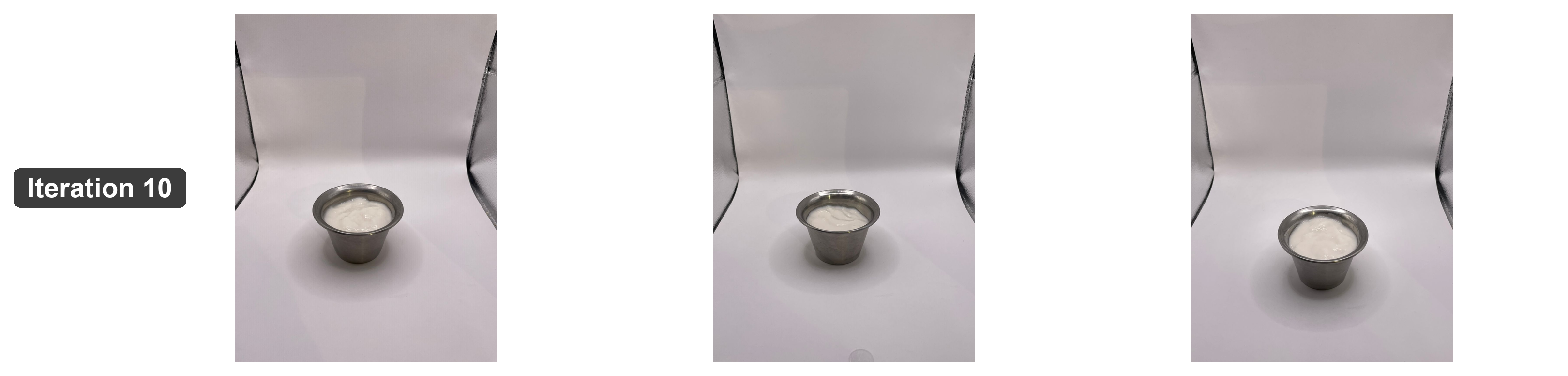}\caption{Plant-based yogurt: iteration 10.}\label{fig:iteration-10}\end{figure}

\paragraph{Iteration photos.} Figures~\ref{fig:iteration-01}--\ref{fig:iteration-10} show the visual evolution of the yogurt across iterations.

\subsection{Cottage cheese campaign}\label{app:cottage}

\begin{figure}[t]
    \centering
    \includegraphics[width=0.9\linewidth]{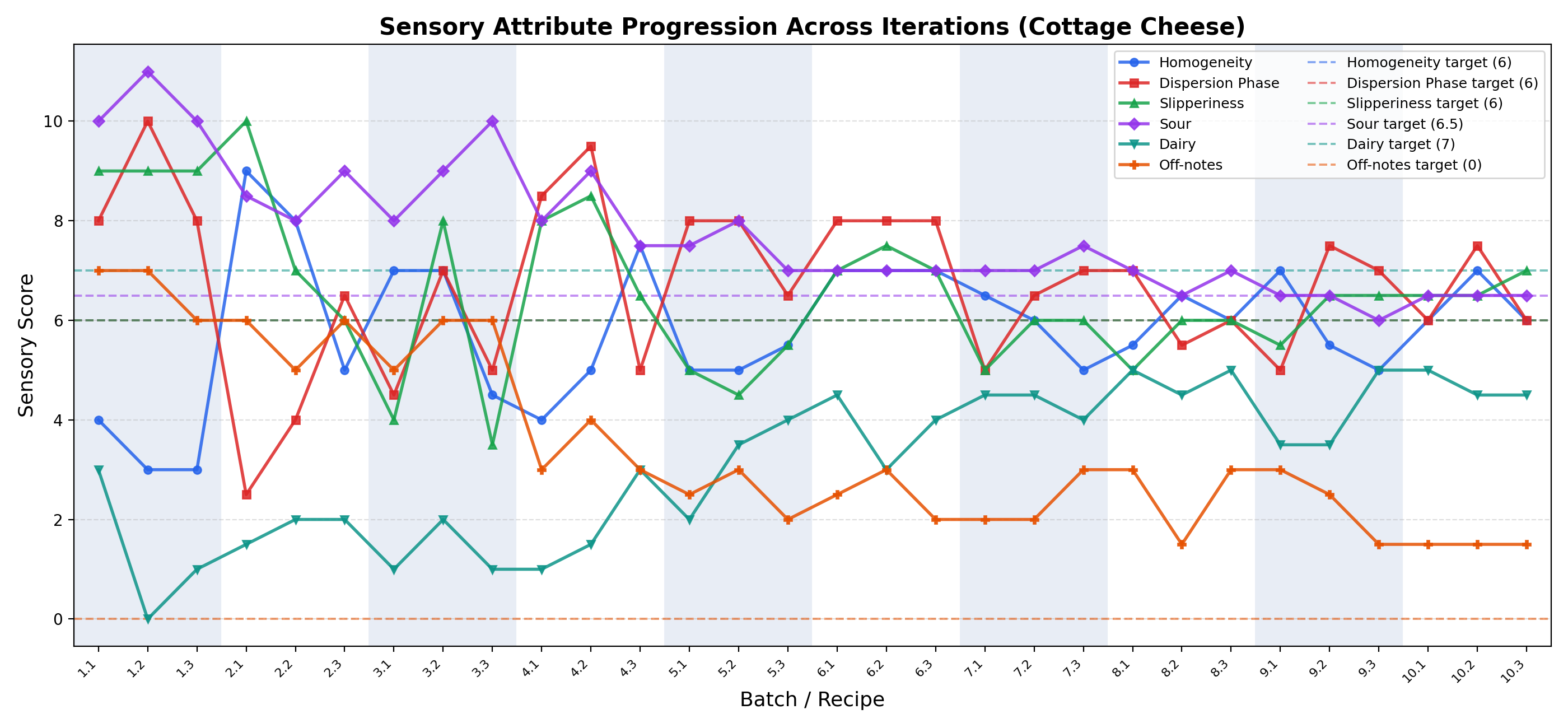}
    \caption{Sensory attribute trajectories across the 10 EGBO iterations (3 recipes per iteration, 30 total) for the plant-based cottage cheese campaign. Solid lines show observed panel scores; dashed lines indicate the target value for each attribute.}
    \label{fig:cottage_sensory_trajectory}
\end{figure}

\paragraph{Sensory convergence.} Figure~\ref{fig:cottage_sensory_trajectory} shows the trajectory of all six sensory attributes. Off-Notes, which required a decrease from 7 to 0, showed the most dramatic improvement, dropping steadily from 7 to 1.5 by the final iterations. Sourness decreased from 10 toward its target of 6.5, stabilizing in the range 6--7 from iteration~7 onward. Dairy, which started at 3 against a target of 7, proved the most difficult attribute to improve, rising gradually to 5 by the final iterations but not fully reaching target (which is expected since this is a plant-based dairy recipe). Homogeneity improved from 4 to values consistently near 6 (target) in the second half of the campaign. Dispersion Phase and Slipperiness, both starting well above their targets of 6, converged toward target by iteration~8, though with residual volatility. The best single recipe (iteration~10, recipe~1) achieved a utility of $0.975$, matching the target exactly on Homogeneity, Dispersion Phase and Sourness, with the largest residual gap on Dairy (5.0 vs.\ target 7).

\begin{figure}[t]
    \centering
    \includegraphics[width=\linewidth]{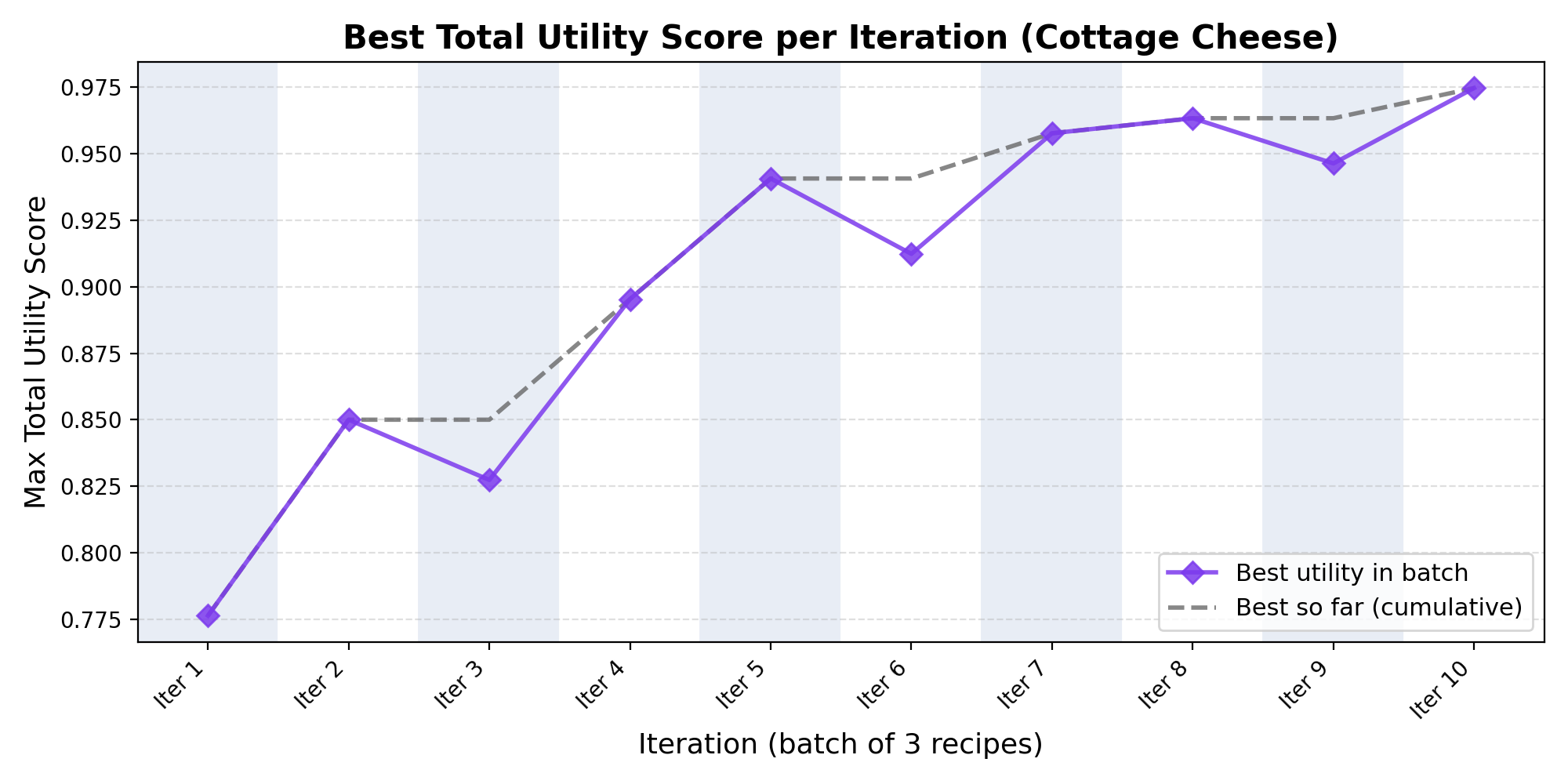}
    \caption{Best total utility score per iteration (solid) and cumulative best-so-far (dashed) across the 10 EGBO iterations for the cottage cheese campaign. Utility rises from $f(x^{(0)}) = 0.776$ at baseline to a peak of $0.975$ at iteration~10.}
    \label{fig:cottage_utility_trajectory}
\end{figure}

\paragraph{Utility progression.} Figure~\ref{fig:cottage_utility_trajectory} reports the maximum utility per iteration and the cumulative best-so-far. The utility rose steadily from $0.776$ at baseline, reaching $0.895$ at iteration~4 and $0.941$ at iteration~5. The best-so-far reached $0.958$ at iteration~7 and continued climbing to the campaign peak of $0.975$ at iteration~10, indicating that the optimizer had not fully plateaued and additional iterations could yield further gains.

\paragraph{Final formulation.} Table~\ref{tab:cottage_results} reports the per-attribute errors, and Table~\ref{tab:cottage-ingredients} the FSA starting recipe and the best recipe found by EGBO (recipe~10.1). The formulation and preparation protocol are described in Appendix~\ref{app:cottage-recipe}.

\begin{table}[t]
\centering
\caption{Plant-based cottage cheese real-world validation. The best recipe (iteration~10, recipe~1) shows the largest gains on Off-Notes and Sourness, with Dairy remaining the hardest attribute to reach. Overall utility improved 26\% relative to baseline. Utilities use the weights $w_j = 0.17$ described in Section~\ref{sec:real}, so the maximum is $1.02$.}
\label{tab:cottage_results}
\small
\begin{tabular}{lccccc}
\toprule
Attribute & Target & Baseline & Best & \multicolumn{2}{c}{Error} \\
\cmidrule(lr){5-6}
          &        &          &      & Before & After \\
\midrule
Homogeneity      & 6   & 4  & 6    & 2   & 0   \\
Dispersion Phase & 6   & 8  & 6    & 2   & 0   \\
Slipperiness     & 6   & 9  & 6.5  & 3   & 0.5 \\
Sourness         & 6.5 & 10 & 6.5  & 3.5 & 0   \\
Dairy            & 7   & 3  & 5    & 4   & 2   \\
Off-Notes        & 0   & 7  & 1.5  & 7   & 1.5 \\
\midrule
Utility $f(x)$ & 1.020 & 0.776 & 0.975 & 0.244 & 0.045 \\
\bottomrule
\end{tabular}
\end{table}

\begin{table}[h!]
\centering
\caption{Plant-based cottage cheese: ingredient composition of the FSA starting recipe (recipe~1.1; $f = 0.776$) and the best recipe found by EGBO (recipe~10.1; $f = 0.975$). Curd-substream ingredients (top) are coagulated and centrifuged before being recombined with the cooked dressing (bottom).}
\label{tab:cottage-ingredients}
\begin{tabular}{lrrl}
\toprule
 & Starting recipe & Best recipe & \\
Ingredient & \% w/w & \% w/w & Substream \\
\midrule
Soy milk, unsweetened, plain, shelf stable & 63.12 & 70.00 & curd     \\
Distilled vinegar                          &  6.00 &  4.10 & curd     \\
\midrule
Tap water                                  & 10.38 &  8.00 & dressing \\
Coconut cream, canned, sweetened           & 18.06 & 15.81 & dressing \\
Lemon juice, raw                           &  0.55 &  0.40 & dressing \\
Nutritional yeast                          &  0.68 &  0.50 & dressing \\
Salt                                       &  0.49 &  0.59 & dressing \\
Tapioca starch                             &  0.68 &  0.57 & dressing \\
Xanthan gum                                &  0.04 &  0.03 & dressing \\
\midrule
Total                                      & 100.00 & 100.00 &         \\
\bottomrule
\end{tabular}
\end{table}

\subsection{Cross-campaign comparison}

\begin{figure}[t]
    \centering
    \includegraphics[width=\linewidth]{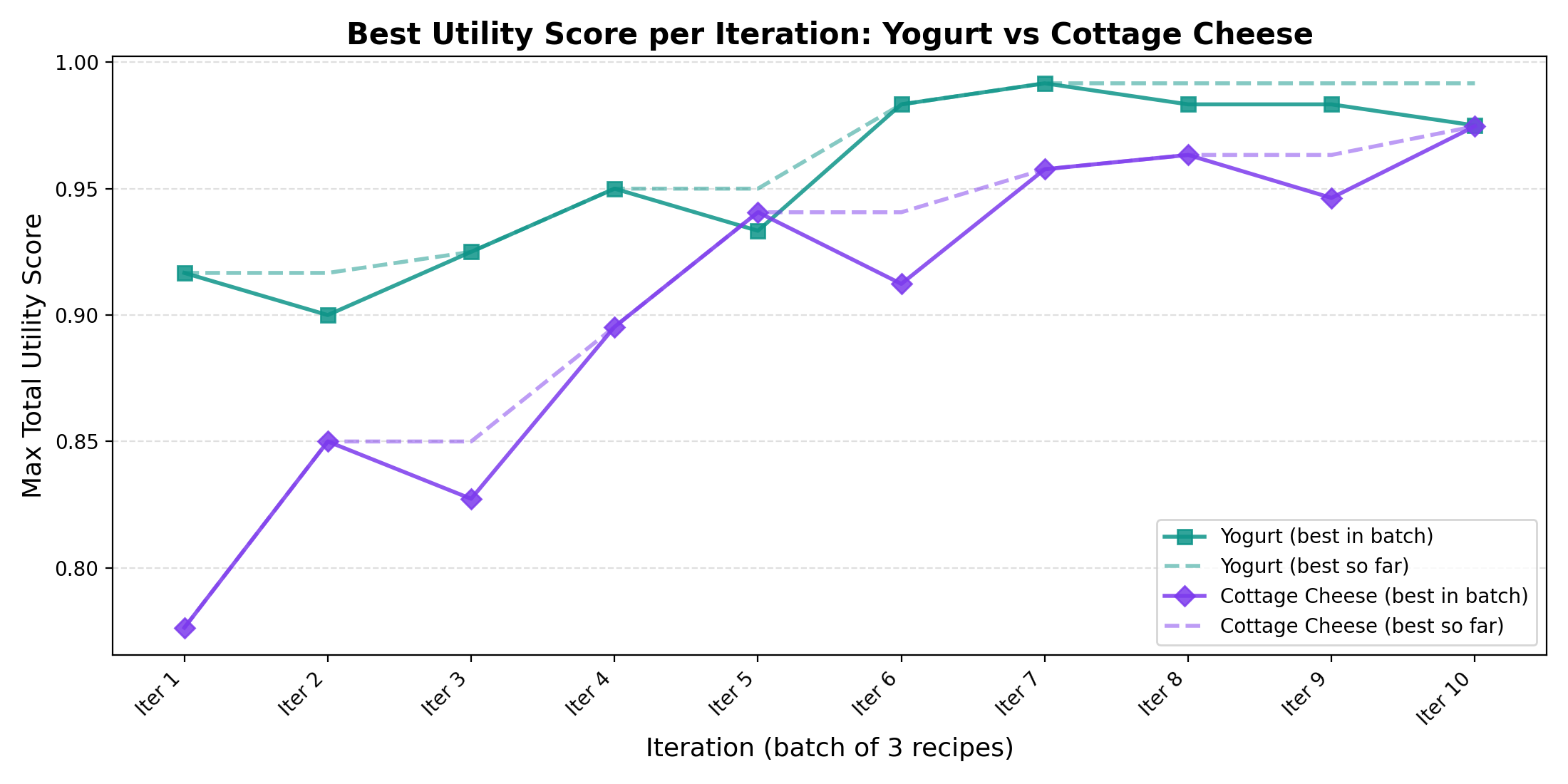}
    \caption{Comparison of best-so-far utility across both real-world campaigns. Yogurt converges faster (peaking at $0.992$ by iteration~7), consistent with its lower-dimensional sensory objective ($m{=}4$, $d{=}7$). Cottage cheese converges more slowly (peaking at $0.975$ at iteration~10), reflecting its higher-dimensional objective ($m{=}6$, $d{=}9$).}
    \label{fig:combined_utility}
\end{figure}

Figure~\ref{fig:combined_utility} overlays the best-so-far utility curves for both campaigns. Both start from similar baselines generated by the FSA ($0.767$ for yogurt, $0.776$ for cottage cheese) and follow a characteristic steep-then-plateau trajectory. Yogurt converges faster, reaching $0.983$ by iteration~6 and peaking at $0.992$ at iteration~7, while cottage cheese reaches $0.958$ by iteration~7 and peaks at $0.975$ at iteration~10. The slower convergence of cottage cheese is consistent with its higher-dimensional sensory objective ($m = 6$ vs.\ $m = 4$) and larger active ingredient set ($d = 9$ vs.\ $d = 7$), both of which increase the effective optimization difficulty. Nevertheless, both campaigns achieved substantial improvements --- 29\% and 26\% respectively --- within the same 10-iteration budget, suggesting that EGBO transfers across product categories with different sensory structures and ingredient spaces, considering also that it started from a $>$ 0.75 utility.

\subsection{EGBO vs.\ Human Food Scientist}\label{app:hfs}

\subsubsection{Protocol}\label{app:human-protocol}

The EGBO vs.\ HFS comparison (Section~\ref{sec:human-comparison}) was conducted under the following instructions and constraints, designed to ensure a controlled comparison while respecting the inherently different workflows of algorithmic and manual formulation.

\paragraph{HFS profile.} HFS is a professional food scientist and chef with prior experience in recipe formulation but no prior experience specifically with plant-based dairy products. The operator was not given access to the EGBO system, its suggestions, or its results at any point during the experiment. HFS participated voluntarily as a professional collaborator; the comparison was designed in mutual agreement and HFS was not a research subject.

\paragraph{Shared constraints.}
\begin{enumerate}
    \item \textbf{Sensory targets.} HFS was given the same four sensory objectives used in the EGBO campaign, assessed on a 0 to 15 scale: Consistency (target 12), Creaminess (target 5), Tanginess (target 5), and Uniformity (target 13).
    \item \textbf{Utility metric.} The scalar utility score (Equation~\ref{eq:scalar-utility}) was computed for each evaluated formulation and reported to the operator after each evaluation.
    \item \textbf{Plant-based constraint.} All ingredients must be plant-based; no animal-derived ingredients were permitted.
    \item \textbf{Time tracking.} HFS documented the time spent on each trial or iteration, including preparation, cooling, tasting, and decision-making time.
    \item \textbf{Documentation.} All ingredient quantities, preparation steps, and sensory observations were recorded for each trial.
\end{enumerate}

\paragraph{Experiment 1: full workflow comparison.}
\begin{enumerate}
    \item \textbf{Objective.} Develop, from scratch, a plant-based yogurt recipe that matches the target sensory profile as closely as possible.
    \item \textbf{No starting recipe.} The operator was free to choose any plant-based ingredients and any preparation method.
    \item \textbf{No budget constraint.} There was no limit on the number of trials. The operator was instructed to continue until satisfied with the recipe or until further progress seemed unlikely.
    \item \textbf{Trial definition.} A trial is defined as a complete attempt starting from raw ingredients. If the operator modifies a formulation mid-preparation, this is considered part of the same trial. If the operator stops and starts over from the beginning, a new trial begins. Trials abandoned before producing a tasteable result were recorded as failed trials.
    \item \textbf{Evaluation.} Formulations that the operator deemed sufficiently promising were evaluated for sensory scores on the four target attributes and the corresponding utility was computed and shared with the operator.
    \item \textbf{Workflow.} The scientist followed a conventional iterative workflow, preparing a candidate formulation, letting it cool, tasting and evaluating it, and then deciding on the next modification or restarting.
\end{enumerate}

\paragraph{Experiment 2: same-start refinement comparison.}
\begin{enumerate}
    \item \textbf{Objective.} Optimize the FSA's starting recipe ($f(x^{(0)}) = 0.767$) toward the same four sensory targets.
    \item \textbf{Fixed starting point.} The operator was given the exact recipe (ingredient list and concentrations) produced by the FSA, the same starting recipe used by EGBO.
    \item \textbf{Budget.} Maximum of 10 iterations, matching the EGBO budget.
    \item \textbf{One recipe per iteration.} The scientist chose to produce one recipe per iteration, judging a batch of 3 independent variants infeasible because each formulation decision depended sequentially on the sensory evaluation of the previous one.
    \item \textbf{Sensory feedback.} After each iteration, the formulation was evaluated by the sensory panel on the four target attributes, and the scores and resulting utility were reported to the operator before the next iteration began.
    \item \textbf{Permitted modifications.} The operator was free to adjust any ingredient concentration, add new plant-based ingredients, or remove ingredients, subject to the plant-based constraint.
    \item \textbf{Time per iteration.} Each iteration consisted of recipe design, preparation, cooling, and sensory evaluation, requiring approximately 4 hours.
\end{enumerate}

\paragraph{Key differences between EGBO and HFS workflows.} Table~\ref{tab:workflow_comparison} summarizes the structural differences between the two workflows.

\begin{table}[h]
\centering
\caption{Structural differences between EGBO and HFS workflows in Experiment~2.}
\label{tab:workflow_comparison}
\small
\begin{tabular}{lcc}
\toprule
 & EGBO & HFS \\
\midrule
Recipes per iteration    & 3 (batch)    & 1 (sequential) \\
Total recipes in 10 iter & 30           & 10 \\
Recipe design time       & Seconds      & Hours \\
Decision basis           & GP surrogate & Expert judgment \\
Regression possible      & Unlikely     & Observed \\
\bottomrule
\end{tabular}
\end{table}

The batch vs.\ sequential distinction is not an artificial handicap imposed on HFS but a genuine consequence of the two workflows. EGBO's acquisition function can propose multiple diverse points in a single optimization step because it reasons over a global surrogate model. HFS, by contrast, reasons sequentially. Each modification is informed by the sensory outcome of the previous one, and designing three independent variants in parallel would require the operator to mentally maintain three separate hypotheses about ingredient interactions without feedback.

\subsubsection{Results}
\begin{figure}[t]
    \centering
    \includegraphics[width=\linewidth]{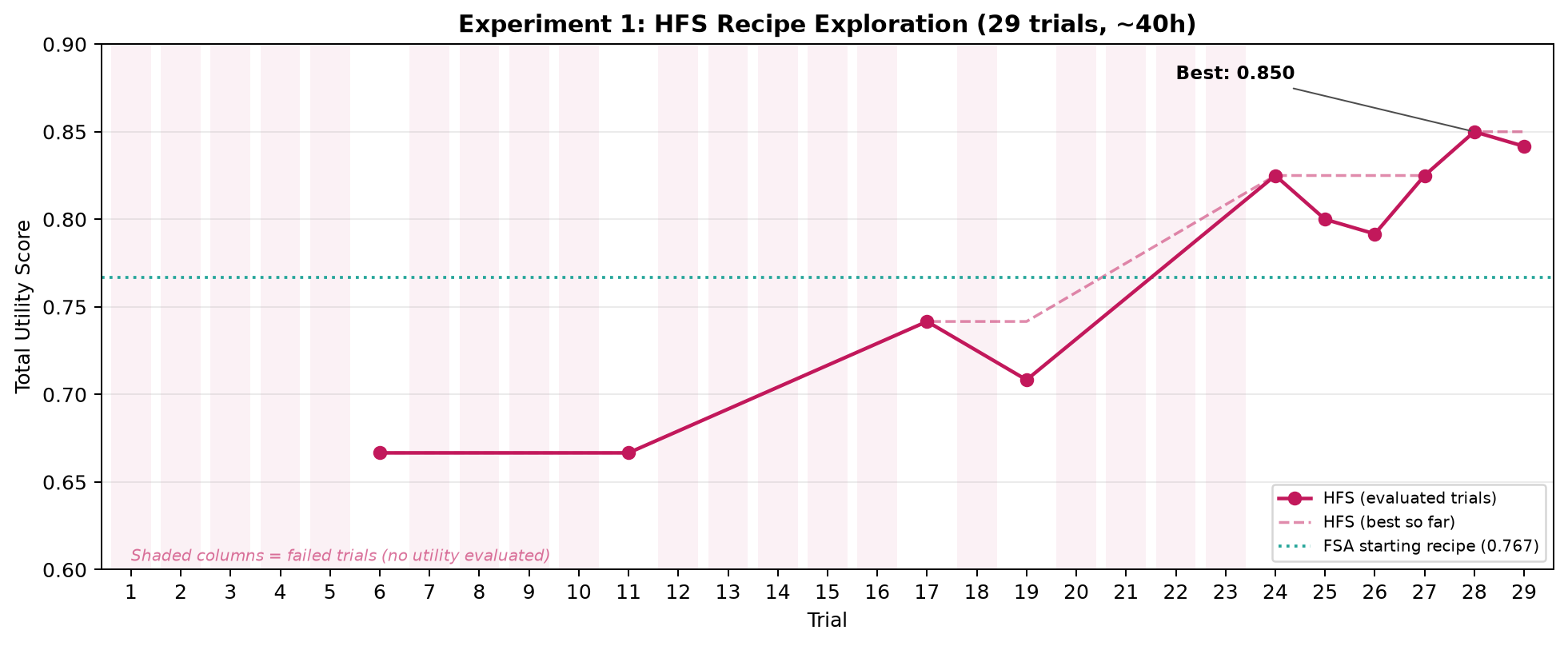}
    \caption{Experiment~1: HFS recipe exploration. The operator executed 29 trials over ${\sim}$40\,h, of which 10 produced formulations evaluated for utility. The dotted line indicates the utility of the FSA's starting recipe (0.767), produced in negligible time. HFS's best result (0.850, trial~28) required the full 40-hour exploration budget.}
    \label{fig:leg1_utility}
\end{figure}

\begin{figure}[t]
    \centering
    \includegraphics[width=\linewidth]{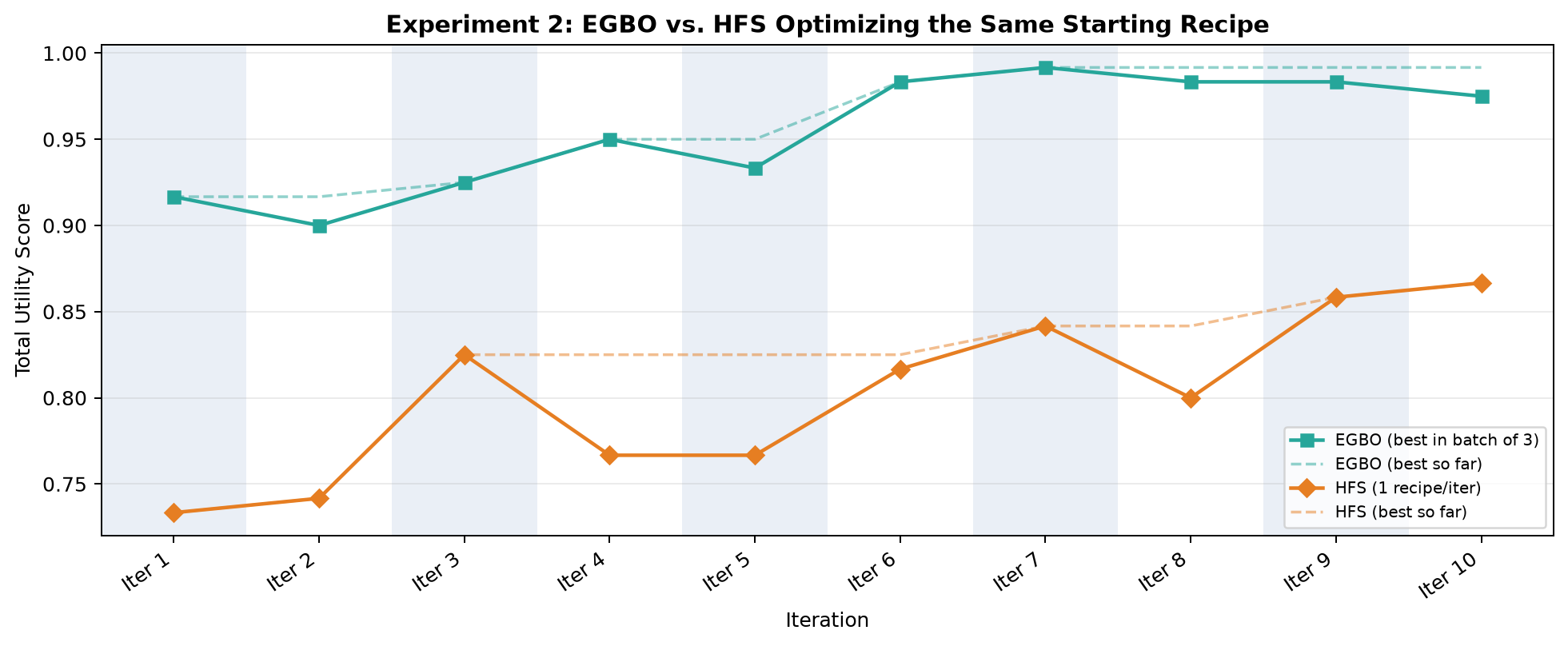}
    \caption{Experiment~2: EGBO vs.\ HFS, both optimizing the same starting recipe ($f(x^{(0)}) = 0.767$) over 10 iterations. EGBO proposes 3 recipes per iteration; HFS produces 1. EGBO reaches $0.992$ by iteration~7, while HFS reaches $0.867$ by iteration~10, a 14.4\% gap.}
    \label{fig:leg2_comparison}
\end{figure}

Figures~\ref{fig:leg1_utility} and~\ref{fig:leg2_comparison} show the utility trajectories for Experiments~1 and~2; Section~\ref{sec:human-comparison} discusses the results.

\clearpage

\end{document}